\documentclass[onefignum, onetabnum]{siamonline190516} %

\usepackage[utf8]{inputenc}
\usepackage{amsfonts, amsmath, amssymb, amsopn} %
\usepackage{braket}
\usepackage{mathtools, xparse}
\usepackage{bbm}
\def\eqref#1{equation~\ref{#1}}

\def\1{\bm{1}}

\def\eps{{\epsilon}}

\DeclareMathAlphabet{\mathsfit}{\encodingdefault}{\sfdefault}{m}{sl}
\SetMathAlphabet{\mathsfit}{bold}{\encodingdefault}{\sfdefault}{bx}{n}

\def\gA{{\mathcal{A}}}

\def\gD{{\mathcal{D}}}
\def\gE{{\mathcal{E}}}
\def\gF{{\mathcal{F}}}
\def\gG{{\mathcal{G}}}

\def\gN{{\mathcal{N}}}
\def\gO{{\mathcal{O}}}

\def\gS{{\mathcal{S}}}

\def\gU{{\mathcal{U}}}

\def\gX{{\mathcal{X}}}

\def\sI{{\mathbb{I}}}

\def\sP{{\mathbb{P}}}

\def\sR{{\mathbb{R}}}
\def\sS{{\mathbb{S}}}

\newcommand{\E}{\mathbb{E}}

\newcommand{\R}{\mathbb{R}}

\newcommand{\sgn}{\mathrm{sgn}}

\newcommand{\diag}{\text{diag}}

\DeclareMathOperator*{\argmax}{arg\,max}
\DeclareMathOperator*{\argmin}{arg\,min}

\DeclarePairedDelimiter\norm{\lVert}{\rVert}
\DeclarePairedDelimiter\bignorm{\big\lVert}{\big\rVert}
\DeclarePairedDelimiter\Bignorm{\Big\lVert}{\Big\rVert}
\DeclarePairedDelimiter\biggnorm{\bigg\lVert}{\bigg\rVert}
\DeclarePairedDelimiter\enorm{\lVert}{\rVert_2}
\DeclareMathOperator{\ST}{s. t. }
\DeclarePairedDelimiter\rbr{(}{)}

\DeclarePairedDelimiterX{\inp}[2]{\big\langle}{\big\rangle}{#1, #2}

\newcommand{\jm}{_{j = 1}^m}
\newcommand{\jms}{_{j = 1}^{m^\star}}
\newcommand{\mhs}{\widehat{m}^\star}
\newcommand{\jmh}{_{j = 1}^{\widehat{m}}}
\newcommand{\jmhs}{_{j = 1}^{\widehat{m}^\star}}

\newcommand{\iP}{_{i = 1}^P}

\newcommand{\kn}{_{k=1}^n}
\newcommand{\iN}{_{i=1}^N}
\newcommand{\iNN}{_{i=1}^{N+1}}
\newcommand{\iNo}{_{i=1}^{N_1}}

\newcommand{\iUt}{_{i=1}^{U_t}}
\newcommand{\iUT}{_{i=1}^{U_T}}
\newcommand{\itildeP}{_{i = 1}^{\widetilde P}}
\newcommand{\Pconv}{P_{\mathrm{conv}}}
\newcommand{\iPconv}{_{i=1}^{\Pconv}}
\newcommand{\iPhat}{_{i = 1}^{\widehat{P}}}
\newcommand{\iPs}{_{h = 1}^{P_s}}
\newcommand{\iPss}{_{h = 1}^{P_s + 1}}

\newcommand{\allDelta}{_{\Delta: X+\Delta \in \gX}}
\newcommand{\allDeltaU}{_{\Delta: X+\Delta \in \gU}}
\newcommand{\allkinn}{\forall k \in [n]}
\newcommand{\alliinP}{\forall i \in [P]}
\newcommand{\alliinPs}{\forall h \in [P_s]}
\newcommand{\alliinPss}{\forall h \in [P_s + 1]}
\newcommand{\alliinPhat}{\forall i \in [\widehat{P}]}

\newcommand{\kone}{_{k: y_k=1}}
\newcommand{\kzero}{_{k: y_k=0}}
\newcommand{\one}{\mathbf{1}}

\newcommand{\yhat}{\widehat{y}}

\newcommand{\prox}{\texttt{prox}}

\newcommand{\ytil}{\tilde{y}}
\newcommand{\stil}{\tilde{s}}
\newcommand{\ttil}{\tilde{t}}

\renewcommand{\th}{^\text{th}}

\newcommand{\lvw}{l_{\mathrm{ADMM}}^{v,w}}
\newcommand{\lua}{l_{\mathrm{ADMM}}^{u,\alpha}}
\newcommand{\rob}{\mathrm{rob}}

\newcommand{\bw}{\mathbf{w}}
\newcommand{\bu}{\mathbf{u}}
\newcommand{\bU}{\mathbf{U}}
\newcommand{\bs}{\mathbf{s}}

\newcommand{\vits}{\widetilde{v}_i^\star}
\newcommand{\wits}{\widetilde{w}_i^\star}
\newcommand{\util}{\widetilde{u}}
\newcommand{\altil}{\widetilde{\alpha}}
\newcommand{\mts}{\widetilde{m}^\star}

\newcommand{\Zbar}{\overline{Z}}
\newcommand{\Dbar}{\overline{D}}
\newcommand{\dbar}{\overline{d}}

\newcommand{\DeltaStar}{\Delta^\star_{v,w}}
\newcommand{\DeltaStarTil}{\widetilde{\Delta}^\star_{v,w}}

\usepackage[numbers]{natbib}
\usepackage{algorithm,algpseudocode}
\usepackage{adjustbox}
\usepackage{booktabs}
\usepackage{subcaption}
\usepackage[shortlabels]{enumitem}
\usepackage{makecell}
\ifpdf
  \DeclareGraphicsExtensions{.eps,.pdf,.png,.jpg}
\else
  \DeclareGraphicsExtensions{.eps}
\fi

\usepackage{enumitem}
\setlist[enumerate]{leftmargin=.5in}
\setlist[itemize]{leftmargin=.5in}

\newsiamremark{remark}{Remark}
\newsiamremark{hypothesis}{Hypothesis}
\crefname{hypothesis}{Hypothesis}{Hypotheses}
\newsiamthm{claim}{Claim}

\headers{Efficient Two-layer ANN Global Optimization}{Y. Bai, T. Gautam, and S. Sojoudi}

\title{
Efficient Global Optimization of Two-layer ReLU Networks: Quadratic-time Algorithms and Adversarial Training\thanks{
This work is an extension of \citep{Bai22a} and was supported by grants from ONR and NSF.
}}

\author{Yatong Bai\thanks{Department of Mechanical Engineering, University of California, Berkeley, (\email{yatong\_bai@berkeley.edu}).}
\and Tanmay Gautam\thanks{Department of Electrical Engineering and Computer Science, University of California, Berkeley, (\email{tgautam23@berkeley.edu}).}
\and Somayeh Sojoudi\thanks{Department of Mechanical Engineering and Department of Electrical Engineering and Computer Science, University of California, Berkeley, (\email{sojoudi@berkeley.edu}).}}

\ifpdf
\hypersetup{
  pdftitle = {Efficient Global Optimization of Two-layer ReLU Networks: Quadratic-time Algorithms and Adversarial Training},
  pdfauthor= {Y. Bai, T. Gautam, and S. Sojoudi}
}
\fi

\begin{document}

\maketitle

\setlength{\parindent}{0em}
\setlength{\parskip}{.5em}

\algblockdefx{MRepeat}{EndRepeat}{\textbf{repeat}}{\textbf{end repeat}}

\begin{abstract}
    The non-convexity of the artificial neural network (ANN) training landscape brings optimization difficulties.
    While traditional back-propagation gradient-based algorithms are effective in certain cases, they can become stuck at spurious local minima and are sensitive to initializations and hyperparameters.
    Recent work has shown that training a ReLU-activated ANN can be reformulated as a convex program, bringing hope to globally optimizing interpretable ANNs.
    However, na\"ively solving the convex training formulation has exponential complexity, and even an approximation heuristic requires cubic time.
    In this work, we characterize the quality of this approximation and develop two efficient algorithms that train ANNs with global convergence guarantees.
    The first algorithm is based on the alternating direction method of multipliers (ADMM).
    It can solve both the exact convex formulation and the approximate counterpart, and generalizes to a family of convex training formulations.
    Linear global convergence is achieved, and the initial several iterations often yield a solution with high prediction accuracy.
    When solving the approximate formulation, the per-iteration time complexity is quadratic.
    The second algorithm, based on the ``sampled convex programs'' theory, solves unconstrained convex formulations and converges to an approximately globally optimal classifier.
    The non-convexity of the ANN training landscape is exacerbated when adversarial training is considered.
    We apply robust convex optimization theory to convex training and develop convex formulations that train ANNs robust to adversarial inputs.
    Our analysis explicitly focuses on one-hidden-layer fully connected ANNs, but can extend to more sophisticated architectures.
\end{abstract}



\section{Introduction}\label{sec:intro}

The artificial neural network (ANN) is one of the most powerful and popular machine learning tools.
Optimizing a typical ANN with non-linear activation functions and a finite width requires solving non-convex optimization problems.
Traditionally, training ANNs relies on stochastic gradient descent (SGD) back-propagation \citep{Rumelhart1986}.
Despite its tremendous empirical success, this algorithm is only guaranteed to converge to a local minimum when applied to the non-convex ANN training objective.
While SGD back-propagation can converge to a global optimizer for one-hidden-layer ``rectified linear unit (ReLU)''-activated networks when the considered network is wide enough \citep{Pilanci20b, du2018gradient} or when the inputs follow a Gaussian distribution \citep{Brutzkus17}, spurious local minima can exist in general applications. 
Moreover, the non-convexity of the training landscape and the properties of back-propagation SGD cause the issues listed below:

\begin{itemize}[leftmargin=8mm]
    \setlength\itemsep{.2em}
    \item \textbf{Poor interpretability:}
    With SGD, it is hard to monitor the training status.
    For example, when the progress slows down, we may or may not be close to a local minimum, and the local minimum may be spurious.
    \item \textbf{High sensitivity to hyperparameters:}
    Back-propagation SGD has several important hyperparameters to tune.
    Every parameter is crucial to the performance, but selecting the parameters can be difficult. SGD is also sensitive to the initialization \citep{He15init}.
    \item \textbf{Vanishing / exploding gradients:}
    With back-propagation, the gradient at shallower layers can be tiny (or huge) if the deeper layer weights are tiny (or huge).
\end{itemize}

While more advanced back-propagation optimizers such as Adam \citep{Kingma15} can alleviate the above issues, avoiding them entirely can be hard.
Since convex programs possess the desirable property that all local minima are global, the existing works have considered convexifying the ANN training problem \citep{Bengio06a, Bach17, arora2018understanding}.
More recently, Pilanci and Ergen proposed ``convex training'' and derived a convex optimization problem with the same global minimum as the non-convex cost function of a one-hidden-layer fully connected ReLU ANN, enabling global ANN optimization \citep{Pilanci20a}.
The favorable properties of convex optimization make convex training immune to back-propagation deficiencies.
Convex training also extends to more complex ANNs such as convolutional neural networks (CNNs) \citep{Ergen21b}, deeper networks \citep{Ergen21a}, and vector-output networks \citep{Sahiner21}.
This work begins with one-hidden-layer ANNs for simplicity, and extends to a family of convex ANN training formulations, including the results for two-hidden-layer sub-networks \cite{Ergen21a, Ergen21c} and one-hidden-layer networks with batch normalization \cite{ErgenBatch}.
Due to space restrictions, the extensions are presented in \Cref{sec:ADMM_ext}.
Moreover, \citep{belilovsky19a} designed a layer-wise training scheme that concatenates one-hidden-layer ANNs into a deep network, where each layer provably reduces the training error.
This approach can be combined with this work, ultimately leading toward training deep networks with convex optimization.

Unfortunately, the $\gO \big( d^3 r^3 (\frac{n}{r})^{3 r} \big)$ computational complexity of the convex training formulation introduced in \citep{Pilanci20a} is exponential in data matrix rank and prohibitively high.
This complexity arises due to the following two reasons:
\begin{itemize}[leftmargin=8mm]
    \setlength\itemsep{.2em}
    \item The size of the convex program grows exponentially in the training data matrix rank $r$.
    This exponential relationship is inherent due to the large number of possible ReLU activation patterns, and thus can be hard to reduce.
    Fortunately, this problem is not a deal-breaker in practice: \cite{Pilanci20a} has shown that a heuristic approximation that forms much smaller convex optimizations performs surprisingly well.
    In this work, we analyze this approximation and theoretically show that for a given level of suboptimality, the required size of the convex training programs is linear in the number of training data points $n$. 
    \item The convex training formulation is constrained.
    A na\"ive algorithm choice for solving a general constrained convex optimization is the interior-point method (IPM) with a cubic per-step computational complexity.
    This paper develops more efficient algorithms that exploit the problem structure and achieve lower computational cost.
    Specifically, an algorithm based on the alternating direction method of multipliers (ADMM) with a quadratic per-iteration complexity, as well as a Sampled Convex Program (SCP)-based algorithm with a linear per-iteration complexity, are introduced.
\end{itemize}

Detailed comparisons among the ADMM-based algorithm, the SCP-based algorithm, the original convex training algorithm in \citep{Pilanci20a}, and back-propagation SGD are presented in \cref{tbl:contributions}.
While IPM can converge to a highly accurate solution with fewer iterations, ADMM can rapidly reach a medium-precision solution, which is often sufficient for machine learning tasks.
Compared with SGD back-propagation, ADMM has a higher theoretical complexity but is guaranteed to converge linearly to a global optimum, enabling efficient global optimization.

\begin{table}
    \caption{
    Comparisons between the proposed ANN training methods and related methods.
    The middle column is the per-epoch complexity when the squared loss is considered.
    $n$ is the number of training points; $d$ is the data dimension; $r$ is the training data matrix rank.
    }
    \label{tbl:contributions}
    \vspace{-1.5mm}
    \footnotesize $\dagger$: Toward the theoretically minimum loss -- further increasing network width will not reduce the training loss; \\
    \footnotesize $\S$: Toward a fixed desired level of suboptimality in the sense defined in \cref{thm:prac}; \\
    \footnotesize $\ddag$: For an arbitrary network width $m$. Since there exists a globally optimal neural network with no more than $n+1$ active hidden-layer neurons \citep{Pilanci20b}, the $\gO(mnd)$ bound for SGD back-propagation evaluates to $\gO (n^2 d)$.
    \vspace{2mm}
    \begin{center}
    \begin{small}
    \begin{tabular}{l|c|c}
        \toprule
        \textbf{Method} & \textbf{Complexity} & \textbf{Global convergence} \\
        \midrule
        \makecell[l]{IPM \citep{Pilanci20a}} & $\gO \big( d^3 r^3 (\frac{n}{r})^{3 r} \big)^\dagger$ & Superlinear to the global optimum. \\ 
        \hline
        ADMM (exact) & $\gO \big( d^2 r^2 (\frac{n}{r})^{2 r} \big)^\dagger$ & \makecell[c]{Rapid to a moderate accuracy; \\ linear to the global optimum.} \\
        \hline
        ADMM (approximate) & $\gO \big( n^2 d^2 \big)^\S$ & \makecell[c]{Rapid to a moderate accuracy; \\ linear to an approximate global optimum.} \\
        \hline
        SCP & $\gO \big( n^2 \big)^\S$ & \makecell[c]{Toward an approximate global optimum; \\ $\gO \big( 1/T \big)$ rate for weakly convex loss; \\ linear for strongly convex loss.} \\
        \hline
        SGD back-propagation & $\gO \big( m n d \big)^\ddag / \gO \big( n^2 d \big)^\dagger$ & \makecell[c]{No spurious valleys if $m \geq 2n+2$; \\ no general results.} \\
        \bottomrule
    \end{tabular}
    \end{small}
    \end{center}
\end{table}

Prior literature has considered applying the ADMM method to ANN training \citep{Taylor16, Wang19}.
These works used ADMM to separate the activations and the weights of each layer, enabling parallel computing.
While the ADMM algorithm in \cite{Wang19} converges at an $\gO(1/t)$ rate ($t$ is the number of iterations) to a critical point of the augmented Lagrangian of the training formulation, there is no guarantee that this critical point is a global optimizer of the ANN training loss.
In contrast, this paper uses ADMM as an efficient convex optimization algorithm and introduces an entirely different splitting scheme based on the convex formulations conceived in \cite{Pilanci20a}.
More importantly, our ADMM algorithm provably converges to a globally optimal classifier.

A parallel line of work has focused on making convex training more efficient.
Specifically, \citep{Ergen21a, Ergen21c} use linear penalty functions to derive unconstrained formulations for convex training.
When the strengths of the penalizations are chosen appropriately, the formulations are exact.
However, the penalization strengths can be difficult to select, since a good choice depends on the optimization landscape of the problem, which is generally unknown.
Note that the solutions found via this penalty method can be used to initialize our ADMM algorithm.
During the review period of this work, Mishkin et al. \citep{Mishkin22} independently proposed a method to accelerate convex training.
The similarities and differences between this work and \citep{Mishkin22} are discussed at the end of \Cref{sec:ADMM}.

Combining the SCP analysis and the convex training framework leads to a further simplified convex training program that solves unconstrained convex optimization problems.
This SCP-based method converges to an approximate global optimum.
The scale of the SCP convex training formulation can be larger than the convex problem solved in the ADMM algorithm.
However, the unconstrained nature enables the use of gradient methods, whose per-iteration complexities are lower than ADMM.
The similarities between the SCP-based algorithm and extreme learning machines (ELMs \citep{Huang04ELM, Gallicchio20}) show that the training of a sparse ELM can be regarded as a convex relaxation of the training of an ANN, providing insights into the hidden sparsity of neural networks.
Due to space restrictions, this result is presented in \Cref{sec:SCP}.

Another major challenge of ANNs is their vulnerability to adversarial attacks.
When the input is perturbed in a carefully designed way that does not significantly alter human perception, ANNs can be tricked into unsafe/incorrect/misaligned outputs drastically different from their normal behaviors.
Such a vulnerability has been observed in computer vision \citep{Szegedy13, Moosavi-Dezfooli16, Goodfellow15} and controls \citep{Huang17}.
As ANNs become popularized in safety-critical applications, it is crucial to analyze their adversarial robustness.
While there have been studies on robustness certification \citep{Anderson20, Ma20, Anderson21a, Bai22b} and achieving certified robustness at test time via ``randomized smoothing'' \citep{Cohen19c, Anderson21b}, efficiently achieving robustness via training remains an important topic.
To this end, ``adversarial training'' \citep{Kurakin17, Goodfellow15, Huang15} is one of the most effective ways to train robust classifiers, compared with other methods such as obfuscated gradients \citep{Athalye18a}.
Specifically, adversarial training replaces the standard loss function with an ``adversarial loss'' and solves a highly challenging bi-level min-max optimization problem.

Adversarial training further exacerbates the aforementioned issues of SGD back-propagation, which arise mostly due to the non-convexity.
As a result, adversarial training can be fragile and volatile in practice, and convergence properties are pessimistic.
Therefore, extending convex training to adversarial training is crucial.
In our conference paper \citep{Bai22a}, we built upon the above results to develop ``convex adversarial training'', explicitly focusing on the cases of hinge loss (for binary classification) and squared loss (for regression).
We theoretically showed that solving the proposed robust convex optimizations trains robust ANNs and empirically demonstrated the efficacy and advantages over traditional methods.
This work extends the analysis to the binary cross-entropy loss and discusses the extensibility to more complex ANN architectures (\Cref{sec:ce_loss} and Appendix \ref{sec:otherANN}).

Previously, researchers have applied convex relaxation techniques to adversarial training.
They obtained convex certifications \citep{raghunathan2018certified, Wong18a} that upper-bounded the inner maximization of adversarial training and used weak duality to develop robust loss functions.
Despite the convex relaxation, the resulting training formulations generally remained non-convex, leaving the fundamental challenges unresolved.
In contrast, we apply robust optimization techniques to the entire min-max adversarial training formulation and obtain convex training problems.

The main contributions of this work are summarized below:
\begin{itemize}[leftmargin=8mm]
	\item A theoretical evaluation of a relaxation that enables tractable convex training (\Cref{sec:prac});
	\item Efficient algorithms to accelerate convex (standard) training (\Cref{sec:ADMM}; Appendix \ref{sec:ADMM_ext});
	\item An extension of the convex adversarial training formulation for one-hidden-layer scalar-output ReLU neural networks (\Cref{sec:convex_adv}).
\end{itemize}

\subsection{Notations} \label{sec:notations}

Throughout this work, we focus on fully connected ANNs with one ReLU-activated hidden layer and a scalar output, defined as
\vspace{-1.5mm}
\begin{equation*}
    \widehat{y} = \sum \jm \big( X u_j + b_j \one_n \big)_+ \alpha_j, \\ \vspace{-1mm}
\end{equation*}
where $X \in \sR^{n \times d}$ is the input data matrix with $n$ data points in $\sR^d$ and $\widehat{y} \in \R^n$ is the ANN output vector.
We use $y \in \R^n$ to denote the corresponding training target output.
The vectors $u_1, \ldots, u_m \in \sR^d$ are the weights of the $m$ hidden layer neurons, the scalars $b_1 \dots, b_m \in \sR$ are the hidden layer bias terms, and the scalars $\alpha_1, \ldots, \alpha_m \in \sR$ represent the output layer weights.
The symbol $(\cdot)_+ = \max \{0, \cdot\}$ indicates the ReLU activation function, which sets all negative entries of a vector or a matrix to zero.
The symbol $\one_n$ defines a column vector with all entries being 1, where the subscript $n$ denotes the dimension of this vector.
The $n$-dimensional identity matrix is denoted by $I_n$.

Furthermore, for a vector $q \in \R^n$, sgn$(q) \in \{-1, 0, 1\}^n$ denotes the signs of the entries of $q$.
$[q \geq 0]$ denotes a boolean vector in $\{0,1\}^n$ with ones at the locations of the non-negative entries of $q$ and zeros at the remaining locations.
The symbol $\diag(q)$ denotes a diagonal matrix $Q \in \R^{n \times n}$ where $Q_{ii} = q_i$ for all $i$ and $Q_{ij} = 0$ for all $i \neq j$.
For a vector $q \in \R^n$ and a scalar $b \in \R$, the inequality $q \geq b$ means that $q_i \geq b$ for all $i\in[n]$.
The symbol $\odot$ denotes the Hadamard product between two vectors and the notation $\norm{\cdot}_p$ denotes the $\ell_p$-norm.
For a matrix $A$, the max norm $\norm{A}_{\max}$ is defined as $\max_{ij} |a_{ij}|$, where $a_{ij}$ is the $(i,j)\th$ entry.  
For a set $\gA$, the notation $|\gA|$ denotes its cardinality, and $\Pi_\gA(\cdot)$ denotes the projection onto $\gA$.
The notation $\prox_f$ denotes the proximal operator associated with a function $f(\cdot)$.
The notation $R\sim\gN(0,I_n)$ indicates that a random variable $R\in\R^n$ is a standard normal random vector, and Unif$(\gS^{n-1})$ denotes the uniform distribution on a $(n-1)$-sphere.
For $P \in \mathbb{N}_+$, we define $[P]$ as the set $\{a \in \mathbb{N}_+ | a \leq P\}$, where $\mathbb{N}_+$ is the set of positive integers.

\section{Practical Convex ANN Training} \label{sec:prac}

\subsection{Prior Work -- Convex ANN Training}

We define the problem of training the above ANN with an $\ell_2$ regularized convex loss function $\ell(\widehat{y},y)$ as:
\begin{equation*} %
    \min_{(u_j, \alpha_j, b_j)\jm} \ell \bigg( \sum\jm \big (X u_j + b_j \one_n \big)_+ \alpha_j, y \bigg) + \frac{\beta}{2} \sum\jm \Big( \norm{u_j}_2^2 + b_j^2 + \alpha_j^2 \Big),
\end{equation*}
where $\beta > 0$ is a regularization parameter.
Without loss of generality, we assume that $b_j = 0$ for all $j\in[m]$.
We can safely make this simplification because concatenating a column of ones to the data matrix $X$ absorbs the bias terms.
The simplified training problem is then:
\begin{equation} \label{eq:nonconvex_general}
    \min_{(u_j, \alpha_j)\jm} \ell \bigg( \sum\jm (X u_j)_+ \alpha_j, y \bigg) + \frac{\beta}{2} \sum\jm \big( \norm{u_j}_2^2 + \alpha_j^2 \big).
\end{equation}

Consider a set of diagonal matrices $\{ \diag([Xu\geq 0]) | u\in\R^d \}$, and let the distinct elements of this set be denoted as $D_1, \dots, D_P$.
The constant $P$ corresponds to the total number of partitions of $\R^d$ by hyperplanes passing through the origin that are also perpendicular to the rows of $X$ \citep{Pilanci20a}. 
Intuitively, $P$ can be regarded as the number of possible ReLU activation patterns associated with $X$.

Consider the convex optimization problem
\begin{align} \label{eq:convex_general}
    \min_{(v_i, w_i)\iP} & \ell \bigg( \sum\iP D_i X (v_i-w_i), y \bigg) + \beta \sum\iP \Big( \norm{v_i}_2 + \norm{w_i}_2 \Big) \\
    \ST \quad & (2D_i-I_n) X v_i\geq 0, \; (2D_i-I_n) X w_i\geq 0, \quad \alliinP \nonumber
\end{align}
and its dual formulation
\begin{align} \label{eq:dual}
    & \max_v -\ell^*(v) \qquad \ST \;\; | v^\top (Xu)_+ | \leq \beta, \;\; \forall u: \norm{u}_2 \leq 1,
\end{align}
which is a convex semi-infinite program, where $\ell^*(v) = \max_z z^\top v - \ell(z,y)$ is the Fenchel conjugate function.
The next theorem, borrowed from Pilanci and Ergen's paper \citep{Pilanci20a}, explains the relationship between the non-convex training problem \cref{eq:nonconvex_general}, the convex problem \cref{eq:convex_general}, and the dual problem \cref{eq:dual} when the ANN is sufficiently wide. 

\begin{theorem}[\citep{Pilanci20a}] \label{THM:PILANCI}
    Let $(v_i^\star,w_i^\star)\iP$ denote a solution of \cref{eq:convex_general} and define $m^\star$ as $|\{i : v_i^\star \neq 0\}| + |\{i : w_i^\star \neq 0\}|$.
    Suppose that the ANN width $m$ is at least $m^\star$, where $m^\star$ is upper-bounded by $n+1$.
    If the loss function $\ell (\cdot, y)$ is convex, then \cref{eq:nonconvex_general}, \cref{eq:convex_general}, and \cref{eq:dual} share the same optimal objective.
    The optimal network weights $(u_j^\star, \alpha_j^\star)\jm$ can be recovered using the formulas
    \begin{equation}\label{eq:recover_weights} 
        \begin{aligned}
            (u_{j_{1 i}}^\star, \alpha_{j_{1 i}}^\star) & = \Big( \dfrac{v_i^\star}{\sqrt{\norm{v_i^\star}_2}}, \sqrt{\norm{v_i^\star}_2} \Big) \hspace{8mm} \text{if $v_i^\star \neq 0$}; \\
            (u_{j_{2 i}}^\star, \alpha_{j_{2 i}}^\star) & = \Big( \dfrac{w_i^\star}{\sqrt{\norm{w_i^\star}_2}}, -\sqrt{\norm{w_i^\star}_2} \Big) \quad \text{if $w_i^\star \neq 0$}.
        \end{aligned}
    \end{equation}
    where the remaining $m-m^\star$ neurons are chosen to have zero weights.
\end{theorem}

The worst-case computational complexity of solving \cref{eq:convex_general} for the case of squared loss is $\gO \big( d^3 r^3 (\frac{n}{r})^{3 r} \big)$ using standard interior-point solvers \citep{Pilanci20a}.
Here, $r$ is the rank of the data matrix $X$, and in many cases $r=d$. 
Such a complexity is polynomial in $n$, significantly better than previous methods, but is exponential in $r$, thus still prohibitively high for many practical applications.
Such high complexity is due to the large number of $D_i$ matrices, which is upper-bounded by $\min \big\{ 2^n, 2r \big( \frac{e(n-1)}{r} \big)^r \big\}$ \citep{Pilanci20a}.

\subsection{A Practical Convex Training Algorithm}

\begin{algorithm}[t]
\begin{algorithmic}[1]
    \State Generate $P_s$ distinct diagonal matrices via $D_h \gets \diag ([X a_h \geq 0])$, where $a_h \sim \gN (0, I_d)$ i.i.d. for all $h \in [P_s]$.
    \State \text{Solve} \vspace{-5mm}
    \begin{align} \label{eq:prac_clean}
        p_{s1}^\star = \min_{(v_h, w_h)\iPs} & \ell \Big( \sum\iPs D_h X (v_h-w_h), y \Big) + \beta \sum\iPs \big( \norm{v_h}_2 + \norm{w_h}_2 \big) \\
        \ST \quad & (2D_h-I_n) X v_h\geq 0, \; (2D_h-I_n) X w_h\geq 0, \quad \alliinPs. \nonumber
    \end{align}; \vspace{-4mm}
    \State \text{Recover $u_1, \ldots, u_{m_s}$ and $\alpha_1, \ldots, \alpha_{m_s}$ from the solution} \text{$(v_{s_h}^\star, w_{s_h}^\star)_{h=1}^{P_s}$ of \cref{eq:prac_clean} using \cref{eq:recover_weights}}.
\end{algorithmic}
\caption{Practical convex training}
\label{alg:train}
\end{algorithm} 

A natural direction of mitigating this high complexity is to reduce the number of $D_i$ matrices by sampling a subset of them.
This idea leads to \cref{alg:train}, which approximately solves the training problem and can train ANNs with widths much less than $m^\star$.
\cref{alg:train} is an instance of the approximation described in \cite[Remark 3.3]{Pilanci20a}, but \citep{Pilanci20a} did not provide theoretical insights regarding its level of suboptimality.
The following theorem bridges the gap by providing a probabilistic bound on the suboptimality of the ANN trained with \cref{alg:train}.

\begin{theorem} \label{thm:prac}
    Consider an additional diagonal matrix $D_{P_s + 1}$ sampled uniformly, and construct
    \begin{align} \label{eq:prac_clean2}
        p_{s2}^\star = \min_{(v_h, w_h)\iPss} & \ell \Big( \sum\iPss D_h X (v_h-w_h), y \Big) + \beta \sum\iPss \big( \norm{v_h}_2 + \norm{w_h}_2 \big) \\
        \ST \quad & (2D_h-I_n) X v_h\geq 0, \; (2D_h-I_n) X w_h\geq 0, \;\; \alliinPss. \nonumber
        \vspace{-3.5mm}
    \end{align}
    It holds that $p_{s2}^\star \leq p_{s1}^\star$.
    Furthermore, if $P_s \geq \min \big\{ \frac{n+1}{\psi \xi} - 1, \frac{2}{\xi} ( n+1-\log\psi ) \big\}$, where $\psi$ and $\xi$ are preset confidence level constants between 0 and 1, then with probability at least $1-\xi$, it holds that $\sP \{ p_{s2}^\star < p_{s1}^\star \} \leq \psi$.
\end{theorem}

The proof of \cref{thm:prac} is presented in \Cref{sec:pracproof}.
Intuitively, \cref{thm:prac} shows that sampling an additional $D_{P_s + 1}$ matrix will not reduce the training loss with high probability when $P_s$ is large.
One can recursively apply this bound $T$ times to show that the solution with $P_s$ matrices is close to the solution with $P_s+T$ matrices for an arbitrary number $T$.
Thus, while the theorem does not directly bound the gap between the approximated optimization problem and its exact counterpart, it states that the optimality gap due to sampling is not too large for a suitable value of $P_s$, and the trained ANN is nearly optimal.

Compared with the exponential relationship between $P$ and $r$, a satisfactory value of $P_s$ is linear in $n$ and is independent from $r$.
Thus, when $r$ is large, solving the approximated formulation \cref{eq:prac_clean} is significantly (exponentially) more efficient than solving the exact formulation \cref{eq:convex_general}.
On the other hand, \cref{alg:train} is no longer deterministic due to the stochastic sampling of the $D_h$ matrices, and yields solutions that upper-bound those of \cref{eq:convex_general}.

Since the confidence constants $\psi$ and $\xi$ are no greater than one, \cref{thm:prac} only applies to overparameterized ANNs, where $P_s \geq n$.
Although \citep{Pilanci20a} has shown that there exists a globally optimal neural network whose width is at most $n+1$ and \cref{thm:prac} seems loose by this comparison, our theorem bounds a different quantity and is meaningful.
Specifically, the bound in \citep{Pilanci20a} does not provide a method that scales linearly: while a globally optimal neural network narrower than $n+1$ exists, finding such an ANN requires solving a convex program with an exponential number of constraints.
In contrast, \cref{thm:prac} characterizes the optimality of a convex optimization with a manageable number of constraints.
In practice, selecting $P_s$ is equivalent to choosing the ANN width.
While \cref{thm:prac} provides a guideline on how $P_s$ should scale with $n$, selecting a much smaller $P_s$ will not necessarily become an issue.
Our experiments in \Cref{sec:prac_exp} show that even when $P_s$ is much less than $n$ (which is much less than $P$), \cref{alg:train} still reliably trains high-performance classifiers.

\section{An ADMM Algorithm for Global ANN Training} \label{sec:ADMM}

The convex ReLU ANN training program \cref{eq:convex_general} may be solved with the IPM.
The IPM is an iterative algorithm that repeatedly performs Newton updates.
Each Newton update requires solving a linear system, which has a cubic complexity, hindering the application of IPM to large-scale optimization problems.
Unfortunately, large-scale problems are ubiquitous in the field of machine learning.
This section proposes an algorithm based on the ADMM, breaking down the optimization problem \cref{eq:convex_general} to smaller subproblems that are easier to solve.
Moreover, when $\ell(\cdot)$ is the squared loss, each subproblem has a closed-form solution.
We will show that the complexity of each ADMM iteration is linear in $n$ and quadratic in $d$ and $P$, and the number of required ADMM steps to reach a desired precision is logarithmic in the precision level.
When other convex loss functions are used, a closed-form solution may not always exist.
We illustrate that iterative methods can solve the subproblems for general convex losses efficiently.
In \Cref{sec:ADMM_ext}, we show that the ADMM algorithm extends to a family of convex training formulations.

Define $F_i \coloneqq D_i X$ and $G_i \coloneqq (2 D_i - I_n) X$ for all $i \in [P]$.
Furthermore, we introduce $v_i$, $w_i$, $s_i$, and $t_i$ as slack variables and let $v_i = u_i$, $w_i = z_i$, $s_i = G_i v_i$, and $t_i = G_i w_i$.
For a vector $q = (q_1, \dots, q_n) \in \R^n$, define the indicator function of the positive quadrant $\sI_{\geq0}$ as
\begin{align*}
    \sI_{\geq0} (q) \coloneqq \begin{cases} 0 & \text{if } q_i \geq 0,\ \forall i\in[N]; \\ +\infty & \text{otherwise.} \end{cases}
\end{align*}

The convex training formulation \cref{eq:convex_general} can be reformulated as a convex optimization problem with positive quadrant indicator functions and linear equality constraints:
\begin{align} \label{eq:3}
    \min_{(v_i,w_i,s_i,t_i,u_i,z_i)\iP} & \ell \Big( \sum\iP F_i (u_i-z_i), y \Big) + \beta \sum\iP \norm{v_i}_2 + \beta \sum\iP \norm{w_i}_2 + \sum\iP \sI_{\geq0} (s_i) + \sum\iP \sI_{\geq0}(t_i) \nonumber \\
    \ST \qquad\ & G_i u_i - s_i = 0, \quad G_i z_i - t_i = 0, \quad v_i - u_i = 0, \quad w_i - z_i = 0, \quad\ \forall i\in[P].
\end{align}

Next, we simplify the notations by concatenating the matrices. 
Define 
\begin{gather*}
    u \coloneqq [u_1^\top \ \cdots \ u_P^\top \ \ z_1^\top \ \cdots \ z_P^\top]^\top, \quad
    v \coloneqq [v_1^\top \ \cdots \ v_P^\top \ \ w_1^\top \ \cdots \ w_P^\top]^\top, \quad \\
    s \coloneqq [s_1^\top \ \cdots \ s_P^\top \ \ t_1^\top \ \cdots \ t_P^\top]^\top, \\
    F \coloneqq [F_1 \ \cdots \ F_P \ \ -F_1 \ \cdots \ -F_P],\ \text{ and } \
    G \coloneqq \mathrm{blkdiag}(G_1, \cdots, G_P, G_1, \cdots, G_P),
\end{gather*}
where blkdiag$(\cdot,\dots,\cdot)$ denotes the block diagonal matrix formed by the submatrices in the parentheses.
The formulation \cref{eq:3} is then equivalent to the compact notation
\begin{align} \label{eq:4}
    \min_{v,s,u} \ell (F u, y) + \beta \norm{v}_{2,1} + \sI_{\geq0}(s) \quad \ST \quad \begin{bmatrix} I_{2dP} \\ G \end{bmatrix} u - \begin{bmatrix} v \\ s \end{bmatrix} = 0,
\end{align}
where $\norm{\cdot}_{2,1}$ denotes the $\ell_1$-$\ell_2$ mixed norm group sparse regularization and $I_{2dP}$ is the idendity matrix in $\sR^{2dP \times 2dP}$.
The corresponding augmented Lagrangian of \cref{eq:4} is:

\begin{align*}
	\\[-12mm]
    L (u, v, & s, \nu, \lambda) := \\[-.5mm]
    & \ell \big( F u, y \big) + \beta \bignorm{v}_{2,1} + \sI_{\geq0} \big( s \big) + \frac{\rho}{2} \Big( \norm{u - v + \lambda}_2^2 - \norm{\lambda}_2^2 \Big) + \frac{\rho}{2} \Big( \norm{G u - s + \nu}_2^2 - \norm{\nu}_2^2 \Big),
\end{align*}
where $\lambda \coloneqq [\lambda_{11} \ \cdots \ \lambda_{1P} \ \ \lambda_{21} \ \cdots \ \lambda_{2P}]^\top \in \R^{2dP}$ and $\nu \coloneqq [\nu_{11} \ \cdots \ \nu_{1P} \ \ \nu_{21} \ \cdots \ \nu_{2P}]^\top \in \R^{2nP}$ are dual variables, $\rho > 0$ is a fixed penalty parameter \citep{Hestenes1969}.

\begin{algorithm}[t]
    \begin{algorithmic}[1] \normalsize
        \MRepeat \vspace{1.5mm}
        \State Primal update \vspace{-3.4mm}
        \refstepcounter{equation} \label{eq:ADMM}
        \begin{align} \label{eq:ADMM1}
            \hspace{8.5mm} u^{k+1} = & \argmin_u \ell (F u, y) + \frac{\rho}{2} \norm{u - v^k + \lambda^k}_2^2 + \frac{\rho}{2} \norm{G u - s^k + \nu^k}_2^2 \hspace{10.2mm} \tag{3.3a}
        \end{align} \vspace{-5mm}
        
        \State Primal update \vspace{-3.1mm}
        \begin{align} \label{eq:ADMM2}
            \hspace{1mm} \begin{bmatrix} v^{k+1} \\ s^{k+1} \end{bmatrix} = & 
            \argmin_{v,s} \beta \norm{v}_{2,1} + \sI_{\geq0}(s) + \frac{\rho}{2} \norm{u^{k+1} - v + \lambda^k}_2^2 + \frac{\rho}{2} \norm{G u^{k+1} - s + \nu^k}_2^2 \tag{3.3b}
        \end{align} \vspace{-5mm}
        
        \State Dual update: \vspace{-5.5mm}
        \begin{align} \label{eq:ADMM3}
            \begin{bmatrix} \lambda^{k+1} \\ \nu^{k+1} \end{bmatrix} = &
            \begin{bmatrix}
                \lambda^k + \frac{\gamma_a}{\rho} (u^{k+1} - v^{k+1}) \\ \nu^k + \frac{\gamma_a}{\rho} (G u^{k+1} - s^{k+1})
            \end{bmatrix} \hspace{2.5mm} \tag{3.3c}
        \end{align} \vspace{-5.5mm}
        \EndRepeat
    \end{algorithmic}
    \caption{An ADMM algorithm for the convex ANN training problem.}
    \label{alg:ADMM}
\end{algorithm}

We can apply the ADMM iterations described in \cref{alg:ADMM} to globally optimize \cref{eq:4}.\footnote{
The ADMM algorithm is presented in the scaled dual form \citep{Boyd11}.
}
Here, $\gamma_a > 0$ is a step-size constant.
As will be shown next, \cref{eq:ADMM2} and \cref{eq:ADMM3} have simple closed-form solutions.
The update \cref{eq:ADMM1} has a closed-form solution when $\ell (\cdot)$ is the squared loss, and can be efficiently solved numerically for general convex loss functions.
When we apply ADMM to solve the approximated convex training formulation \cref{eq:prac_clean}, \cref{alg:ADMM} becomes a subalgorithm of \cref{alg:train}.
The following theorem certifies the linear convergence of the ADMM algorithm, with the proof provided in \Cref{sec:ADMMproof}:
\begin{theorem} \label{thm:ADMM}
    If $\ell(\yhat, y)$ is strictly convex and continuously differentiable with a uniform Lipschitz continuous gradient with respect to $\yhat$, then the sequence $\{(u^k, v^k, s^k, \lambda^k, \nu^k)\}$ generated by \cref{alg:ADMM} converges linearly to an optimal primal-dual solution for \cref{eq:4}, provided that the step size $\gamma_a$ is sufficiently small.
\end{theorem}

Many popular loss functions satisfy the conditions of \cref{thm:ADMM}.
Examples include the squared loss (for regression) and the binary cross-entropy loss coupled with the tanh or the sigmoid output activation (for binary classification).

\subsection{\texorpdfstring{$s$}{s} and \texorpdfstring{$v$}{v} Updates} \label{sec:svupdates}

\begingroup
\allowdisplaybreaks
The update step \cref{eq:ADMM2} can be separated for $v^{k+1}$ and $s^{k+1}$ as:
\refstepcounter{equation} \label{eq:ADMMb}
\begin{align}
    v^{k+1} = & \argmin_v \beta \norm{v}_{2,1} + \frac{\rho}{2} \norm{u^{k+1} - v + \lambda^k}_2^2; \tag{3.4a} \label{eq:vupdate} \\
    s^{k+1} = & \argmin_s \sI_{\geq0}(s) + \norm{G u^{k+1} - s + \nu^k}_2^2 = \argmin_{s\geq0} \norm{G u^{k+1} - s + \nu^k}_2^2. \tag{3.4b} \label{eq:supdate}
\end{align}

Note that \cref{eq:vupdate} can be separated for each $v_i$ and $w_i$ (allowing parallelization) and solved analytically using the formulas
\begin{align*}
    v_i^{k+1} = & \argmin_v \beta \norm{v_i}_{2} + \frac{\rho}{2} \norm{u_i^{k+1} - v + \lambda_{1i}^k}_2^2 = \prox_{\frac{\beta}{\rho} \norm{\cdot}_2} \big( u_i^{k+1} + \lambda_{1i}^k \big) \\[-1.5mm]
    = & \bigg( 1 - \frac{\beta}{\rho \cdot \bignorm{u_i^{k+1} + \lambda_{1i}^k}_2} \bigg)_+ \big( u_i^{k+1} + \lambda_{1i}^k \big), \qquad\quad \forall i\in[P], \\[2mm]
    w_i^{k+1} = & \argmin_v \beta \norm{w_i}_{2} + \frac{\rho}{2} \norm{s_i^{k+1} - w + \lambda_{2i}^k}_2^2 = \prox_{\frac{\beta}{\rho} \norm{\cdot}_2} \big( z_i^{k+1} + \lambda_{2i}^k \big) \\[-1.5mm]
    = & \bigg( 1 - \frac{\beta}{\rho \cdot \bignorm{z_i^{k+1} + \lambda_{2i}^k}_2} \bigg)_+ \big( z_i^{k+1} + \lambda_{2i}^k \big), \qquad\quad \forall i \in [P],
\end{align*}
\endgroup
where $\prox_{\frac{\beta}{\rho} \norm{\cdot}_2}$ denotes the proximal operation on the function $f(\cdot) = \frac{\beta}{\rho} \norm{\cdot}_2$.
The computational complexity of finding $v_i$ and $w_i$ is $\gO (d)$.
Similarly, \cref{eq:supdate} can also be separated for each $s_i$ and $t_i$ and solved analytically using the formulas
\begin{align*}
    s_i^{k+1} = & \argmin_{s_i\geq0} \bignorm{G_i u_i^{k+1} - s_i + \nu_{1i}^k}_2^2 = \Pi_{\geq0} \big( G_i u_i^{k+1} + \nu_{1i}^k \big) = \big( G_i u_i^{k+1} + \nu_{1i}^k \big)_+, \quad \forall i\in[P]; \\
    t_i^{k+1} = & \argmin_{t_i\geq0} \bignorm{G_i z_i^{k+1} - s_i + \nu_{2i}^k}_2^2 = \Pi_{\geq0} \big( G_i z_i^{k+1} + \nu_{2i}^k \big) = \big( G_i z_i^{k+1} + \nu_{2i}^k \big)_+, \quad\ \forall i\in[P].
\end{align*}
where $\Pi_{\geq0}$ denotes the projection onto the non-negative quadrant.
The computational complexity of finding $s_i$ and $t_i$ is $\gO (n)$.
The updates \cref{eq:vupdate} and \cref{eq:supdate} can be performed in $\gO(nP+dP)$ time in total.

\subsection{\texorpdfstring{$u$}{u} Updates}

The $u$ update step depends on the specific structure of $\ell (\cdot)$.
For the squared loss, the $u$ update step can be solved in closed form.
For many other loss functions, the update can be performed with numerical methods.

\subsubsection{Squared Loss} \label{sec:ADMM_sql}

The squared loss $\ell (\widehat{y}, y) = \frac{1}{2} \enorm{\widehat{y} - y}^2$ is a commonly used loss function in machine learning.
It is widely used for regression tasks, but can also be used for classification.
For the squared loss, \cref{eq:ADMM1} amounts to
\begin{equation} \label{eq:ADMM1_sql}
    u^{k+1} = \argmin_u \Big\{ \enorm{F u - y}^2 + \frac{\rho}{2} \norm{u - v^k + \lambda^k}_2^2 + \frac{\rho}{2} \norm{G u - s^k + \nu^k}_2^2 \Big\}.
\end{equation}

Setting the gradient with respect to $u$ to zero yields that 
\begin{align} \label{eq:uupdate}
    \big( I + \tfrac{1}{\rho} F^\top F + G^\top G \big) u^{k+1} = \tfrac{1}{\rho} F^\top y + v^k - \lambda^k + G^\top s^k - G^\top \nu^k.
\end{align}

Therefore, the $u$ update can be performed by solving the linear system \cref{eq:uupdate} in each iteration.
While solving a linear system $Ax=b$ for a square matrix $A$ has a cubic time complexity in general, by taking advantage of the structure of \cref{eq:uupdate}, a quadratic per-iteration complexity can be achieved.
Specifically, the matrix $I + \tfrac{1}{\rho} F^\top F + G^\top G$ is symmetric, positive definite, and fixed throughout the ADMM iterations.
In general, we can solve $A x = b$ for some symmetric $A \in \sS^{2dP \times 2dP}$, $A \succ 0$ and $b \in \sR^{2dP}$ via the procedure:
\begin{enumerate}[leftmargin=8mm]
    \setlength\itemsep{.2em}
    \item Perform Cholesky decomposition $A = L L^\top$, where $L$ is lower-triangular (cubic complexity in $2dP$);
    \item Solve $L \widehat{b} = b$ by forward substitution (quadratic complexity in $2dP$);
    \item Solve $L^\top x = \widehat{b}$ by back substitution (quadratic complexity in $2dP$).
\end{enumerate}

Throughout the ADMM iterations, the first step only needs to be performed once, while the second and third steps are required for every iteration.
Since the dimension of the matrix $(I + \tfrac{1}{\rho} F^\top F + G^\top G)$ is $2dP \times 2dP$, the per-iteration time complexity of the $u$ update is $\gO(d^2 P^2)$, making it the most time-consuming step of our algorithm when $d$ and $P$ are large.
Thus, the overall complexity of a full ADMM primal-dual iteration for squared loss is $\gO( nP + d^2 P^2 )$, which is quadratic.
In contrast, the linear system for IPM's Newton updates can be different for each iteration, and thus each iteration has a cubic complexity.
Hence, the proposed ADMM method achieves a notable speed improvement over IPM.

When the approximated formulation \cref{eq:prac_clean} is considered and $P_s$ diagonal matrices are sampled in place of the full set of $P$ matrices, obtaining a given level of optimality requires $P_s$ to be linear in $n$, as discussed in \Cref{sec:prac}.
Coupling with the above analysis, we obtain an overall $\gO(d^2 n^2)$ per-iteration complexity, a significant improvement over the $\gO \big( d^3 r^3 (\frac{n}{r})^{3r} \big)$ per-iteration complexity in \citep{Pilanci20a}.
The total computational complexity for reaching a point $u^k$ satisfying $\enorm{u^k - u^\star} \leq \epsilon_a$ is $\gO(d^2 n^2 \log(1/\epsilon_a))$, where $u^\star$ is an optimal value of $u$ and $\epsilon_a > 0$ is a predefined precision threshold.
In \Cref{sec:ADMMexp}, we use numerical experiments to demonstrate that the ADMM algorithm's high efficiency enables convex ANN training for image classification tasks for the first time.
Moreover, our experiments show that a high prediction accuracy only requires moderate optimization precision, which can be reached within a few ADMM iterations.

\subsubsection{General Convex Loss Functions}

When a general convex loss function $\ell(\widehat{y},y)$ is considered, a closed-form solution to \cref{eq:ADMM1} does not always exist, and one may need to use iterative methods, such as gradiant descent, to solve \cref{eq:ADMM1}.
However, for large-scale problems, a full gradient evaluation is prohibitively expensive.
To address this issue, we exploit the symmetric and separable property of each $u_i$ and $z_i$ in \cref{eq:ADMM1} and propose a randomized block coordinate descent (RBCD) method in \cref{alg:BCD}.
Steps 5 and 6 of \cref{alg:BCD} are derived via the differentiation chain rule.
Note that \cref{eq:ADMM1} is always strongly convex because its second term is strongly convex while the first and third terms are convex.
Hence, our RBCD algorithm converges linearly \citep[Theorem 1]{DBLP:journals/mp/LuX15}.
The theoretical convergence rate is faster when the convexity of \cref{eq:ADMM1} is stronger and $P$ is smaller.

\begin{algorithm}{}
\begin{algorithmic}[1]
    \State Initialize $\yhat = \sum\iP F_i (u_i - z_i)$;
    \State Fix $\stil_i = G_i^\top (s_i - \nu_{1i})$, $\ttil_i = G_i^\top (t_i - \nu_{2i})$ for all $i\in[P]$;
    \State Select accuracy thresholds $\tau>0, \varphi>0$;
    \Repeat
        \State $\ytil \leftarrow \nabla_{\yhat} \ell(\yhat,y)$
        \State Uniformly select $i$ from $[P]$ at random;
        \State $u_i^+ \leftarrow u_i - \gamma_r F_i^\top \ytil - \gamma_r \rho (u_i - v_i + \lambda_{1i} + G_i^\top G_i u_i - \stil_i )$;
        \State $z_i^+ \leftarrow z_i + \gamma_r F_i^\top \ytil - \gamma_r \rho (z_i - w_i + \lambda_{2i} + G_i^\top G_i z_i - \ttil_i )$;
        \State $\yhat^+ \leftarrow \yhat + F_i \big( (u_i^+ - z_i^+) - (u_i + z_i) \big)$;
    \Until $\enorm{\nabla_u L(u,v,s,\nu,\lambda)} \leq \dfrac{\varphi}{\max\{\tau, \enorm{u}\}}$.
\end{algorithmic}
\caption{
Randomized Block Coordinate Descent (RBCD). \\
\small The superscript $^+$ denotes the updated quantities for each iteration; $\gamma_r$ denotes the step size.
}
\label{alg:BCD}
\end{algorithm}

In practice, the RBCD step size $\gamma_r$ can be adaptively chosen via the backtracking line search.
While \cref{alg:BCD} updates one block in each iteration, it is also possible to update multiple blocks at once by sampling multiple indices.
Moreover, each iteration can use the gradient associated with a random portion of the dataset as a surrogate for the entire dataset.

Furthermore, it holds that $G_i^\top G_i = X^\top X$ for all $i \in [P]$.
To understand this, recall that $G_i = (2D_i - I_n) X$ by definition.
Since $(2D_i - I_n)$ is a diagonal matrix with all entries being $\pm 1$, it holds that $(2D_i - I_n)^\top (2D_i - I_n) = I_n$, and thus $G_i^\top G_i = X^\top (2D_i - I_n)^\top (2D_i - I_n) X = X^\top X$.
Therefore, we pre-compute $X^\top X$, removing the need to compute $G_i^\top G_i$ in each iteration.
The most expensive steps of each RBCD update thus have the following complexities:
\begin{table}[H]
	\vspace{-2mm}
    \centering
    \begin{small}
    \begin{tabular}{c|c|c|c}
        $F_i^\top \ytil$ & $F_i \Big( (u_i^+ - z_i^+) - (u_i + z_i) \Big)$ & $(X^\top X) u_i$ & $(X^\top X) z_i$ \\
        \midrule
        $\gO(nd)$ & $\gO(nd)$ & $\gO(d^2)$ & $\gO(d^2)$
    \end{tabular}
    \end{small}
    \vspace{-2.5mm}
\end{table}

While it can be costly to solve \cref{eq:ADMM1} to a high accuracy using iterative methods, especially during the early iterations of ADMM, \citep[Proposition 6]{Eckstein17} has shown that even when \cref{eq:ADMM1} is solved approximately, as long as the accuracy threshold $\varphi$ of each ADMM iteration forms a convergent sequence, the ADMM algorithm can eventually converge to the global optimum of \cref{eq:4}.
Each iterative solution of the $u$-update subproblem can also take advantage of warm-starting by initializing at the result of the previous ADMM iteration.
As a result, we alternate between an ADMM update and several RBCD updates in a delicate manner.

Compared to the parallel independent work \citep{Mishkin22}, our method sees some connections but is overall distinct.
The authors of \citep{Mishkin22} considered two approaches, one using an unconstrained relaxation to the constrained convex training formulation and the other directly tackling the constrained formulation.
While \citep{Mishkin22} also proposes to reformulate the constraints into an augmented Lagrangian, it uses a separation scheme different from ours.
Specifically, we separate the group-sparse regularization in addition to the constraints, whereas \citep{Mishkin22} only separates the constraints.
As a result, our ADMM separation allows the primal update subproblem \cref{eq:ADMM1} to be solved in closed form for the case of squared loss, whereas \citep{Mishkin22} requires the FISTA algorithm for the primal update step.
For general loss functions, our separation embeds strong convexity into the subproblem \cref{eq:ADMM1}, allowing the randomized block coordinate descent (RBCD) subroutine to converge linearly.
Furthermore, our ADMM algorithm also achieves linear convergence, whereas \citep{Mishkin22} claims a slower $\gO (\frac{1}{\epsilon\delta})$ dual convergence rate.

\section{Convex Adversarial Training} \label{sec:convex_adv}

The inherent difficulties with adversarial training can be addressed by taking advantage of the convex training framework and the related algorithms.

\subsection{Adversarial Training Background} \label{adv_train_section}

A classifier is considered robust against adversarial perturbations if it assigns the same class to all inputs within a perturbation set.
We need the perturbation set to define the input distortion allowances, because an unlimited distortion breaks even the most robust models, and is impractical because it can be easily detected and rejected.
We consider a $\ell_\infty$-bounded perturbation set with radius $\epsilon > 0$, a common problem formulation proposed in \citep{Goodfellow15}:
\begin{align} \label{perturbation_set}
    \gX = \Big\{ & X + \Delta \in \sR^{n \times d} \ \Big| \ 
    \Delta = [\delta_1, \ldots, \delta_n]^\top,\ \delta_k\in\R^d,\ \norm{\delta_k}_\infty \leq \epsilon,\ \forall k \in [n] \Big\}. \nonumber
\end{align}

We consider the ``white box'' setting, where the adversary has complete knowledge about the ANN.
A common method for training robust classifiers is to minimize the loss associated with the worst-case perturbation, i.e., the attack resulting in the maximum loss within the perturbation set.
More concretely, we solve the following min-max problem proposed in \citep{madry2018towards}:
\begin{equation} \label{eq:robust_general}
\begin{aligned}
        \min_{(u_j, \alpha_j)\jm} 
        \begin{pmatrix}
            \displaystyle \max\allDelta \ell \bigg( \sum_{j=1}^m \big( (X+\Delta) u_j \big)_+ \alpha_j, y \bigg) + \frac{\beta}{2} \sum_{j=1}^m \big( \norm{u_j}_2^2 + \alpha_j^2 \big) \hfill
        \end{pmatrix}.
    \end{aligned}
\end{equation}

This process of ``training with adversarial data'' is often referred to as ``adversarial training'', as opposed to ``standard training'' that trains on clean, unperturbed data.
In the prior literature, Fast Gradient Sign Method (FGSM) and Projected Gradient Descent (PGD) were commonly used to numerically solve the inner maximization of \cref{eq:robust_general} and generate adversarial examples \citep{madry2018towards}.
Specifically, PGD generates adversarial examples $\tilde{x}$ by running the iterations
\begin{equation}\label{eq:PGD}
    \tilde{x}^{t+1} = \Pi_{\gX} \bigg( \tilde{x}^t + \gamma_p \cdot \sgn \Big( \nabla_x \ell \big( \sum\jm (x^\top u_j)_+ \alpha_j, y \big) \Big) \bigg)
\end{equation}
for $t = 0, 1, \cdots, T$, where $\tilde{x}^t$ is the perturbed data vector at the $t^{\th}$ iteration, $\gamma_p > 0$ is the step size, and $T \geq 1$ is the number of iterations.
The initial vector $\tilde{x}^0$ is the unperturbed data $x$. FGSM can be regarded as a special case of PGD where $T = 1$.

\subsection{The Convex Adversarial Training Formulation}

While adversarial training with PGD adversaries has demonstrated some success, this approach suffers from several limitations.
Since the optimization landscapes are generally non-concave over the perturbation $\Delta$, there is no guarantee that PGD will find the true worst-case adversary.
Furthermore, traditional adversarial training solves complicated bi-level min-max optimization problems, exacerbating the instability of non-convex ANN training.
Our experiments show that back-propagation gradient methods can struggle to converge when solving \cref{eq:robust_general}.
Moreover, solving the bi-level optimization \cref{eq:robust_general} requires an algorithm with a computationally cumbersome nested loop structure.
To conquer such difficulties, we leverage \cref{THM:PILANCI} to re-characterize \cref{eq:robust_general} as robust, convex upper-bound problems that can be efficiently solved globally.

We first develop a result about adversarial training involving general convex loss functions.
The connection between the convex training objective and the non-convex ANN loss function holds only when the linear constraints in \cref{eq:convex_general} are satisfied.
For adversarial training, we need this connection to hold at all perturbed data matrices $X+\Delta\in\gX$.
Otherwise, if some matrix $X+\Delta$ violates the linear constraints, then this perturbation $\Delta$ can correspond to a low convex objective value but a high actual loss.
To ensure the correctness of the convex reformulation throughout $\gX$, we introduce some robust constraints below.
 
Since the $D_i$ matrices in \cref{eq:convex_general} reflect the ReLU patterns of $X$, these matrices can change when $X$ is perturbed.
Therefore, we include all distinct diagonal matrices $\diag([(X+\Delta) u \geq 0])$ that can be obtained for all $u \in \R^d$ and \textit{all} $\Delta: X+\Delta \in \gU$, denoted as $D_1$, $\dots$, $D_{\widehat{P}}$, where $\widehat{P}$ is the total number of such matrices.
Since $D_1$, $\dots$, $D_{\widehat{P}}$ include $D_1$, $\dots$, $D_P$ in \cref{eq:convex_general}, we have $\widehat{P} \geq P$.
While $\widehat{P}$ is at most $2^n$ in the worst case, since $\epsilon$ is often small, we expect $\widehat{P}$ to be relatively close to $P$, where $P \leq 2r \big( \frac{e(n-1)}{r} \big)^r$ as discussed above.

Finally, we replace the objective of the convex standard training formulation \cref{eq:convex_general} with its robust counterpart, giving rise to the optimization problem
\refstepcounter{equation} \label{eq:rob_gen_cvx}
\begin{align}
    & \min_{(v_i, w_i)\iPhat} 
    \begin{pmatrix}
        \displaystyle \max\allDeltaU \ell \bigg( \sum\iPhat D_i (X+\Delta) (v_i - w_i), y \bigg) + \ \beta \sum\iPhat \big( \norm{v_i}_2 + \norm{w_i}_2 \big) \hfill
    \end{pmatrix} \tag{5.4a} \label{eq:rob_gen_cvx_1} \\
    \ST \; & \min\allDeltaU (2 D_i - I_n) (X+\Delta) v_i \geq 0, 
    \; \min\allDeltaU (2 D_i - I_n) (X+\Delta) w_i \geq 0, \;\; \alliinPhat, \tag{5.4b} \label{eq:rob_gen_cvx_2}
\end{align}
where $\gU$ is any convex additive perturbation set. The next theorem shows that \cref{eq:rob_gen_cvx} is an upper-bound to the robust loss function \cref{eq:robust_general}, with the proof provided in \Cref{sec:CVX_MINIMAX}.

\vspace{.25mm}
\begin{theorem} 
\label{THM:CVX_MINIMAX}
    Let $(v_{\rob_i}^\star, w_{\rob_i}^\star)\iPhat$ denote a solution of \cref{eq:rob_gen_cvx} and define $\mhs$ as $|\{i : v_{\rob_i}^\star \neq 0\}| + |\{i : w_{\rob_i}^\star \neq 0\}|$.
    When the ANN width $m$ satisfies $m \geq \mhs$, the optimization problem \cref{eq:rob_gen_cvx} provides an upper-bound on the non-convex adversarial training problem \cref{eq:robust_general}.
    The robust ANN weights $(u_{\rob_j}^\star, \alpha_{\rob_j}^\star)\jmh$ can be recovered using \cref{eq:recover_weights}.
\end{theorem}

When the perturbation set is zero, \cref{THM:CVX_MINIMAX} reduces to \cref{THM:PILANCI}.
In light of \cref{THM:CVX_MINIMAX}, we use optimization \cref{eq:rob_gen_cvx} as a surrogate for the optimization \cref{eq:robust_general} to train the ANN.
Since \cref{eq:rob_gen_cvx} includes all $D_i$ matrices in \cref{eq:convex_general}, we have $\widehat{P} \geq P$.
While $\widehat{P}$ is at most $2^n$ in the worst case, since $\epsilon$ is often small, we expect $\widehat{P}$ to be relatively close to $P$, where $P \leq 2r \big( \frac{e(n-1)}{r} \big)^r$ as discussed above.
As will be shown in \Cref{sec:prac_adv}, an approximation to \cref{eq:rob_gen_cvx} can be applied to train ANNs with widths much less than $\mhs$.

The robust constraints in \cref{eq:rob_gen_cvx_2} force all points within the perturbation set to be feasible. 
Intuitively, for every $j \in [\widehat{m}^\star]$, \cref{eq:rob_gen_cvx_2} forces the ReLU activation pattern sgn$\big( (X+\Delta) u_{\rob_j}^\star \big)$ to stay the same for all $\Delta$ such that $X+\Delta\in\gU$.
Moreover, if $\Delta_{\rob}^\star$ denotes a solution to the inner maximization in \cref{eq:rob_gen_cvx_1}, then $X + \Delta_{\rob}^\star$ corresponds to the worst-case adversarial inputs for the recovered ANN.

\begin{corollary}
\label{CORO:ROB_CONSTRAINT}
    For the perturbation set $\gX$, the constraints in \cref{eq:rob_gen_cvx_2} are equivalent to 
    \begin{equation} \label{eq:robust_constraint}
    \begin{aligned}
        (2 D_i - I_n) X v_i \geq \epsilon \norm{v_i}_1, \quad
        (2 D_i - I_n) X w_i \geq \epsilon \norm{w_i}_1, \quad & \alliinPhat.
    \end{aligned}
    \end{equation}
\end{corollary}
The proof of \cref{CORO:ROB_CONSTRAINT} is provided in \Cref{sec:ROB_CONSTRAINT}.
Note that the left side of each inequality in \cref{eq:robust_constraint} is a vector while the right side is a scalar, which means that each element of the corresponding vector should be greater than or equal to that scalar.

We will show that the new problem can be efficiently solved in important cases.
Specifically, \cref{eq:rob_gen_cvx} reduces to a classic convex optimization problem when $\ell(\widehat{y}, y)$ is the hinge loss, the squared loss, or the binary cross-entropy loss.
Due to space restrictions, the result for the squared loss is presented in \Cref{sec:sqr_loss}.

\subsection{Practical Convex Adversarial Training Algorithm} \label{sec:prac_adv}

Since \cref{thm:prac} does not rely on assumptions about the matrix $X$, it applies to an arbitrary $X+\Delta$ matrix, and naturally extends to the convex adversarial training formulation \cref{eq:rob_gen_cvx}.
Therefore, an approximation to \cref{eq:rob_gen_cvx} can be applied to train robust ANNs with widths much less than $\mhs$.
Similar to the strategy rendered in \cref{alg:train}, we use a subset of the $D_i$ matrices for practical adversarial training. 
Since the $D_i$ matrices depend on the perturbation $\Delta$, we also add randomness to the data matrix $X$ in the sampling process to cover $D_i$ matrices associated with different perturbations, leading to \cref{alg:adv_train}.
$P_a$ and $S$ are preset parameters that determine the number of random weight samples, with $P_a \times S \geq P_s$.

\begin{algorithm}[t]
    \begin{algorithmic}[1]
        \For{$h=1$ to $P_a$}
            \State $a_h \sim \gN (0, I_d)$ i.i.d.
            \State $D_{h1} \leftarrow \diag ([X a_h \geq 0])$
            \For{$j=2$ to $S$}
                \State $R_{hj} \leftarrow [r_1, \dots, r_d]$, where $r_\kappa \sim \gN(\mathbf{0}, I_n), \forall \kappa\in[d]$
                \State $D_{hj} \leftarrow \diag ([\overline{X}_{hj} a_h \geq 0])$, where $\overline{X}_{hj} \leftarrow X + \epsilon \cdot \sgn(R_{hj})$
                \State Discard repeated $D_{hj}$ matrices
                \State \textbf{break if} $P_s$ distinct $D_{hj}$ matrices has been generated
            \EndFor
        \EndFor
        \State Solve \vspace{-5mm}
        \begin{align} \label{eq:rob_gen_cvx_s}
            \min_{(v_i, w_i)\iPhat} & 
            \begin{pmatrix}
                \displaystyle \max\allDeltaU \ell \bigg( \sum\iPs D_h (X+\Delta) (v_h - w_h), y \bigg) + \ \beta \sum\iPs \big( \norm{v_h}_2 + \norm{w_h}_2 \big) \hfill
            \end{pmatrix} \\
            \ST \quad & \min\allDeltaU (2 D_h - I_n) (X+\Delta) v_h \geq 0, \quad \alliinPs \nonumber, \\
            & \min\allDeltaU (2 D_h - I_n) (X+\Delta) w_h \geq 0, \quad \alliinPs \nonumber.
        \end{align}
        \vspace{-2mm}
        \State \text{Recover $u_1, \ldots, u_{m_s}$ and $\alpha_1, \ldots, \alpha_{m_s}$ from the solution $(v_{\rob s_h}^\star,w_{\rob s_h}^\star)_{h=1}^{P_s}$ of \cref{eq:rob_gen_cvx_s}} \text{using \cref{eq:recover_weights}}.
    \end{algorithmic}
    \caption{Practical convex adversarial training}
    \label{alg:adv_train}
\end{algorithm}

\subsection{Convex Hinge Loss Adversarial Training} \label{sec:hinge}

While the inner maximization of the robust problem \cref{eq:rob_gen_cvx} is still hard to solve in general, it is tractable for some loss functions.
The simplest case is the piecewise-linear hinge loss $\ell(\widehat{y}, y) = (1-\widehat{y} \odot y)_+$, which is widely used for classification.
Here, we focus on binary classification with $y \in \{-1, 1\}^n$.\footnote{
Other $\ell_p$ norm-bounded additive perturbation sets can be similarly analyzed, as shown in \Cref{sec:lpnorm}.
Moreover, extending this section's analysis to arbitrary convex piecewise-affine loss functions is straightforward.
}

Consider the training problem for a one-hidden-layer ANN with $\ell_2$ regularized hinge loss:
\begin{equation} \label{eq:hinge_std}
    \min_{(u_j, \alpha_j)\jm}
    \bigg( \frac{1}{n} \cdot \one^\top \Big( \one - y \odot \sum\jm (X u_j)_+ \alpha_j \Big)_+ + \frac{\beta}{2} \sum\jm \big( \norm{u_j}_2^2 + \alpha_j^2 \big) \bigg).
\end{equation}

The adversarial training problem considering the $\ell_\infty$-bounded adversarial data perturbation set $\gX$ is:
\begin{align} \label{eq:hinge_adv}
    & \min_{(u_j, \alpha_j)\jm} 
    \begin{pmatrix}
        \displaystyle \max\allDelta \frac{1}{n} \cdot \one^\top \bigg( \one - y \odot \sum\jm \big( (X+\Delta) u_j \big)_+ \alpha_j \bigg)_+ + \frac{\beta}{2} \sum\jm \big( \norm{u_j}_2^2 + \alpha_j^2 \big) \hfill \\
    \end{pmatrix}
\end{align}

Applying \cref{THM:CVX_MINIMAX} and \cref{CORO:ROB_CONSTRAINT} leads to the following formulation as an upper bound on \cref{eq:hinge_adv}:
\begin{align} \label{eq:hinge_ADV_D_minmax}
    \min_{(v_i, w_i)\iPhat} & 
    \begin{pmatrix}
        \displaystyle \max\allDelta \frac{1}{n} \cdot \one^\top \bigg( \one - y \odot \sum\iPhat D_i (X+\Delta) (v_i - w_i) \bigg)_+ \hspace{-1.5mm} + \beta \sum\iPhat \big( \norm{v_i}_2 + \norm{w_i}_2 \big) \hfill
    \end{pmatrix} \nonumber \\
    \ST \; & \;\;\; (2 D_i - I_n) X v_i \geq \epsilon \norm{v_i}_1, \;\;\;
    (2 D_i - I_n) X w_i \geq \epsilon \norm{w_i}_1, \;\;\; \alliinPhat.
\end{align}

For the purpose of generating the $D_1, \dots, D_{\widehat{P}}$ matrices, instead of enumerating an infinite number of points in $\gX$, we only need to enumerate all vertices of $\gX$, which is finite.
This is because the solution $\Delta_\mathrm{hinge}^\star$ to the inner maximum always occurs at a vertex of $\gX$, as will be shown in \cref{thm:inner_max}. 
Solving the inner maximization of \cref{eq:hinge_ADV_D_minmax} in closed form leads to the next theorem, whose proof is provided in \Cref{sec:INNER_MAX}.

\begin{theorem} \label{thm:inner_max}
    For the binary classification problem, the inner maximum of \cref{eq:hinge_ADV_D_minmax} is attained at $\Delta_\mathrm{hinge}^\star = - \epsilon \cdot \sgn \Big( \sum\iPhat D_i y (v_i - w_i)^\top \Big)$, and the bi-level optimization problem \cref{eq:hinge_ADV_D_minmax} is equivalent to the classic optimization problem:
    \vspace{-1mm}
    \begin{align} \label{eq:hinge_adv_D}
        \min_{(v_i, w_i)\iPhat} \frac{1}{n} & \sum_{k=1}^n \left( 1 - y_k \sum\iPhat d_{ik} x_k^\top (v_i - w_i) + \epsilon \biggnorm{\sum\iPhat d_{ik} (v_i - w_i)}_1 \right)_+ \hspace{-2mm} + \beta \sum\iPhat \big( \norm{v_i}_2 + \norm{w_i}_2 \big)  \nonumber \\[.1mm]
        \ST \quad & (2 D_i - I_n) X v_i \geq \epsilon \norm{v_i}_1, \;\;
        (2 D_i - I_n) X w_i \geq \epsilon \norm{w_i}_1, \;\; \alliinPhat,
    \end{align}
    where $d_{ik}$ denotes the $k\th$ diagonal element of $D_i$.
\end{theorem}

The problem \cref{eq:hinge_adv_D} is a finite-dimensional convex program that upper-bounds \cref{eq:hinge_adv}, the robust counterpart of \cref{eq:hinge_std}.
We can thus solve \cref{eq:hinge_adv_D} to robustly train the ANN.

\subsection{Convex Binary Cross-Entropy Loss Adversarial Training} \label{sec:ce_loss}

The binary cross-entropy loss is also widely used in binary classification.
Here, we consider a scalar-output ANN with a scaled tanh output layer for binary classification with $y \in \{0, 1\}^n$.
The loss function $\ell(\cdot)$ in this case is $\ell (\yhat, y) = -2 \yhat^\top y + \one^\top \log(e^{2\yhat}+1)$.
The non-convex adversarial training formulation considering the $\ell_\infty$-bounded data uncertainty set $\gX$ is then:
\vspace{-.8mm}
\begin{align} \label{eq:ce_adv}
    \min_{(u_j, \alpha_j)\jm} & \begin{pmatrix}
        \displaystyle \max_{\norm{\Delta}_{\max} \leq \epsilon} \frac{1}{n} \sum\kn \Big( -2\yhat_k y_k + \log(e^{2\yhat_k}+1) \Big)
    \end{pmatrix} + \frac{\beta}{2} \sum\jm \big( \norm{u_j}_2^2 + \alpha_j^2 \big) \\[-1.6mm]
    \text{where} \;\; &\; \yhat := \sum\jm \big( (X+\Delta) u_j \big)_+ \alpha_j. \nonumber
\end{align}
Applying \cref{THM:CVX_MINIMAX} and \cref{CORO:ROB_CONSTRAINT} leads to the following optimization problem as an upper bound on \cref{eq:ce_adv}:
\begin{equation} \label{eq:ce_adver1}
    \begin{aligned}
        \min_{(v_i, w_i)\iPhat} & \begin{pmatrix}
            \displaystyle \max_{\norm{\Delta}_{\max} \leq \epsilon} \frac{1}{n} \sum\kn \bigg( -2\yhat_k y_k + \log(e^{2\yhat_k}+1) \bigg)
        \end{pmatrix} + \beta \sum\iPhat \Big( \norm{v_i}_2 + \norm{w_i}_2 \Big) \\
        \ST \quad & (2D_i-I_n) X v_i \geq \epsilon \norm{v_i}_1, \ (2D_i-I_n) X w_i \geq \epsilon \norm{w_i}_1, \quad \alliinPhat, \\[-1mm]
        & \yhat_k = \sum\iPhat d_{ik} x_k^\top (v_i-w_i) + \sum\iPhat d_{ik} \delta_k^\top (v_i-w_i).
    \end{aligned}
\end{equation}

Consider the convex optimization formulation
\begin{align} \label{eq:convex_ce_adver}
    \min_{ (v_i, w_i)\iPhat } & \frac{1}{n} \bigg( \sum\kn f \circ g_k \big( \{v_i,w_i\}\iPhat \big) \bigg) + \beta \sum\iPhat \Big( \norm{v_i}_2 + \norm{w_i}_2 \Big) \nonumber \\
    \ST \quad & (2D_i-I_n) X v_i \geq \epsilon \norm{v_i}_1, \ (2D_i-I_n) X w_i \geq \epsilon \norm{w_i}_1, \quad \alliinPhat \nonumber \\[2mm]
    & f(u) := \log(e^{2u}+1), \\[-2mm]
    & g_k \big( \{v_i,w_i\}\iPhat \big) := (2y_k-1) \sum\iPhat d_{ik} x_k^\top (v_i-w_i) + \epsilon \cdot \biggnorm{\sum\iPhat d_{ik} (v_i-w_i)}_1, \;\; \allkinn. \nonumber
\end{align}

The next theorem establishes the equivalence between \cref{eq:convex_ce_adver} and \cref{eq:ce_adver1}. The proof is provided in \Cref{sec:CEadvthm}.
\begin{theorem} \label{thm:CEadv}
     The optimization \cref{eq:convex_ce_adver} is a convex program that is equivalent to the bi-level optimization \cref{eq:ce_adver1}, and can be used as a surrogate for \cref{eq:ce_adv} to train robust ANNs.
     The worst-case perturbation is $\Delta_\mathrm{BCE}^\star = -\epsilon \cdot \sgn \Big( (2 y - 1) \sum\iPhat D_i (v_i-w_i)^\top \Big)$.
\end{theorem}

Note that the worst-case perturbation occurs at the same location as for the hinge loss case, which is a vertex in $\gX$.
Thus, for the purpose of generating the $D_1, \dots, D_{\widehat{P}}$ matrices, we again only need to enumerate all vertices of $\gX$ instead of all points in $\gX$.

\section{Numerical Experiments} \label{sec:exp}

Due to space restrictions, we focus on binary classification with the hinge loss, and defer the squared loss results to \Cref{sec:sl_simulation}.

\subsection{Approximated Convex Standard Training} \label{sec:prac_exp}
\begin{figure*}
    \centering
    \begin{subfigure}{0.48\textwidth}
        \centering
        \includegraphics[width=\textwidth, height=.75\textwidth] {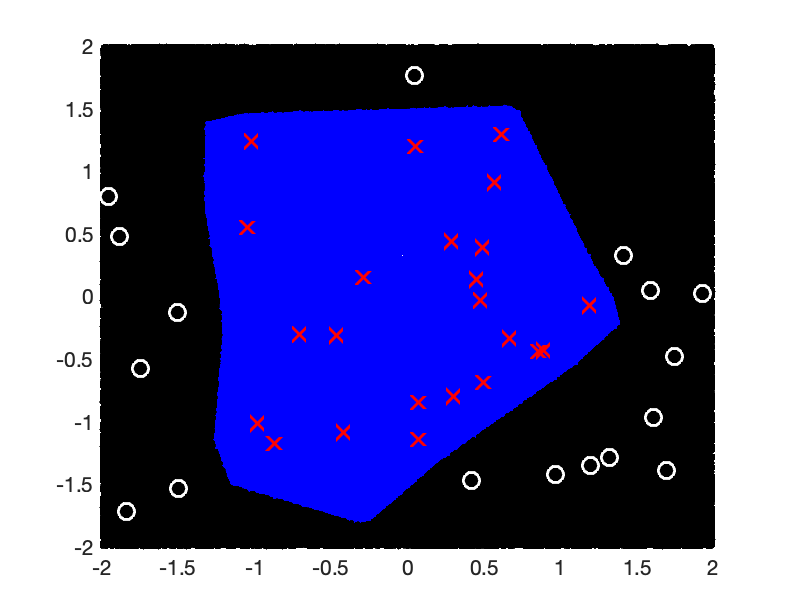}
        \caption{
        A randomized 2-dimensional dataset used in this experiment.
        The red crosses are positive training points and the white circles are negative training points.
        The region classified as positive is in blue, whereas the negative region is in black.
        }
        \label{fig:stability1}
    \end{subfigure}
    \quad
    \begin{subfigure}{0.48\textwidth}
        \centering
        \includegraphics[width=\textwidth, height=.82\textwidth] {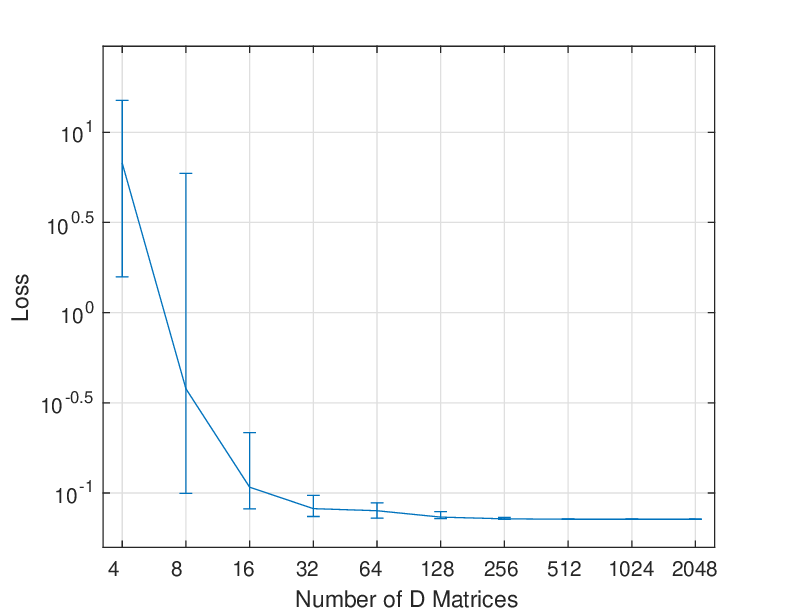}
        \caption{
        The optimized training loss for each $P_s$.
        When $P_s$ reaches 128, the mean and variance of the optimized loss become very small.
        }
        \label{fig:stability2}
    \end{subfigure}
    \vspace{-3.5mm}
    \caption{Analyzing the effect of $P_s$ on convex standard training.}
    \label{fig:stability}
    \vspace{-1mm}
\end{figure*}

In this subsection, we use numerical experiments to demonstrate the efficacy of practical standard training (\cref{alg:train}) and to show the level of suboptimality of the ANN trained using \cref{alg:train}.\footnote{
For all non-ADMM experiments in this paper, CVX \citep{cvx} and CVXPY \citep{agrawal2018rewriting,diamond2016cvxpy} with the MOSEK \citep{mosek} solver was used for solving optimization on a laptop computer, unless otherwise stated.
Off-the-shelf solvers supported by CVX and CVXPY often treat the convex training problem as a general SOCP.
Among all solvers that we experimented with on the convex training formulation, MOSEK is the most efficient.
}
The experiment was performed on a randomly generated dataset with $n=40$ and $d=2$ shown in \cref{fig:stability1}.
The upper bound on the number of ReLU activation patterns is $4 \big( \frac{e(39)}{2} \big)^2 = 11239$.
We ran \cref{alg:train} to train ANNs using the hinge loss with the number of $D_h$ matrices equal to $4, 8, 16, \dots, 2048$ and compared the optimized loss.\footnote{
To reliably sample $P_s$ matrices, $P_a \cdot S$ in \cref{alg:adv_train} was set to a large number (81920), and the sampling was terminated when a sufficient number of $D_h$ matrices was generated. The regularization strength $\beta$ was chosen to be $10^{-4}$.
}
We repeated this experiment 15 times for each setting, and plotted the loss in \cref{fig:stability2}.
The error bars show the loss values achieved in the best and the worst runs.
When there are more than 128 matrices (much less than the theoretical bound on $P$), \cref{alg:train} yields consistent and favorable results.
Further increasing the number of $D$ matrices does not produce a significantly lower loss.
By \cref{thm:prac}, $P_s = 128$ corresponds to $\psi \xi = 0.318$.

\subsection{The ADMM Convex Training Algorithm} \label{sec:ADMMexp}

We now present the experiment results with the ADMM training algorithm.
We use \cref{alg:ADMM} to solve the approximate convex training formulation \cref{eq:prac_clean} with the sampled $D_h$ matrices.
In \Cref{sec:ADMMparam}, we discuss our experiments' ADMM hyperparameter settings and present guidelines on selecting them.

\subsubsection{Squared Loss (closed form \texorpdfstring{$u$}{u} Updates) -- Convergence} \label{sec:ADMMconv} \label{sec:ADMMSLexp1}

For the case of the squared loss, the closed-form solution \cref{eq:uupdate} is used for the $u$ updates. We first demonstrate the convergence of the proposed ADMM algorithm using illustrative random data with dimensions $n=6, d=5, P_s=8$.
CVX \citep{cvx} with the IPM-based MOSEK solver \citep{mosek} was used to solve the optimal objective of \cref{eq:convex_general} as the ground truth.

In the figures, we use $l^\star_\mathrm{CVX}$ to denote the CVX optimal objective and use $l^\star_\mathrm{ADMM}$ to denote the objective that ADMM converges to as the number of iterations $k$ goes to infinity.
There are several methods to calculate the training loss obtained by ADMM.
For fair comparisons among ADMM, CVX, and SGD, we use \cref{eq:recover_weights} to recover the ANN weights $(u_j, \alpha_j) \jm$ from the ADMM optimization variables $(v_h^k, w_h^k)_{h=1}^{P_s}$, and use $(u_j, \alpha_j) \jm$ to calculate the true non-convex training loss \cref{eq:nonconvex_general}.
The loss at each iteration calculated via this method is denoted as $\lua$, and the ADMM solution $l^\star_\mathrm{ADMM}$ is also calculated via this method.
At each iteration, we also compute the convex objective of \cref{eq:convex_general} using $(v_h^k, w_h^k)_{h=1}^{P_s}$, denoted as $\lvw$.
Since ADMM uses dual variables to enforce the constraints, while the ADMM solution is feasible as $k$ goes to infinity, the intermediate iterations may not be feasible.
When the constraints in \cref{eq:convex_general} are satisfied, it holds that $\lua = \lvw$. Otherwise, $\lua$ may be different from $\lvw$.
The gap between $\lua$ and $\lvw$ indirectly characterizes the feasibility of the ADMM intermediate solutions.
When this gap is small, $(v_h^k, w_h^k)_{h=1}^{P_s}$ should be almost feasible.
When this gap is large, the constraints may have been severely violated.

\begin{figure}
    \centering
    \begin{subfigure}{\textwidth}
        \includegraphics[width=1\textwidth]{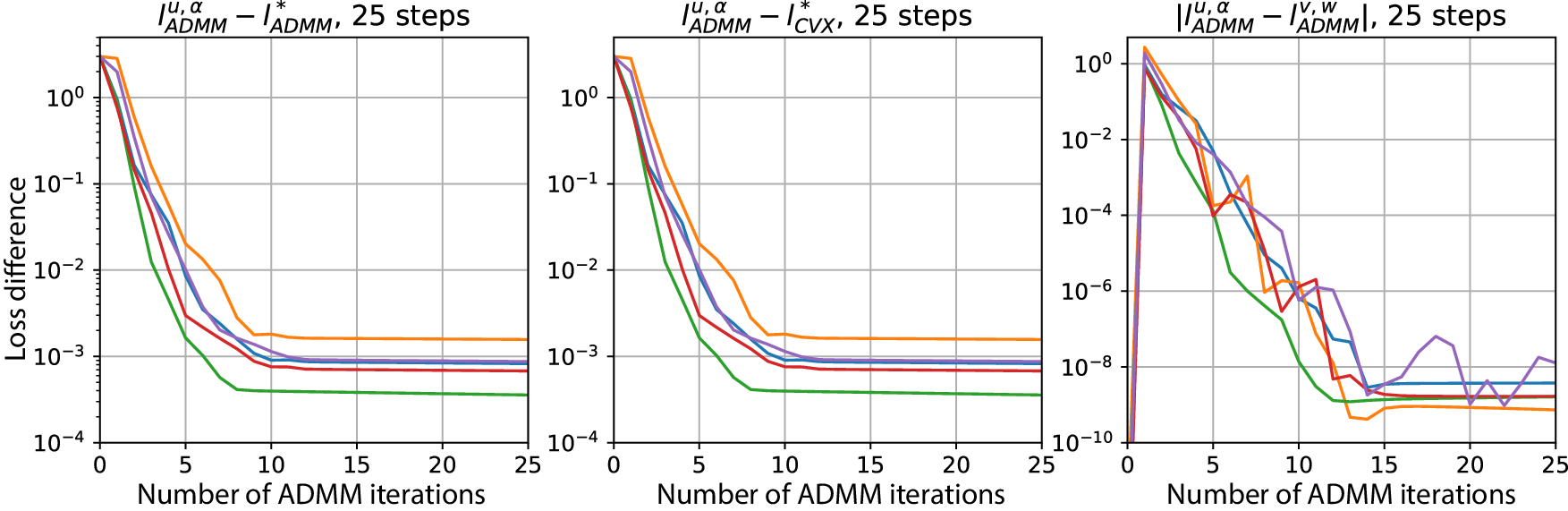}
        \vspace{-4.5mm}
        \caption{$\lua - l^\star_{\mathrm{ADMM}} \hspace{10cm}$}
        \label{fig:ADMMConv2_a}
    \end{subfigure} \\[-10.2mm]
    \hspace{51mm}
    \begin{subfigure}{.31\textwidth}
        \caption{$\lua - l^\star_{\mathrm{CVX}}$}
        \label{fig:ADMMConv2_b}
    \end{subfigure}
    \begin{subfigure}{.31\textwidth}
        \caption{$\big| \lua - \lvw \big|$}
        \label{fig:ADMMConv2_c}
    \end{subfigure}
    \vspace{-2.5mm}
    \caption{
    Gap between the cost returned by ADMM for the first 25 iterations and the true optimal cost for the five independent runs.
    }
    \label{fig:ADMMConv2}
    \vspace{-2mm}
\end{figure}

While it can be expensive for ADMM to converge to a high precision (note that the algorithm is guaranteed to linearly converge to a global minimum given an ample computation time according to \cref{thm:ADMM}), an approximate solution is usually sufficient for achieving a high validation accuracy since decreasing the training loss excessively could induce overfitting.
Therefore, when performing the experiments, we apply early stopping \citep{Prechelt12}, a common training technique that improves generalization. 
\cref{fig:ADMMConv2_a}, \ref{fig:ADMMConv2_b} shows that a precision of $10^{-3}$ can be achieved within 25 iterations.
Moreover, \cref{fig:ADMMConv2_c} shows that the solution after 25 iterations violates the constraints insignificantly.
This behavior of ``converging rapidly in the first several steps and slowing down (to a linear rate) afterward'' is typical for the ADMM algorithm.
As will be shown next, a medium-accuracy solution returned by only a few ADMM iterations can achieve a better prediction performance than the CVX solution.
In \Cref{sec:ADMMasym}, we present empirical results that demonstrate the asymptotic convergence properties of ADMM.

\begin{figure}
    \centering
    \begin{subfigure}{.88\textwidth}
        \centering
        \includegraphics[width=.96\textwidth, height=.37\textwidth, trim={0 0 95mm 0}, clip] {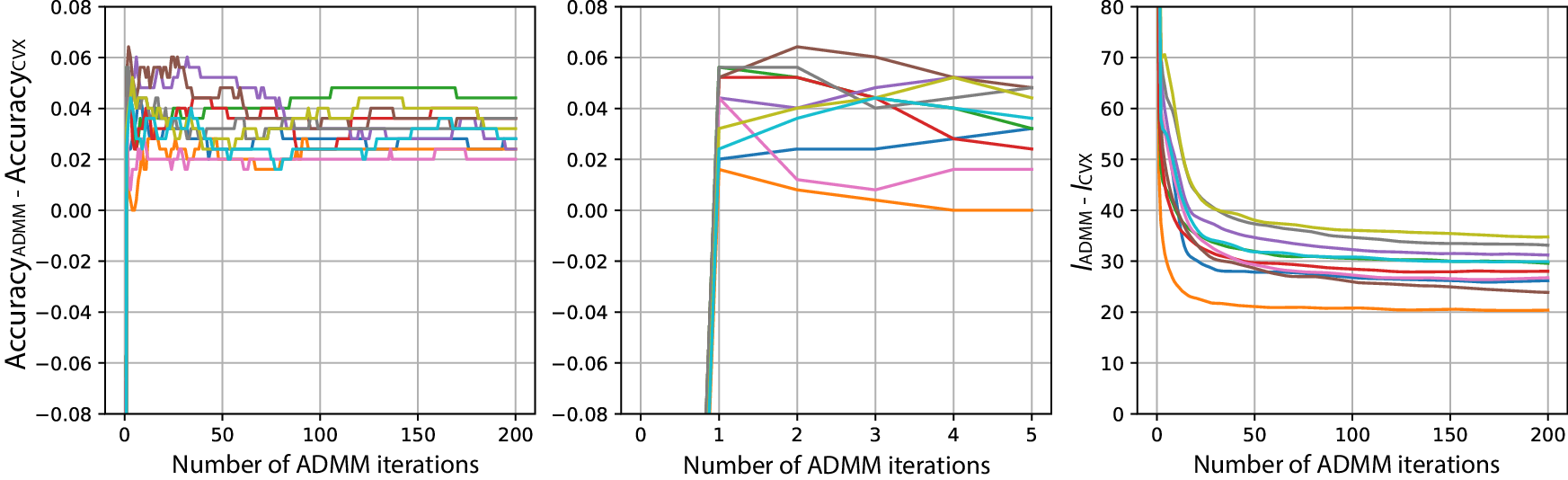}
        \vspace{-.5mm}
        \caption{$\mathrm{Accuracy}_{\mathrm{ADMM}} - \mathrm{Accuracy}_{\mathrm{CVX}}$ (positive \\ means the ADMM solution outperforms CVX). $\hspace{6cm}$}
        \label{fig:ADMMAcc_a}
    \end{subfigure} \\[-14.2mm]
    \hspace{74mm}
    \begin{subfigure}{.26\textwidth}
        \caption{\cref{fig:ADMMAcc_a} zoomed-in to the first five iterations.}
        \label{fig:ADMMAcc_b}
    \end{subfigure}
    \vspace{-3mm}
    \caption{Comparing the ANNs trained with ADMM and with CVX over ten independent runs on the mammographic masses dataset.}
    \label{fig:ADMMAcc}
    \vspace{-1mm}
\end{figure}

To visualize how the prediction performance achieved by the model changes as the ADMM iteration progresses, we run the ADMM iterations on the ``mammographic masses'' dataset from the UCI Machine Learning Repository \citep{Dua:2019}, and record the prediction accuracy on the validation set at each iteration.
70\% of the dataset is randomly selected as the training set, and the other 30\% is used as the validation set.
\cref{fig:ADMMAcc} plots the difference between the ADMM accuracy and the CVX accuracy at each iteration.
In all experiments, all variables in the ADMM algorithm are initialized to be zero.

All ten runs achieve superior validation accuracy throughout the first 200 iterations compared with the CVX baseline.
Even the first five iterations outperform the baseline, with the best run outperforming CVX by 6\%.
After about 80 iterations, the accuracy stabilizes at around 2\% to 4\% better than CVX.
In conclusion, the prediction performance of the classifiers trained by ADMM is superior even when only a few iterations are run.

\begin{figure}
    \centering
    \begin{subfigure}{.48\textwidth}
        \centering
        \adjustbox{trim={0} {0} {.5\width} {0},clip} {\includegraphics[width=1.72\textwidth, height=.64\textwidth] {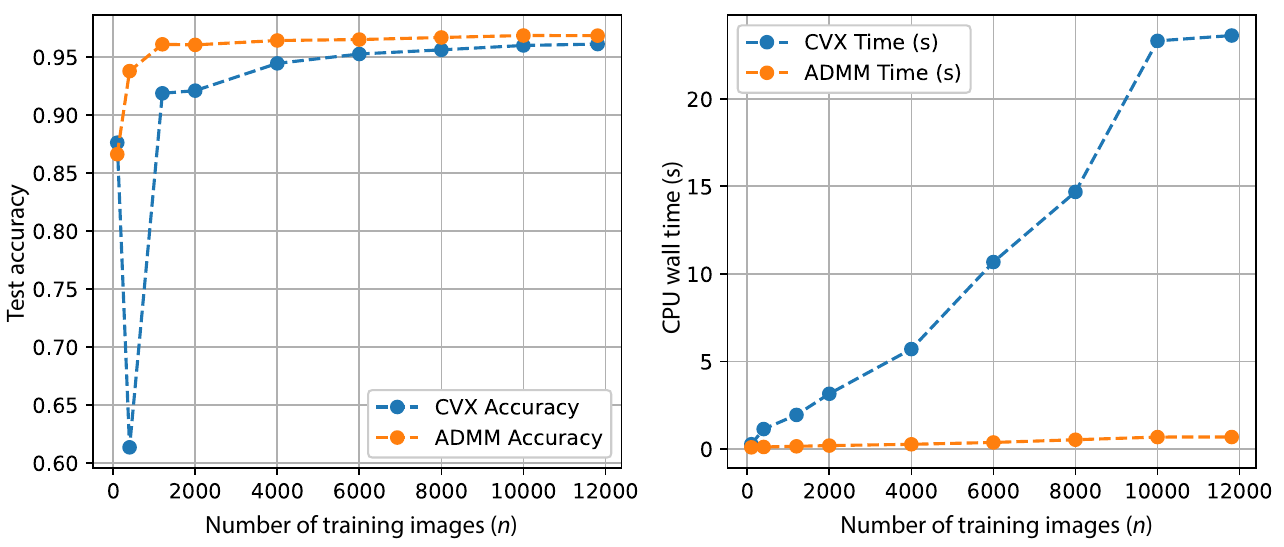}}
        \vspace{-1.2mm}
        \caption{Average validation accuracy for each $n$.}
        \label{fig:ADMMComn_a}
    \end{subfigure}
    \begin{subfigure}{.48\textwidth}
        \centering
        \adjustbox{trim={.5\width} {0} {0} {0},clip} {\includegraphics[width=1.72\textwidth, height=.64\textwidth] {Figures/Complexity_n.pdf}}
        \vspace{-1.2mm}
        \caption{Average CPU wall time for each $n$.}
        \label{fig:ADMMComn_b}
    \end{subfigure} \\[-.5mm]
    \begin{subfigure}{.48\textwidth}
        \centering
        \adjustbox{trim={0} {0} {.5\width} {0},clip} {\includegraphics[width=1.72\textwidth, height=.64\textwidth] {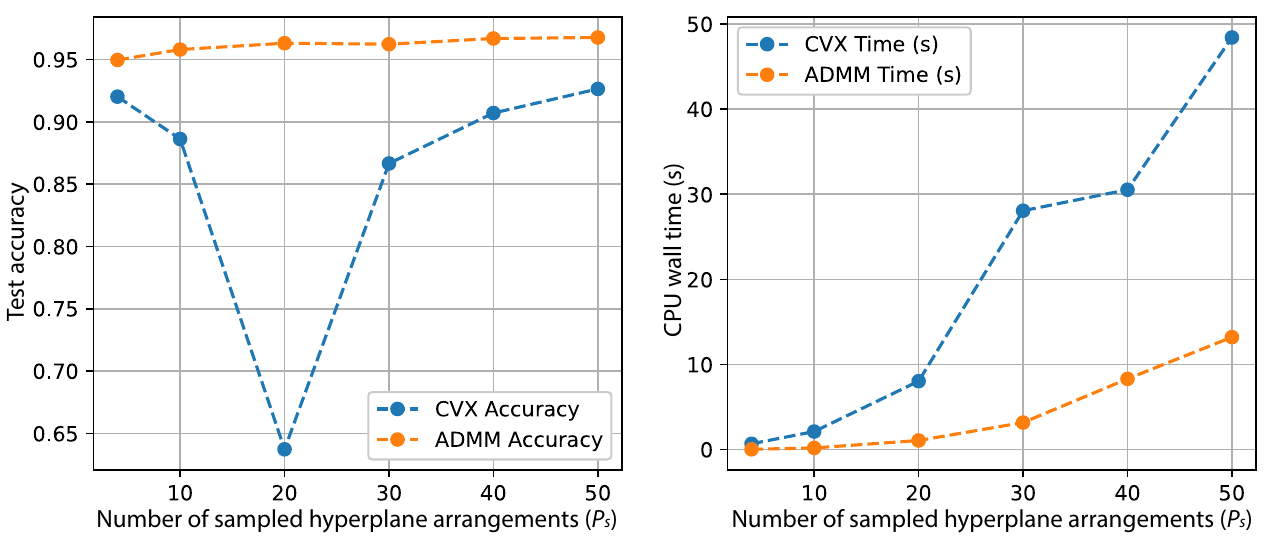}}
        \vspace{-1.2mm}
        \caption{Average validation accuracy for each $P_s$.}
        \label{fig:ADMMComP_c}
    \end{subfigure}
    \begin{subfigure}{.48\textwidth}
        \centering
        \adjustbox{trim={.5\width} {0} {0} {0},clip} {\includegraphics[width=1.72\textwidth, height=.64\textwidth] {Figures/Complexity_P.pdf}}
        \vspace{-1.2mm}
        \caption{Average CPU wall time for each $P_s$.}
        \label{fig:ADMMComP_d}
    \end{subfigure}
    \vspace{-4.5mm}
    \caption{Analyzing the effect of $n$ and $P_s$ on ADMM convex training with the MNIST dataset.}
    \label{fig:ADMMCom}
    \vspace{-2mm}
\end{figure}

\subsubsection{Squared Loss (Closed Form \texorpdfstring{$u$}{u} Updates) -- Complexity} \label{sec:ADMMSLexp2}

To demonstrate the computational complexity of the proposed ADMM method, we used the ADMM method to train ANNs on the downsampled MNIST handwritten digits dataset with $d=100$.
The task was to perform binary classification between digits ``2'' and ``8''. We first fix $P_s=8$ and vary $n$ from 100 to 11809.\footnote{
11809 is the total number of 2's and 8's in the training set.
} We independently repeat the experiment five times for each $n$ setting, and present the average results in \cref{fig:ADMMComn_a}, \ref{fig:ADMMComn_b}.
In each experiment, ADMM is allowed to run six iterations, which is sufficient to train an accurate ANN.
For all choices of $n$ except $n=100$, the ANNs trained with ADMM attain higher accuracy than CVX networks.
This is because while ADMM and CVX solve the same problem, the medium-precision solution from ADMM generalizes better than the high-precision CVX solution.
More importantly, as $n$ increases, the CPU time required for CVX grows much faster than ADMM's execution time, which increases linearly in $n$.
While it is also theoretically possible to run the IPM to a medium precision, even a few IPM iterations become too expensive when $n$ is large.
Moreover, since the IPM uses barrier functions to approximate the constraints, a medium-precision solution produced by the IPM may have feasibility issues, while the ADMM solution sequence generally has good feasibility, as illustrated in \cref{fig:ADMMConv2}.

Similarly, we fix $n=1000$ and vary $P_s$ from 4 to 50.
The average result over five runs is shown in \cref{fig:ADMMComP_c}, \ref{fig:ADMMComP_d}.
Once again, the proposed ADMM algorithm achieves a higher accuracy for each $P_s$, and the average CPU time of ADMM grows much slower than the CVX CPU time.
When $P_s$ is 20, all five CVX runs achieve low validation accuracy, possibly because the structure of the true underlying distribution cannot be well approximated with a combination of 20 linear classifiers.
\cref{fig:ADMMComP_c}, \ref{fig:ADMMComP_d} also show that the CPU time scales quadratically with $P_s$, confirming our theoretical analysis of the $\gO(n P_s + d^2 P_s^2)$ per-iteration complexity.

\subsubsection{Squared Loss (Closed Form \texorpdfstring{$u$}{u} Updates) -- MNIST, Fashion MNIST, and CIFAR-10} \label{sec:ADMMSLexp3}

We now demonstrate the effectiveness of the proposed ADMM algorithm on all images of ``2'' and ``8'' in the MNIST dataset without downsampling ($n=11809$ and $d=784$).
The parameter $P_s$ was chosen to be 24, corresponding to a network width of at most 48.
The prediction accuracy on the validation set, the training loss, and the CPU time are shown in \cref{tbl:ADMMMnist}.
The baseline method ``CVX'' corresponds to using CVX to globally optimize the ANN by solving \cref{eq:convex_general}, while ``Back-prop'' denotes the conventional method that performs an SGD local search on the non-convex cost function \cref{eq:nonconvex_general}.

\cref{tbl:ADMMMnist} shows that the training loss returned by ADMM is higher than the true optimal cost but lower than the back-propagation solution.
Note that the difference between the ADMM training loss and the CVX loss is due to the early stopping strategy applied to ADMM.
ADMM will converge to the true global optimal with a sufficient computation time, but we prematurely terminate the algorithm once the validation accuracy becomes satisfactory so that the rapid initial convergence of ADMM can be fully exploited.
In contrast, back-propagation does not have this guarantee due to the non-convexity of \cref{eq:nonconvex_general}.
Moreover, back-propagation is highly sensitive to the initialization and the hyperparameters. While ADMM also requires a pre-specified step size $\gamma_a$, it is much more stable: its convergence to a primal optimum does not depend on the step size \citep[Appendix A]{Boyd11}.
An optimal step size speeds up the training, but a suboptimal step size is also acceptable.

ADMM achieves a higher validation accuracy than both CVX and back-propagation SGD.
Once again, while ADMM and CVX solve the same problem, the CVX solution suffers from overfitting and thus cannot generalize well to the validation data.

The training time of ADMM is considerably shorter than CVX.
Specifically, assembling the matrix $I+\frac{1}{\rho} F^\top F + G^\top G$ required 22\% of the time, and the Cholesky decomposition needed 34\% of the time, while each ADMM iteration only took 4.4\% of the time.
Thus, running more ADMM iterations will not considerably increase the training time. 

\begin{table}
    \centering
    \caption{Average experiment results with the squared loss on the MNIST dataset over five independent runs. We run 10 ADMM iterations for each setting.}
    \label{tbl:ADMMMnist}
    \begin{small}
    \begin{tabular}{l|c|c|c|c}
        \toprule
        Method      & Validation Accuracy & CPU Time (s)  & Training Loss & Global Convergence \\
        \midrule
        Back-prop   & 98.86 \%      & 74.09         & 422.4         & No                 \\
        CVX         & 70.99 \%      & 14879         & 1.146         & Yes                \\
        ADMM        & 98.90 \%      & 802.2         & 223.2         & Yes                \\
        \bottomrule
    \end{tabular}
    \end{small}
\end{table}

\begin{table}
    \begin{center}
    \caption{Average experiment results with the squared loss over five independent runs.}
    \label{tbl:ADMMFMnist}
    \vspace{-.5mm}
    Fashion MNIST (42 ADMM iterations, $P_s$ set to 18) \\
    \begin{small}
    \begin{tabular}{l|c|c|c}
        \toprule
        Method         & Validation Accuracy      & CPU Time (s) & Training Loss \\
        \midrule
        Back-prop      & 99.04\% \tiny{(.0735\%)} & 183.6        & 175.1 \tiny{(4.246)} \\
        ADMM           & 98.73\% \tiny{(.0200\%)} & 167.1        & 129.7 \tiny{(13.24)} \\
        Back-prop (DS) & 98.34\% \tiny{(.0917\%)} & 18.31        & 433.0 \tiny{(10.40)} \\
        ADMM (DS)      & 98.80\% \tiny{(.0585\%)} & 6.840        & 380.1 \tiny{(17.74)} \\
        \bottomrule
    \end{tabular}
    \end{small}
    
    \vspace{2mm}
    Downsampled CIFAR-10 (30 ADMM iterations, $P_s$ set to 18) \\
    \begin{small}
    \begin{tabular}{l|c|c|c}
        \toprule
        Method         & Validation Accuracy     & CPU Time (s) & Training Loss  \\
        \midrule
        Back-prop (DS) & 90.90\% \tiny{(.305\%)} & 122.7        & 991.5 \tiny{(11.68)} \\
        ADMM (DS)      & 86.89\% \tiny{(.132\%)} & 118.6        & 607.6 \tiny{(10.76)} \\
        \bottomrule
    \end{tabular}
    \end{small}
    \end{center}

    \vspace{3.2mm}
    \begin{small}
    \begin{itemize}[leftmargin=5mm]
        \item ``DS'' denotes image downsampling with a stride of 2.
        \item The numbers in the parentheses are the standard deviations over five runs.
        \item Note that the ADMM algorithm is theoretically guaranteed to converge to an approximate global minimum, whereas back-propagation does not have this property.
    \end{itemize}
    \end{small}
    \vspace{1mm}
\end{table}

\begin{figure}
    \centering
    \includegraphics[width=.48\textwidth, trim={3mm 3mm 3mm 3mm}, clip] {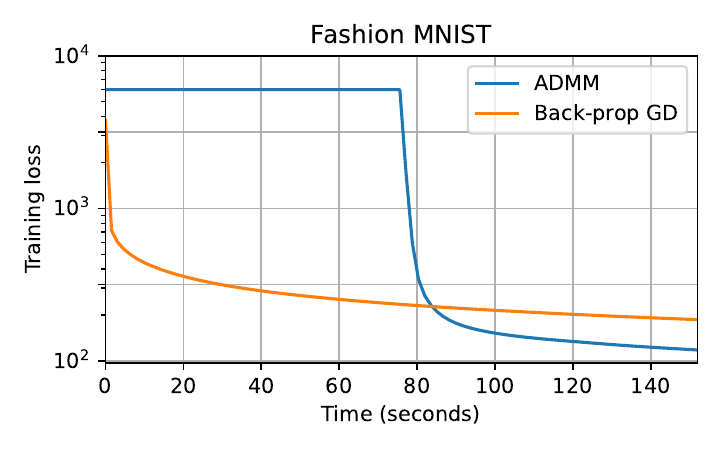} \;\;
    \includegraphics[width=.48\textwidth, trim={3mm 3mm 3mm 3mm}, clip] {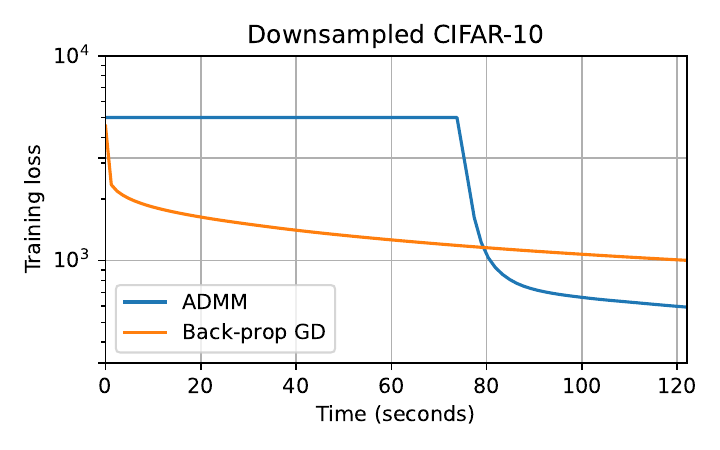}
    \caption{
    The learning curves of the closed-form ADMM algorithm and back-propagation gradient descent.
    The flat parts of the ADMM curves represent the pre-processing time.
    }
    \vspace{-3mm}
    \label{fig:learning_curves}
\end{figure}

Next, we compare ADMM with back-propagation on the more challenging Fashion MNIST \citep{Xiao17FMNIST} and CIFAR-10 datasets.
For Fashion MNIST, we perform binary classification between the ``pullover'' and the ``bag'' classes on both full data ($n=12000$, $d=784$) and downsampled data ($n=12000$, $d=196$).
For CIFAR-10, we perform binary classification between ``birds'' and ``ships'', and downsample the images to $16 \times 16 \times 3$.
The results are presented in \cref{tbl:ADMMFMnist}, and we plot the training loss with respect to time in \cref{fig:learning_curves}.
The results show that ADMM converges faster and achieves a lower loss within the same allowed time, even though it requires preprocessing before the iterations start.
However, on these datasets, the classifiers learned via back-propagation generalize better to the validation set.
Gradient descent is known to have favorable properties for machine learning, where solutions with similar losses can have vastly different properties.
For applications where training data is abundant, ADMM is well-suited since the generalization gap would be small.

We also note that ADMM is extremely efficient on the downsampled Fashion MNIST dataset, since the faster convergence of ADMM overshadows the higher complexity associated with the decomposition when the data dimension is smaller.
This result shows that ADMM is particularly suitable for data with a dimension of around 200.

\subsubsection{Binary Cross-Entropy Loss (Iterative \texorpdfstring{$u$}{u} Updates) -- MNIST} \label{sec:ADMMCEexp}

To verify the efficacy of using the RBCD method to solve \cref{eq:ADMM1}, we similarly experiment with the binary cross-entropy loss coupled with a tanh output activation.
The resulting loss function is $\ell (\yhat, y) = -2 \yhat^\top y + \one^\top \log(e^{2\yhat}+1)$. 
Since the augmented Lagrangian's gradient in the stopping condition of \cref{alg:BCD} is difficult to obtain, we use the objective improvement amount as a surrogate.

\begin{table}
    \centering
    \caption{
    Average experiment results with the binary cross-entropy loss over five runs.
    The main advantage of ADMM-RBCD is its theoretically guaranteed global convergence.
    }
    MNIST (34 ADMM iterations, $P_s = 24$)
    \begin{small}
    \begin{tabular}{l|c|c|c}
        \toprule
        Method       & Validation Accuracy & CPU Time (s) & Training Loss \\
        \midrule
        Back-prop    & 98.91 \%            & 62.06         & 437.6        \\
        CVX          & 98.21 \%            & 14217         & 1.007        \\
        ADMM-RBCD    & 98.89 \%            & 555.8         & 310.3        \\
        \bottomrule
    \end{tabular}
    \end{small}
    \label{tbl:ADMMRBCD}
\end{table}

The experiment results are shown in \cref{tbl:ADMMRBCD}.
On the MNIST dataset, the ADMM-RBCD algorithm achieves a high validation accuracy while requiring a training time $94.6\%$ shorter than the time of globally optimizing the cost function \cref{eq:convex_general} with CVX.
ADMM-RBCD also requires less time to reach a comparable accuracy than the closed-form ADMM method with the squared loss.
On the other hand, ADMM-RBCD is still slower than back-propagation local search, trading the training speed for the global convergence guarantee.
The extremely slow pace of CVX forbids its application to even medium-scaled problems, while ADMM-RBCD makes convex training much more practical by balancing efficiency and optimality.

\subsubsection{GPU Acceleration}

The success of modern deep learning relies on the parallelized computing enabled by GPUs. Using GPUs to accelerate the proposed ADMM algorithm is straightforward.
All operations required in the ADMM algorithm (\cref{alg:ADMM}) are already implemented in existing GPU-supporting deep learning libraries like PyTorch \citep{PyTorch}.
Specifically, \cref{eq:ADMM3} consists of parallelizable algebraic operations, and we have shown that \cref{eq:ADMM2} reduces to parallelizable element-wise operations.
If the RBCD algorithm is used to solve \cref{eq:ADMM1}, then all operations are again parallelizable (as is the case for traditional back-propagation gradient descent), and auto-differentiation can be used to obtain the closed-form gradients.

To verify the effectiveness of GPU acceleration and show that ADMM-RBCD scales to wider neural networks and higher dimensions with the help of GPUs, we use the method to train binary classifiers with $P_s$ set to 120 on the CIFAR-10 dataset.
The average validation accuracy over five runs is 91.23\%.
On a MacBook Pro laptop computer, this task takes 474.5 seconds on average.
Repeating the experiment on an Nvidia V100 GPU only requires 24.64 seconds, which is a 19.25x speed-up.

\subsubsection{Summary of ADMM Experiment Results}

Based on the above experiment results, we summarize some advantages of our ADMM methods below:
\begin{enumerate}[leftmargin=10mm]
    \item While the closed-form ADMM algorithm has a higher theoretical complexity compared with back-propagation, it is guaranteed to linearly converge to a global optimum if allowed to run for a sufficiently long time, enabling efficient global optimization of neural networks. Back-propagation does not have this property.
    \item The closed-form ADMM algorithm often converges rapidly in the first few iterations. Since a moderately accurate solution is sufficient for many machine learning tasks, this fast initial convergence is highly advantageous.
    \item For datasets with a relatively small number of dimensions, the closed-form ADMM algorithm is more efficient than back-propagation (as shown in \cref{tbl:ADMMFMnist}), since the faster convergence outweighs the increased complexity.
    \item Compared with closed-form ADMM, ADMM-RBCD applies to general convex loss functions, and scales better to wide ANNs, but is less efficient, as illustrated in \cref{tbl:ADMMRBCD}. ADMM-RBCD is then a trade-off between CVX (high solution quality) and back-propagation (efficient), while maintaining the theoretically provable global convergence.
\end{enumerate}

In summary, the proposed ADMM method is particularly suited for applications where:
\begin{itemize}[leftmargin=10mm]
    \item Abundant training data exists (a low empirical risk translates to a low true risk);
    \item Accuracy is more important than computational efficiency;
    \item The number of dimensions is not too large.
\end{itemize}

\subsection{Convex Adversarial Training}

All experiment results in this section are obtained using CVX with the MOSEK solver based on the interior-point method.

\subsubsection{Hinge Loss Convex Adversarial Training -- 2D Illustration} \label{sec:AT_exp_2D}

\begin{figure}
    \centering
    \adjustbox{trim={.12\width} {.1\height} {.07\width} {.01\height},clip}{\includegraphics[width=.42\textwidth]{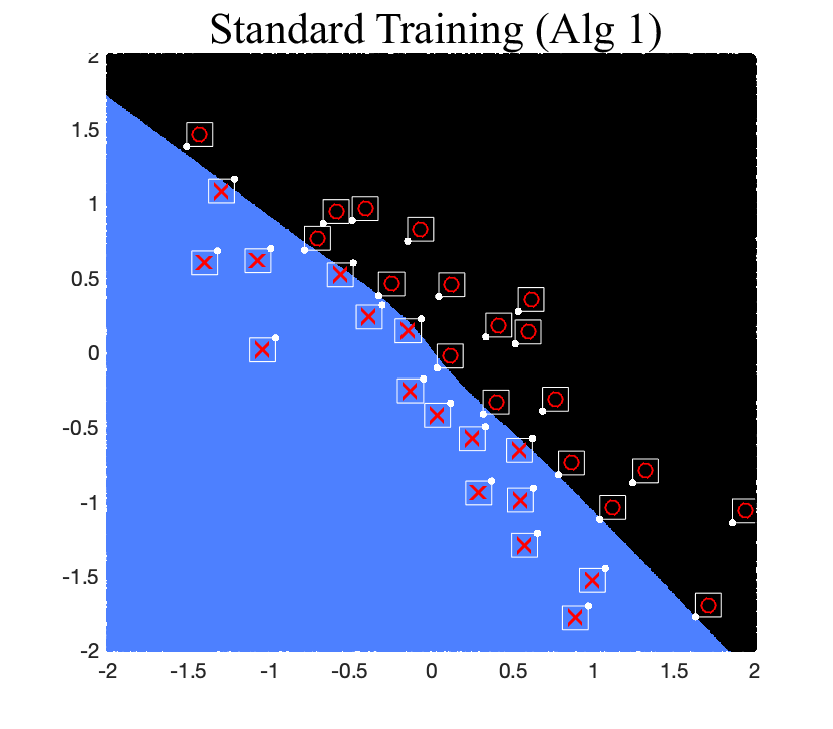}} \adjustbox{trim={.12\width} {.1\height} {.07\width} {.01\height},clip}{\includegraphics[width=.42\textwidth]{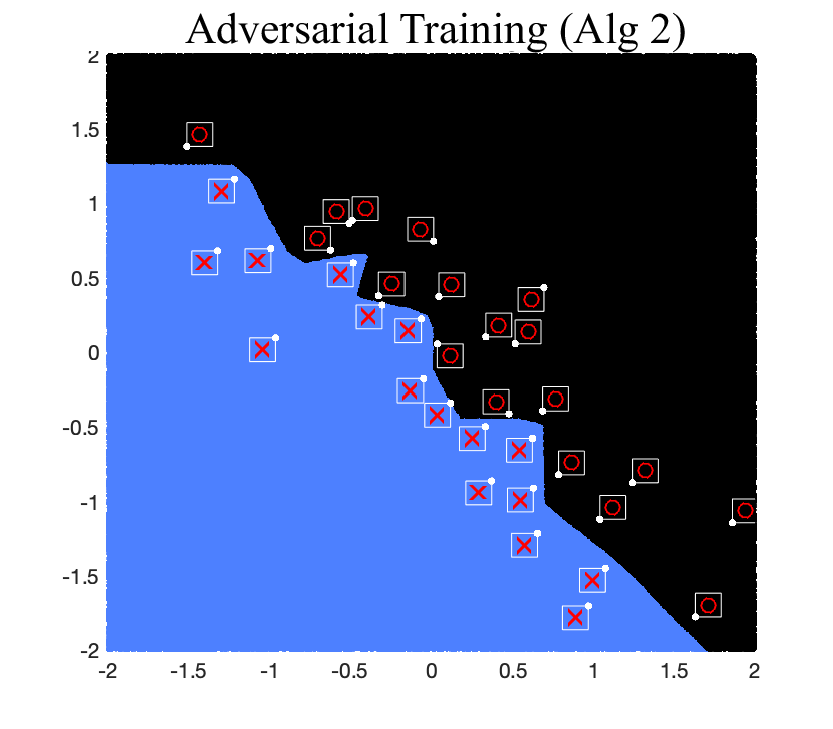}}
    
    \vspace{0.3mm}
    \begin{small}
    \begin{tabular}{l}
    	Red crosses: positive training points; $\quad\;$ Red circles: negative training points. \hfill \\
    	Blue region: classified as positive; $\qquad\;\;\,$ Black region: classified as negative. \\
    	The white box around each training data: the $\ell_\infty$ perturbation bound. \\
    	The white dot at a vertex of each box: the worst-case perturbation.
    \end{tabular}	
    \end{small}
    
    \vspace{-.5mm}
    \captionof{figure}{
    Visualization of the binary decision boundaries in a 2-dimensional space.
    \Cref{alg:adv_train} fits the perturbation boxes while the standard training fits the training points.
    }
    \label{fig:2D-toy}
    \vspace{-1.5mm}
\end{figure}

To analyze the decision boundaries obtained from convex adversarial training, we ran \cref{alg:train} and \cref{alg:adv_train} on 34 random points in a two-dimensional space for binary classification.
The algorithms were run with the parameters $P_s = 360$ and $\epsilon=0.08$.
A bias term was included by concatenating a column of ones to the data matrix $X$.
The decision boundaries shown in \cref{fig:2D-toy} confirm that \cref{alg:adv_train} fits the perturbation boxes as designed, coinciding with the theoretical prediction \citep[Figure 3]{madry2018towards}.
In \Cref{sec:more_2d_adv_exp}, we compare the decision boundaries of convex training and back-propagation methods, and discuss how the regularization strength $\beta$ affects the decision boundaries.
In \Cref{sec:adv_opt_landscape}, we compare the convex and the non-convex optimization landscapes and demonstrate robust certifications around the training data.

\subsubsection{Hinge Loss Convex Adversarial Training -- Image Classification} \label{sec:exp_adv_CIFAR}

We now verify the real-world performance of the proposed convex training methods on a subset of the CIFAR-10 image classification dataset \citep{cifar10} for binary classification between ``birds'' and ``ships''.
The subset consists of 600 images downsampled to $d = 7 \times 7 \times 3 = 147$.\footnote{
The parameters are $\epsilon=10/255$, $\beta = 10^{-4}$, and $P_s=36$, corresponding to an ANN width of at most 72.
}
We use clean data and adversarial data generated with FGSM and PGD to compare \cref{alg:train}, \cref{alg:adv_train}, traditional back-propagation standard training (abbreviated as GD-std), and the widely used adversarial training method: use FGSM or PGD to solve for the inner maximum of \cref{eq:hinge_adv} and use back-propagation to solve the outer minimization (abbreviated as GD-FGSM and GD-PGD).
The implementation details of FGSM and PGD are discussed in \Cref{sec:PGDdetails}.

\newcommand{\pc}{\small{\% }}
\begin{table}[t]
\centering
\caption{
Average optimal objective and accuracy on clean and adversarial data over seven runs on the CIFAR-10 dataset. The standard deviations across the runs are shown in parentheses.
}
\begin{small}
\begin{tabular}{lccccc}
\toprule
Method & Clean accuracy & FGSM adv. & PGD adv. & Objective & CPU Time (s) \\
\midrule
GD-std                 & $79.56\pc$ \tiny{$(.414\%)$} & $47.09\pc$ \tiny{$(.4290\%)$} & $45.60\pc$ \tiny{$(.4796\%)$} & $.3146$								& $108.4$ \\
GD-FGSM                & $75.30\pc$ \tiny{$(3.10\%)$} & $61.03\pc$ \tiny{$(4.763\%)$} & $60.99\pc$ \tiny{$(4.769\%)$} & $.8370$								& $154.9$ \\
GD-PGD                 & $76.56\pc$ \tiny{$(.604\%)$} & $62.48\pc$ \tiny{$(.2215\%)$} & $62.44\pc$ \tiny{$(.1988\%)$} & $.8220$								& $1764$  \\
\cref{alg:train}       & $81.01\pc$ \tiny{$(.809\%)$} & $.4857\pc$ \tiny{$(.1842\%)$} & $.3571\pc$ \tiny{$(.1239\%)$} & \footnotesize{$6.910\times 10^{-3}$}	& $37.77$ \\
\cref{alg:adv_train}   & $78.36\pc$ \tiny{$(.325\%)$} & $66.95\pc$ \tiny{$(.4564\%)$} & $66.81\pc$ \tiny{$(.4862\%)$} & $.6511$								& $1544$  \\
\bottomrule
\end{tabular}
\end{small}
\label{tbl:cifar}
\end{table}

\Cref{tbl:cifar} shows the results on our CIFAR-10 subset.
Convex standard training (\cref{alg:train}) achieves a higher clean accuracy and a much lower training loss than GD-std, supporting the findings of \cref{thm:prac}.
The non-robust convex-trained model is highly sensitive to adversarial perturbations.
This is because standard training has no control over the loss of the perturbed inputs, and the high optimization accuracy of convex training exacerbates this issue, making convex adversarial training (\Cref{alg:adv_train}) paramount.
As shown in \cref{tbl:cifar}, \Cref{alg:adv_train} achieves a higher accuracy on clean and adversarial data alike compared to GD-FGSM and GD-PGD.
While \cref{alg:adv_train} solves the upper-bound problem \cref{eq:hinge_adv_D}, it returns a lower training objective than GD-FGSM and GD-PGD, showing that back-propagation fails to find an optimal network.
In addition to achieving superior results and higher observed stability, \cref{alg:train} and \cref{alg:adv_train} are theoretically guaranteed to converge to their global optima, hence particularly suitable for safety-critical applications.

We also compare the aforementioned SDP relaxation adversarial training method \citep{raghunathan2018certified} and the LP relaxation method \citep{Wong18a} against our work on the CIFAR-10 subset.
While an iteration of the LP or the SDP method is faster than a GD-PGD iteration, the ANNs trained with the LP or SDP method achieve worse accuracy and robustness than those trained with \cref{alg:adv_train}: the LP method achieves a 74.05\% clean accuracy and a 58.65\% PGD accuracy, whereas the SDP method achieves 73.35\% on clean data and 40.45\% on PGD adversaries.\footnote{
For SDP, the robustness parameter is chosen as $\lambda = .04$, since a larger $\lambda$ causes the algorithm to fail.
}
These results support that \cref{alg:adv_train} trains more robust ANNs and that the LP and SDP relaxations can be extremely loose and unstable.
While \citep{raghunathan2018certified, Wong18a} applied the convex relaxation method to the adversarial training problem, their training formulations are non-convex.

The presence of an $\ell_1$ norm term in the upper-bound formulations \cref{eq:hinge_adv_D} and \cref{eq:convex_ce_adver} indicates that adversarial training with a small $\epsilon$ has a regularizing effect, which can improve generalization, supporting the finding of \citep{Kurakin17}.
In the above experiments, \cref{alg:adv_train} outperforms \cref{alg:train} on adversarial data, highlighting the contribution of \cref{alg:adv_train}: a novel convex adversarial training procedure that reliably trains robust ANNs.

\section{Concluding Remarks} \label{sec:conclu}

In this paper, we used the SCP theory to characterize the quality of the solution obtained from an approximation method, providing theoretical insights into practical convex training.
We then developed a separating scheme and applied the ADMM algorithm to a family of convex training formulations.
When combined with the approximation method, the algorithm achieves a quadratic per-iteration computational complexity and a linear convergence towards an approximate global optimum.
We also introduced a simpler unconstrained convex training formulation based on an SCP relaxation.
The characterization of its solution quality shows that ELMs are convex relaxations to ANNs.
Compared to traditional back-propagation, our training algorithms possess theoretical convergence rate guarantees and enjoy the absence of spurious local minima.
Compared with na\"ively solving the convex training formulation with general-purpose solvers, our algorithms have much-improved complexities, making a significant step towards practical convex training.

On the robustness side, we used the robust convex optimization analysis to derive convex programs that train adversarially robust ANNs.
Compared with traditional adversarial training methods, including GD-FGSM and GD-PGD, the favorable properties of convex optimization endow convex adversarial training with the following advantages:

\begin{itemize}[leftmargin=8mm]
    \setlength\itemsep{.2em}
    \item \textbf{Global convergence to an upper bound:} Convex adversarial training provably converges to an upper bound to the global optimum cost, offering superior interpretability.
    \item \textbf{Guaranteed adversarial robustness on training data:} As shown in \cref{thm:inner_max}, the inner maximization over the robust loss function is solved exactly. 
    \item \textbf{Hyperparameter-free:} \cref{alg:adv_train} can automatically determine its step size with line search, not requiring any preset parameters.
    \item \textbf{Immune to vanishing/exploding gradients:} The convex training method avoids this problem completely because it does not rely on back-propagation.
\end{itemize}

Overall, our analysis makes it easier and more efficient to train interpretable and robust ANNs with global convergence guarantees, facilitating safety-critical ANN applications.

\bibliographystyle{siamplain}
\bibliography{Papers}

\newpage
\appendix

\section*{Appendix}

\section{Extending the ADMM Approach to More Sophisticated ReLU Networks} \label{sec:ADMM_ext}

Since the emergence of convex training, convex formulations have been developed to train various types of neural networks.
Most formulations share the structure 
\begin{align} \label{eq:ct_general}
	\min_{\bw, \bw^\prime} \; & \ell \left( \mathbf{F} (\bw - \bw^\prime), y \right) + \beta \left( \norm{\bw}_{2,1} + \norm{\bw^\prime}_{2,1} \right) \\
	\ST \quad & \mathbf{G} \bw \geq 0, \quad \mathbf{G} \bw^\prime \geq 0 \nonumber,
\end{align}
where $\mathbf{F}$ and $\mathbf{G}$ are matrices formed by the training data matrix $X$ and those matrices that represent all possible ReLU activation patterns, $\norm{\cdot}_{2, 1}$ denotes the norm that is a mixture of the $\ell_2$ norm and the $\ell_1$ norm under some partition scheme, and $\bw$ and $\bw^\prime$ are the optimization variables from which the neural network weights can be recovered.

\cref{alg:ADMM} can be extended to all convex training formulations with this structure by first reforming the problem into the equality-constrained form
\begin{align} \label{eq:ct_eq_general}
	\min_{\bu, \bu^\prime, \bw, \bw^\prime, \bs, \bs^\prime} \; & \ell \left( \mathbf{F} (\bu - \bu^\prime), y \right) + \beta \left( \norm{\bw}_{F,1} + \norm{\bw^\prime}_{F,1} \right) + \sI_{\geq 0} (\bs) + \sI_{\geq 0} (\bs^\prime) \\
	\ST \quad & \begin{bmatrix} I \\ \mathbf{G} \end{bmatrix} \begin{bmatrix} \bu & \bu^\prime \end{bmatrix} = \begin{bmatrix} \bw & \bw^\prime \\ \bs & \bs^\prime \end{bmatrix} \nonumber
\end{align}
and constructing the augmented Lagrangian
\begin{align*}
	L (\bu, \bu^\prime, \bw, & \bw^\prime, \bs, \bs^\prime, \lambda, \lambda^\prime, \nu, \nu^\prime) := \nonumber \\[-4mm]
	& \ell \Big( \mathbf{F} (\bu - \bu^\prime), y \Big) + \beta \left( \begin{Vmatrix} \bw \\ \bw^\prime \end{Vmatrix}_{2,1} \right) + \sI_{\geq 0} \left( \begin{bmatrix} \bs \\ \bs^\prime \end{bmatrix} \right) + \frac{\rho}{2} \left( \begin{Vmatrix} \bu - \bw + \lambda \\ \bu^\prime - \bw^\prime + \lambda^\prime \\ \mathbf{G} \bu - \bs + \nu \\ \mathbf{G} \bu^\prime - \bs^\prime + \nu^\prime \end{Vmatrix}_2^2 - \begin{Vmatrix} \lambda \\ \lambda^\prime \\ \nu \\ \nu^\prime \end{Vmatrix}_2^2 \right),
\end{align*}
where $(\lambda, \lambda^\prime)$ and $(\nu, \nu^\prime)$ are again dual variables and $\rho > 0$ is a fixed penalty parameter.
Minimizing over $(\bw, \bw^\prime)$, $(\bu, \bu^\prime)$, and $(\bs, \bs^\prime)$ separately in an alternating manner and performing dual updates on $(\lambda, \lambda^\prime)$ and $(\nu, \nu^\prime)$ gives us an ADMM algorithm that tackles \cref{eq:ct_general}.

\subsection{Two-Hidden-Layer Sub-Networks}

We now discuss extending our methods to deeper and more practical ANN architectures.
In \cite{Ergen21a}, the authors have shown that training multiple two-hidden-layer ReLU sub-networks with a weight decay regularization is equivalent to solving a higher-dimensional convex problem with sparsity induced by group $\ell_1$ regularization.

Consider an architecture with $K$ parallel sub-networks, each of which is a two-hidden-layer ReLU network.
The neural network output can be parameterized as $\left( \left( X \bw_{1k} \right)_+ \bw_{2k} \right)_+ w_{3k}$, where $\bw_{1k} \in \R^{m_0 \times m_1}$, $\bw_{2k} \in \R^{m_1 \times m_2}$, $w_{3k} \in \R^{m_3}$ are the hidden and output layer weights for the $k^{\textrm{th}}$ sub-network.
Note that $m_0 = d$, whereas $m_1$ and $m_2$ denote the numbers of neurons in the first and the second hidden layer.
The regularized training problem is formalized as
\begin{align} \label{eq:nonconvextrainingdeep}
    \min_\theta \ell \big( \left( \left( X \bw_{1k} \right)_+ \bw_{2k} \right)_+ w_{3k}, y \big) + \frac{\beta}{2} \sum_{k=1}^K \left( \norm{\bw_{2k}} + w_{3k}^2 \right),
\end{align}
where $\beta > 0$ is a regularization parameter.
In \cite{Ergen21a}, it has been shown that the non-convex training problem (\ref{eq:nonconvextrainingdeep}) can be equivalently stated as the following convex problem:
\begin{align} \label{eq:convextrainingdeep}
    \min_{\bw, \bw^\prime} \; & \; \ell \big( \tilde{X} (\bw - \bw^\prime), y \big) + \beta \left( \norm{\bw}_{2,1} + \norm{\bw^\prime}_{2,1} \right) \nonumber \\ 
    \ST \; & \; \text{vec} \left( \begin{bmatrix} (2\mathbf{D}_{1ij}-I_n) X \\ (2 \mathbf{D}_{2l} - I_n) \mathbf{D}_{1ij} X \end{bmatrix} \begin{bmatrix} \bw_{ijl}^+ & \bw_{ijl}^- \end{bmatrix} \right) \geq 0, \quad \forall i \in [P_1],\ j \in [m_1],\ l \in [P_2], \\
    & \; \text{vec} \left( \begin{bmatrix} (2 \mathbf{D}_{1ij} - I_n) X \\ (2 \mathbf{D}_{2l} - I_n) \mathbf{D}_{1ij} X \end{bmatrix} \begin{bmatrix} \bw_{ijl}^{+^\prime} & \bw_{ijl}^{-^\prime} \end{bmatrix} \right) \geq 0, \quad \forall i \in [P_1],\ j \in [m_1],\ l \in [P_2], \nonumber
\end{align}
where
\begin{itemize}[leftmargin=8mm]
    \item The vectors $\bw$ and $\bw^\prime \in \sR^{2 d m_1 P_1 P_2}$ are constructed by concatenating \linebreak $\left\{ \{ \{ \{\bw_{ijl}^\pm\}_{i=1}^{P_1} \}_{j=1}^{m_1} \}_{l=1}^{P_2} \right\}_\pm$ and $\left\{ \{ \{ \{ {\bw_{ijl}^\pm}^\prime \}_{i=1}^{P_1} \}_{j=1}^{m_1} \}_{l=1}^{P_2} \right\}_\pm$, respectively;
    \item Consider all $\overline{\bw} \in \R^d$, $\bw_1 \in \R^{d \times m_1}$, and $\bw_2 \in \R^{m_1}$. $P_1$ denotes the total number of possible sign patterns of $X \overline{\bw}$ , and $P_2$ denotes the number of possible sign patterns of $(\mathbf{X W}_1)_+ \bw_2$;
    \item The fixed diagonal binary mask matrices $\mathbf{D}_{1ij} \in \R^{n\times n}$ and $\mathbf{D}_{2l} \in \R^{n\times n}$ with $i \in [P_1]$, $j \in [m_1]$, $l \in [P_2]$ encode all possible ReLU patterns;
    \item For a vector $\bu \in \R^{dP}$, the notation $\norm{\bu}_{2,1} := \sum_{i=1}^P \norm{\bu_i}_2$ denotes the $d$-dimensional group norm operator with $\bu_i$ being the $i^{\text{th}}$ $d$-dimensional partition of $\bu$;
	\item $\tilde{X}_s$ is defined as $\begin{bmatrix} \mathbf{D}_{21} \mathbf{D}_{111} X & \dots & \mathbf{D}_{2l} \mathbf{D}_{1ij} X & \dots & \mathbf{D}_{2P_2} \mathbf{D}_{1P_1m_1} X \end{bmatrix}$, and $\tilde{X}$ is defined as $\begin{bmatrix} \tilde{X}_s & 0 \\ 0 & \tilde{X}_s \end{bmatrix}$.
\end{itemize}

We observe that both the objective function and the constraint set of \cref{eq:convextrainingdeep} follow the same structure as \cref{eq:ct_general}, namely, the objective consists of a convex loss with $\ell_1$-$\ell_2$ regularization and the feasible set is defined by linear inequality constraints.
Specifically, \cref{eq:convextrainingdeep} can be represented in the equality-constrained form below:
\begin{align} \label{eq:3layerAL_eq}
    \min_{\bw, \bw^\prime, \bu, \bu^\prime, (\bs, \bs^\prime)_{ijl}} & \ell \big( \tilde{X} (\bu-\bu^\prime), y \big) + \beta \left( \norm{\bw}_{2,1} + \norm{\bw^\prime}_{2,1} \right) + \sum_{i, j, k} \left( \sI_{\geq0} ( \bs_i ) + \sI_{\geq0} ( \bs_i^\prime ) \right) \nonumber \\[-2mm]
    \ST \quad\ & \bu = \bw, \quad \bu^\prime = \bw^\prime, \nonumber \\
    & \bs_{ijl} = \text{vec} \left( \begin{bmatrix} (2\mathbf{D}_{1ij}-I_n) X \\ (2\mathbf{D}_{2l}-I_n) \mathbf{D}_{1ij} X \end{bmatrix} \begin{bmatrix} \bw_{ijl}^+ & \bw_{ijl}^- \end{bmatrix} \right), \quad \forall i \in [P_1],\ j \in [m_1],\ l \in [P_2], \\
    & \bs_{ijl}^\prime = \text{vec} \left( \begin{bmatrix} (2\mathbf{D}_{1ij}-I_n) X \\ (2\mathbf{D}_{2l}-I_n) \mathbf{D}_{1ij} X \end{bmatrix} \begin{bmatrix} \bw_{ijl}^{+^\prime} & \bw_{ijl}^{-^\prime} \end{bmatrix} \right), \quad \forall i \in [P_1],\ j \in [m_1],\ l \in [P_2], \nonumber
\end{align}
which is a special case of \cref{eq:ct_eq_general}.
The ADMM algorithm thus extends to \cref{eq:3layerAL_eq}, the convex training problem for architectures consisting of parallel two-hidden-layer ReLU networks.

The work \citep{Ergen21c} has similarly analyzed three-layer ReLU networks, but considers an alternative regularization technique --- path regularization.
Since the convex training formulation with path regularization also follows the structure of \cref{eq:ct_general}, our ADMM algorithm similarly applies.

\newpage
\subsection{One-Hidden-Layer Networks with Batch Normalization}

In \cite{ErgenBatch}, exact convex representations of weight-decay regularized ReLU networks with (full-batch) batch normalization (BN) have been introduced.
While \cite{ErgenBatch} provides discussions on training deeper neural networks with BN, the paper only presents convex training formulations for the one-hidden-layer case.
Consider a one-hidden-layer scalar-output ReLU network with the weights $\bw^{(1)} \in \R^{m_0 \times m_1}$ and $\bw^{(2)} \in \R^{m_1}$, where $m_0 = d$ is the input dimension and $m_1$ is the network width.
Let $X \in \R^{n \times d}$ denote the training data and $y \in \R^n$ be the label matrix.
The regularized training problem of this network with BN is given by 
\begin{align} \label{eq:BNregularizedtraining}
    \min_{\bw^{(1)}, \bw^{(2)}, \gamma, \alpha} \ell \left( \big( \textrm{BN}_{\gamma, \alpha} (X \bw^{(1)}) \big)_+ \bw^{(2)}, y \right) + \frac{\beta}{2} \left( \norm{\gamma}_2^2 + \norm{\alpha}_2^2 + \norm{\bw^{(1)}}_F^2 + \norm{\bw^{(2)}}_2^2 \right),
\end{align}
where $\ell$ is a convex loss function and $\textrm{BN}_{\gamma, \alpha}(\cdot)$ represents the BN operator associated with a scaling parameter $\gamma$ and a shifting parameter $\alpha$ \cite{BN}.
The non-convex training problem \cref{eq:BNregularizedtraining} can be equivalently cast as the convex optimization problem
\begin{align} \label{eq:BNregularizedtrainingconvex}
        \min_{\bw_i, \bw^\prime_i \in \R^{r+1}} & \ell \left( \sum_{i=1}^P D_i \bU^\prime_i \left( \bw_i - \bw^\prime_i \right), y \right) + \beta \sum_{i=1}^P \left( \norm{\bw_i}_2 + \norm{\bw^\prime_i}_2 \right) \\
    \ST & \quad (2 D_i - I_n) \bU^\prime \bw_i \geq 0, \; (2 D_i - I_n) \bU^\prime \bw^\prime_i \geq 0, \quad \forall i \in [P], \nonumber
\end{align}
where the diagonal matrices $D_1, \ldots, D_P$ represent all ReLU activation patterns associated with $X \bw$ for an arbitrary weight vector $\bw \in \R^d$ and $P$ denotes the cardinality of the set of all possible $\mathbf{D}$ matrices.
Furthermore, $\bU \in \R^{n\times r}$ and $\bU^\prime \in \R^{n\times (r+1)}$ are computed using the compact singular value decomposition (SVD) of the zero-mean data matrix, where $r = \textrm{rank} (X)$.
More specifically, $\left( I_n -\frac{1}{n} \mathbf{1} \mathbf{1}^\top \right) X = \bU \Sigma \mathbf{V}^\top$ and $\bU^\prime = [\bU \ \ \frac{1}{\sqrt{n}} \mathbf{1}]$ \cite{ErgenBatch}.

Note that \cref{eq:BNregularizedtrainingconvex} has the same structure as the convex reformulation of the standard one-hidden-layer ReLU network training problem \cref{eq:convex_general}.
The main difference is that in \cref{eq:BNregularizedtrainingconvex}, $\bU^\prime$ plays the role of the data matrix $X$.
As such, \cref{alg:ADMM} and the convex adversarial training analyses extend to the convex training formulation of ReLU networks with BN without modifications.

\section{SCP-Based Convex Training} \label{sec:SCP}

While the practical training formulation \cref{eq:prac_clean} and the ADMM algorithm (\cref{alg:ADMM}) vastly improve the efficiency and the practicality of globally optimizing ANNs, the complexity of the aforementioned methods can still be too high for large-scale machine learning problems due to the complicated structure of \cref{eq:convex_general}.
In this section, we propose a ``sampled convex program (SCP)''-based alternative approach to approximately globally optimize scalar-output one-hidden-layer ANNs.
This approach constructs scalable unconstrained convex optimization problems with simpler structures.
Unconstrained convex optimization problems are much easier to numerically solve compared to constrained ones.
Scalable and simple first-order methods can be easily applied to unconstrained convex programs, while the same cannot be said for constrained optimization in general due to feasibility issues.

Compared with the ADMM approach in \cref{alg:ADMM}, the SCP approach is easier to implement and has a lower per-iteration complexity.
The trade-off is that while \cref{alg:ADMM} can be applied to find the exact global minimum of \cref{eq:nonconvex_general} (albeit with an exponential complexity with respect to the data matrix rank), the SCP approach only finds an approximate global solution.
In the approximate case, the qualities of the ADMM solution and the SCP solution can both be characterized.

\subsection{One-Shot Sampling of Hidden-Layer Weights}

The paper \cite{Pilanci20a} has shown that the non-convex training formulation \cref{eq:nonconvex_general} has the same global optimum as
\begin{equation} \label{eq:noncvx2}
    p^\star = \min_{(u_j, \alpha_j)\jm} \ell \Big( \sum\jm (X u_j)_+ \alpha_j, y \Big) + \frac{\beta}{2} \sum\jm |\alpha_j| \quad \ST\quad \norm{u_j}_2 \leq 1,\ \forall j\in[m].
\end{equation}

Note that we can replace the perturbation set $\{ u \;|\; \norm{u}_2 \leq 1 \}$ with $\{ u \;|\; \norm{u}_2 = 1 \}$ without changing the optimum.
This is because for any pair $(u_j, \alpha_j)$ such that $\enorm{u_j} < 1$, replacing $(u_j, \alpha_j)$ with the scaled weights $(\frac{u_j}{\enorm{u_j}}, \enorm{u_j} \cdot \alpha_j)$ will reduce the regularization term of \cref{eq:noncvx2} while keeping the loss term unchanged.
Therefore, the optimal $u_j^\star$ must satisfy $\enorm{u_j^\star}=1$.

To approximate the semi-infinite program \cref{eq:noncvx2}, we randomly sample a total of $N$ vectors, namely $u_1, \dots, u_N$, on the $\ell_2$ unit norm sphere $\gS^{d-1}$ following a uniform distribution.
It is well-known that such a procedure can be performed by randomly sampling $\widehat{u}_i \sim \gN(0, I_d)$ for all $i\in[N]$ and projecting each $\widehat{u}_i$ onto the unit $\ell_2$ norm sphere by calculating $u_i = \frac{\widehat{u}_i}{\norm{\widehat{u}_i}_2}$ for all $i\in[N]$.
Next, $u_1, \dots, u_N$ are used to construct the following SCP:
\vspace{-2mm}
\begin{align} \label{eq:SCP2}
    p_{s3}^\star = \min_{(\alpha_i)\iN} \ell \Big( \sum\iN \big( X u_i \big)_+ \alpha_i, y \Big) + \beta \sum\iN |\alpha_i|,
    \vspace{-1mm}
\end{align}
where the sampled hidden-layer weights $(u_i)\iN$ are fixed.

The finite-dimensional unconstrained convex formulation \cref{eq:SCP2} is a relaxation of \linebreak \cref{eq:noncvx2}, and can be used as a surrogate for the optimization problem \cref{eq:nonconvex_general} to approximately globally optimize one-hidden-layer ANNs.
The formulation \cref{eq:SCP2} optimizes the ANN's output layer while freezing the hidden layer.
When the squared loss $\ell (\yhat, y) = \frac{1}{2} \enorm{\yhat-y}^2$ is considered, \cref{eq:SCP2} is a Lasso regression problem.
Intuitively, the sampled hidden-layer weights map the training data points into a higher-dimensional space.
While some of the sampled weights will inevitably be far from the optimum weights for the ANN, the $\ell_1$ regularization term promotes sparsity, encouraging assigning zero weights to ``disable'' the suboptimal hidden neurons.

The SCP training formulation \cref{eq:SCP2} recovers the training problems of one-hidden-layer random vector functional link (RVFL) \citep{RVFLbound} and ELM.
Such an equivalence shows that training an ELM is a convex relaxation of ANN training.
Compared with traditional ELMs, \cref{eq:SCP2} contains a sparsity-promoting regularization, and requires a different initialization of the untrained hidden layer weights, providing insights into the implicit sparsity-seeking property of ANNs.

The method in this subsection is referred to as ``one-shot sampling'' because all hidden layer weights are sampled in advance, in contrast with the iterative sampling procedure described in \Cref{sec:itersampling}.
The ANNs trained with \cref{eq:SCP2} can be suboptimal in terms of empirical loss compared with the network that globally minimizes the non-convex cost function, but are expected to be close to the optimal classifier.
The next theorem characterizes the level of suboptimality of the SCP optimizer, with the proof provided in \Cref{sec:SCPproof}.

\begin{theorem} \label{thm:SCP}
    Suppose that an additional hidden neuron $u_{N+1}$ is randomly sampled on the unit Euclidean norm sphere via a uniform distribution to augment the ANN.
    Consider the following formulation to train the augmented network:
    \begin{align} \label{eq:SCPaug}
        p_{s4}^\star = \min_{(\alpha_i)\iNN} \ell \Big( \sum\iNN \big( X u_i \big)_+ \alpha_i, y \Big) + \beta \sum\iNN |\alpha_i|.
    \end{align}
    It holds that $p_{s4}^\star \leq p_{s3}^\star$.
    Furthermore, if $N \geq \min \big\{ \frac{n+1}{\psi \xi} - 1, \frac{2}{\xi} ( n+1-\log\psi ) \big\}$, where $\psi$ and $\xi$ are preset confidence level constants between 0 and 1, then with probability no smaller than $1-\xi$, it holds that $\sP \{ p_{s4}^\star < p_{s3}^\star \} \leq \psi$.
\end{theorem}

Intuitively, this bound means that uniformly sampling another hidden layer weight $u_{N+1}$ on the unit norm sphere will not improve the training loss with high probability.
For a fixed level of suboptimality, the required scale of the SCP formulation \cref{eq:SCP2} has a linear relationship with respect to the number of training data points. 
Somewhat surprisingly, from the perspective of the probabilistic optimality, the bound provided by \cref{thm:SCP} is the same as the bound associated with \cref{alg:train} presented in \cref{thm:prac}, because both bounds are obtained via the SCP analysis framework.

The main advantage of the SCP-based training approach is that the unconstrained optimization \cref{eq:SCP2} is much easier and faster to solve than the constrained optimization \cref{eq:prac_clean}.
The iterative soft-thresholding algorithm (ISTA) \citep{Beck2009AFI} and its accelerated or stochastic variants can be readily applied to solve \cref{eq:SCP2}.
Specifically, ISTA converges at a linear rate if $\ell \big( \sum\iN (X u_i)_+ \alpha_i, y \big)$ is strongly convex over each $\alpha_i$, and converges at a $\gO(1/T)$ rate for weakly convex cases, where $T$ is the iteration count.
As a result, with the same amount of computational resources, one can solve \cref{eq:SCP2} with $N \gg P_s$, allowing for training wider networks (with stronger representation powers) within a reasonable amount of time.
Numerical experiments in \Cref{sec:SCP_exp} verify that the SCP relaxation \cref{eq:SCP2} can train larger-scale classifiers with a reasonable computing effort.

When $\ell(\cdot)$ is the squared loss, the SCP formulation \cref{eq:SCP2} evaluates to $\min_\alpha \enorm{H \alpha - y}^2 + \beta \norm{\alpha}_1$, where $H = \begin{bmatrix} (X u_1)_+ & \dots & (X u_N)_+ \end{bmatrix} \in \sR^{n \times N}$ and $\alpha = (\alpha_1, \dots, \alpha_N) \in \sR^N$.
The ISTA update is then $\alpha^+ = \prox_{\gamma_s\beta \norm{\cdot}_1} (\alpha - \gamma_s H^\top H \alpha + \gamma_s H^\top y)$, where $\prox_{\gamma_s\beta \norm{\cdot}_1} (\cdot)$ evaluates to $\sgn(\cdot) \max(|\cdot|-\gamma_s\beta, 0)$, $\alpha^+$ denotes the updated $\alpha$ at each iteration, and $\gamma_s$ is a step size that can be determined with backtracking line search.
Since $H^\top H$ and $H^\top y$ are fixed and only need to be calculated once, the per-iteration complexity is $\gO(N^2)$.
Since $N$ is linear in $n$ for a fixed solution quality (see \cref{thm:SCP}), the per-iteration complexity amounts to $\gO(n^2)$, and the overall complexity amounts to $\gO(n^2 \log (1/\epsilon_a))$ and $\gO(n^2 / \epsilon_a)$ for strongly and weakly convex loss functions, respectively, where $\epsilon_a$ is the desired optimization precision.

\cref{thm:prac} also implies that when the neural network is wide, the hidden layer weights are less important than the output layer weights.
The role of the hidden layers is to map the data to features in higher-dimensional spaces, facilitating the output layer to extract the most important information.

\subsection{Iterative Sampling of Hidden-Layer Weights} \label{sec:itersampling}

While the efficacy of SCP-based convex training with a one-shot sampling of the hidden layer neurons can be proved theoretically and experimentally, the probabilistic optimality bound provided in \cref{thm:SCP} may be too conservative in some cases.
To provide a more accurate and robust estimation of the level of suboptimality of the SCP relaxation \cref{eq:SCP2}, we propose a scheme (\cref{alg:scheme}) that iteratively samples hidden layer neurons used in \cref{eq:SCP2} to train classifiers.

The convex semi-infinite training formulation \cref{eq:noncvx2} has a dual problem: \citep[Appendix A.4]{Pilanci20a}
\begin{align} \label{eq:UCP}
    d^\star = \max_{v\in\R^n} -\ell^*(v) \quad \ST \quad |v^\top (X u)_+| \leq \beta, \;\; \forall u: \norm{u}_2 \leq 1,
\end{align}
where $\ell^*(\cdot)$ is the Fenchel conjugate function defined as $\ell^*(v) = \max_z z^\top v - \ell(z, y)$.
When $m \geq m^*$, where $m^*$ is upper-bounded by $n+1$, strong duality holds $p^\star = d^\star$.
Moreover, the dual problem \cref{eq:UCP} is a convex semi-infinite problem, which is a category of uncertain convex programs (UCP) \citep{Calafiore2005}.

We then use the sampled vectors $u_1, \dots, u_N$ to construct the following SCP that approximates the UCP \cref{eq:UCP}:
\begin{align} \label{eq:SCP}
    d_{s3}^\star = \max_{v\in\R^n} -\ell^*(v) \quad \ST \quad & |v^\top (X u_i)_+| \leq \beta, \;\; \forall i\in[N].
\end{align}

Similarly, strong duality holds between \cref{eq:SCP} and \cref{eq:SCP2} and it holds that $p_{s3}^\star = d_{s3}^\star$.
The level of suboptimality of the dual solution $v^\star$ to \cref{eq:SCP} can be easily verified by checking the feasibility of $v^\star$ to the UCP \cref{eq:UCP}.

While it is easier to check the quality of the dual solution, it is desirable to solve the primal problem \cref{eq:SCP2} because the primal is unconstrained and thus easier to solve.
Suppose that $(\alpha_i^\star)\iN$ is a solution to \cref{eq:SCP2}.
By following the procedure described in \Cref{sec:SCPdual}, one can recover the optimal dual variable $v^\star$ from $(\alpha_i^\star)\iN$ by exploiting the strong duality between \cref{eq:SCP2} and \cref{eq:SCP}.
Next, we independently sample another set of $N_1$ hidden layer weights $(u_i^1)\iNo \sim \text{Unif}(\gS^{n-1})$ and check if $| v^{\star\top} (X u_i^1)_+ | > \beta$ for each $i\in[N_1]$.
If $| v^{\star\top} (X u_i^1)_+ | > \beta$ for a particular $i$, then adding $u_i^1$ to the set of sampled constraint set of \cref{eq:SCP} will change (reduce) the value of $d_{s3}^\star$ and thereby reduce the relaxation gap between $p_{s3}^\star$ and $p^\star$.
In other words, by incorporating $u_i^1$ as another hidden layer node, the considered ANN can be improved.

Define the notations
\vspace{-1mm}
\begin{gather*}
    Z_i \coloneqq \begin{cases} 1 & \text{if } | v^{\star\top} (X u_i^1)_+ | > \beta \\ 0 & \text{otherwise}
    \end{cases}, \text{ for all } \forall i\in[N_1], \\
    \Zbar \coloneqq \frac{\sum\iNo Z_i}{N_1}, \quad \text{and} \quad
    \theta \coloneqq \E_{u \sim \text{Unif}(\gS^{d-1})} [Z_i] = \sP_{u \sim \text{Unif} (\gS^{d-1})} \Big[ |v^{\star\top} (X u)_+| > \beta \Big].
\end{gather*}

By Hoeffding's inequality, it holds that $\sP \big( \theta - \overline{Z} \geq t \big) \leq \exp(-2 N_1 t^2)$.
Therefore, with probability at least $1-\xi$, it holds that $\theta \leq \Zbar + \frac{\log (1/\xi)}{2 N_1}$, where $\xi\in(0,1]$.
In other words, by evaluating the feasibility of the additional set of hidden layer weights $u_1^1 \dots u_{N_1}^1$, one can obtain a probabilistic bound on the level of suboptimality of the solution to \cref{eq:SCP} constructed with $u_1 \dots u_N$: as long as $\Zbar + \frac{\log (1/\xi)}{2 N_1} \leq \psi$ for a constant $\psi \in (0, 1]$, it holds that $\theta \leq \psi$ with probability at least $1-\xi$.

We now introduce a scheme of training scalar-output fully connected ReLU ANNs to an arbitrary degree of suboptimality by repeating the evaluation and sampling procedure, described in \cref{alg:scheme}.
Let $T$ denote the total iterations of \cref{alg:scheme}, $U_t$ denote the total number of hidden layer neurons at iteration $t$, and $N_t$ denote the number of hidden layer neurons sampled at iteration $t$.
In light of \cref{thm:SCP}, it holds that the solution $(\alpha^\star_i)\iUT$ yielded by \cref{alg:scheme} satisfies the following property with probability at least $1-\xi$: if an additional vector $\tilde{u}$ is sampled on the unit Euclidean norm sphere $\gS^{d-1}$ via a uniform distribution, then adding $\tilde{u}$ to the set of hidden layer weights used in \cref{eq:SCP2} will not improve the training loss of the ANN with probability at least $1-\psi$.

\begin{algorithm}[t]
    \begin{algorithmic}[1]
        \State Let $t=0$; sample $\widehat{u}^0_1, \dots, \widehat{u}^0_{N_0} \sim \gN(0,I_d)$ i.i.d., and let $u^0_i = \frac{\widehat{u}^0_i}{\norm{\widehat{u}^0_i}_2}$ for all $i\in[N_0]$.
        \State Construct $\gU^0 \coloneqq \{u^0_1, \dots, u^0_{N_0}\}$; let $U_0 = N_0$.
        \Repeat
            \State Solve $(\alpha_i^t)\iUt = \argmin_{(\alpha_i)\iUt} \ell \big( \sum\iUt \big( X u^t_i \big)_+ \alpha_i, y \big) + \beta \sum\iUt |\alpha_i|$, the same formulation as \cref{eq:SCP2}.
            \State Update $v^t = y - \sum\iUt (X u_i)_+ \alpha_i^t$.
            \State Sample $\widehat{u}^{t+1}_1, \dots, \widehat{u}^{t+1}_{N_{t+1}} \sim \gN(0,I_d)$ i.i.d., and let $\bar{u}^{t+1}_i = \frac{\widehat{u}^{t+1}_i}{\norm{\widehat{u}^{t+1}_i}_2}$ for all $i\in[N_{t+1}]$. 
            \State Construct $\gE^{t+1} = \big\{ \bar{u}_i^{t+1} \ \big|\ |v^{t\top} (X \bar{u}_i^{t+1})_+| > \beta \big\}$ to be the set of newly sampled weight vectors that tighten the dual constraint.
            \State Construct $\gU^{t+1} = \gU^t \cup \gE^{t+1}$ and rename all vectors in $\gU^{t+1}$ as $u_1^{t+1}, \dots, u_{U_{t+1}}^{t+1}$, where $U_{t+1}$ is the cardinality of $\gU^{t+1}$.
            \State $t \leftarrow t+1$.
            \vspace{1mm}
        \Until{$\frac{|\gE^t|}{N_t} + \frac{\log(1/\xi)}{2 N_t} \leq \psi$ or/and $U_{t-1} \geq \frac{n+1}{\psi\xi}-1$, where $\psi$ and $\xi$ are preset thresholds.}
    \end{algorithmic}
    \caption{Convex ANN training based on iterative sampling hidden-layer weights}
    \label{alg:scheme}
\end{algorithm}

\section{Additional Experiments}

\subsection{ADMM Asymptotic Convergence} \label{sec:ADMMasym}

In this part of the appendix, we present empirical evidences that demonstrate the asymptotic convergence properties of ADMM (\cref{alg:ADMM}).
We use the same data as in \Cref{sec:ADMMconv}, and the experiment settings are presented in \Cref{sec:ADMMparam}.

\begin{figure}
    \centering
    \begin{subfigure}{\textwidth}
        \adjustbox{trim={.065\width} {.01\height} {.075\width} {.05\height},clip} {\includegraphics[width=1.16\textwidth]{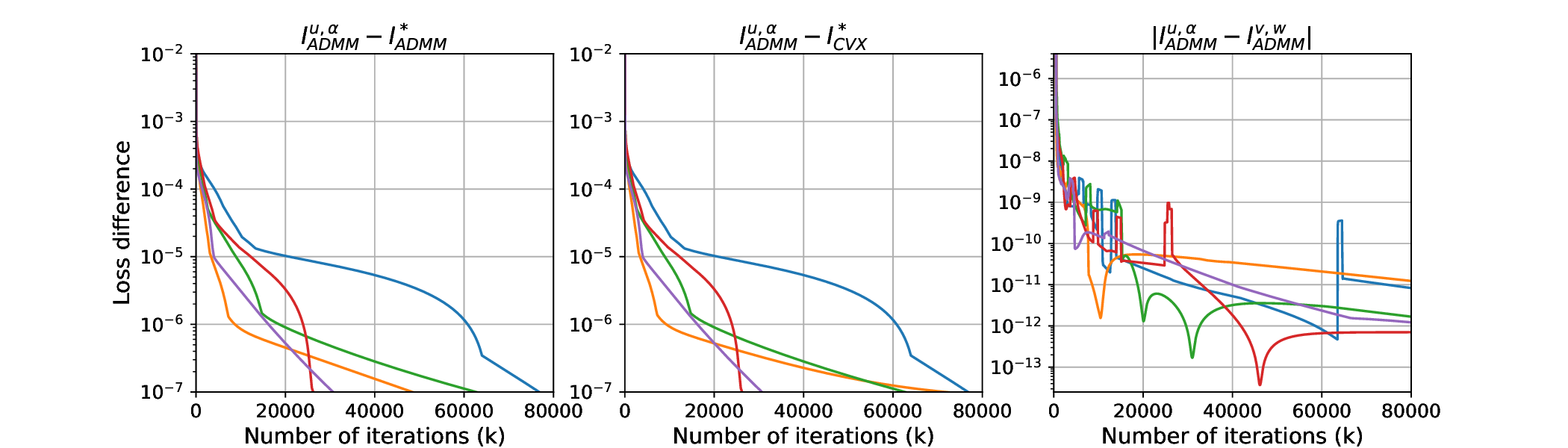}}
        \vspace{-3mm}
        \caption{$\lua - l^\star_{\text{ADMM}} \hspace{10cm}$}
        \label{fig:ADMMConv1_a}
    \end{subfigure} \\[-11.5mm]
    \hspace{51mm}
    \begin{subfigure}{.31\textwidth}
        \caption{$\lua - l^\star_{\text{CVX}}$}
        \label{fig:ADMMConv1_b}
    \end{subfigure}
    \begin{subfigure}{.31\textwidth}
        \caption{$\big| \lua - \lvw \big|$}
        \label{fig:ADMMConv1_c}
    \end{subfigure}
    \caption{Gap between the cost returned by ADMM at each iteration and the true optimal cost for five independent runs.}
    \label{fig:ADMMConv1}
\end{figure}

\cref{fig:ADMMConv1_a} shows that the training loss converges to a stationary value at a linear rate, verifying the findings of \cref{thm:ADMM}.
Note that the $D_h$ matrices randomly generated in the five runs are different, resulting in different optimization landscapes and different linear convergence bounds.
\cref{fig:ADMMConv1_b} shows that ADMM converges towards the CVX ground truth, verifying the correctness of the ADMM solution.
\cref{fig:ADMMConv1_c} shows that $\lvw$ and $\lua$ are close throughout the ADMM iterations, implying that $v_i$ and $w_i$ violate the constraints of \cref{eq:convex_general} insignificantly at every step.
Together, these figures confirm that the ADMM algorithm optimizes \cref{eq:nonconvex_general} effectively as designed.
The learning curves of the five runs look quite different because different random $D_h$ matrices can make the optimization landscape quite different.
However, as illustrated in \cref{fig:ADMMConv2}, the initial rapid convergence behavior is very consistent.

\subsection{The SCP Convex Training Formulation} \label{sec:SCP_exp}

\begin{figure}
    \centering
    \begin{subfigure}{.493\textwidth}
        \centering
        \includegraphics[width=\textwidth,trim={0 .5mm .2mm 5mm},clip] {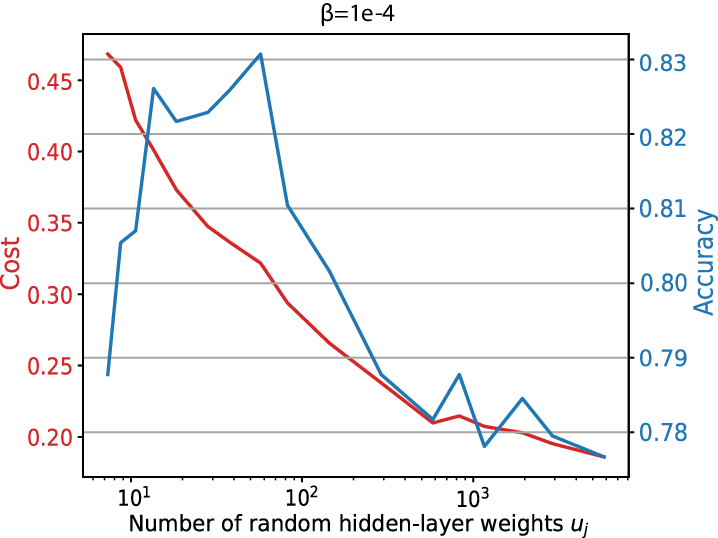}
        \vspace{-5mm}
        \caption{$\beta = 10^{-4}$}
        \label{fig:SCPexp1}
    \end{subfigure}
    \begin{subfigure}{.497\textwidth}
        \centering
        \includegraphics[width=\textwidth,trim={0 .5mm .2mm 5mm},clip] {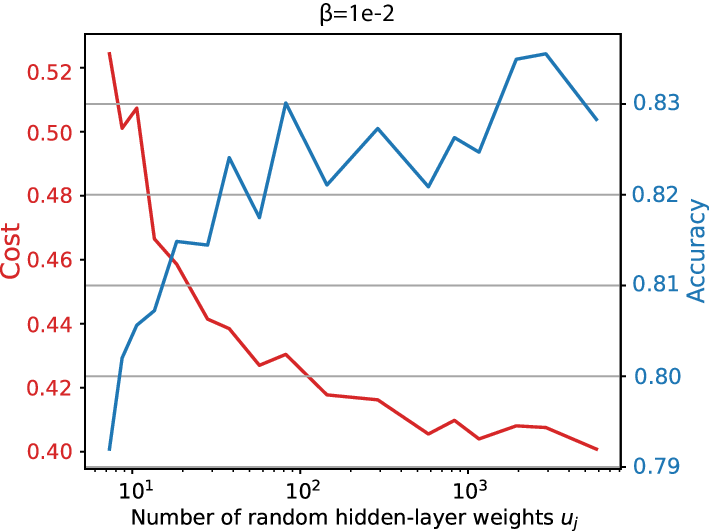}
        \vspace{-5mm}
        \caption{$\beta = 10^{-2}$}
        \label{fig:SCPexp2}
    \end{subfigure}
    \vspace{-5mm}
    \caption{Average accuracy and average cost with different choices of $N$ for two different selections of the regularization strength $\beta$.}
    \label{fig:SCPexp}
    \vspace{-2mm}
\end{figure}

In this subsection, we demonstrate the efficacy of the SCP relaxed training using the one-shot random sampling approach to choose $u_1, \dots, u_N$ and explore the effect of the number of sampled weights $N$.
We independently sample different numbers of hidden-layer-weights and use the SCP training formulation \cref{eq:SCP2} to train ANNs on the ``mammographic masses'' dataset \citep{Dua:2019}.
We remove instances containing NaNs and randomly select 70\% of the data for the training set and 30\% for the test set, resulting in $n=581$ and $d=5$.
We use two different regularization strengths: $\beta = 10^{-4}$ and $\beta = 10^{-2}$.
The training loss and the test accuracy of each $N$ setting are plotted in \cref{fig:SCPexp}.
The ANN training process is stochastic due to the randomly generated hidden-layer weights $u_j$ and the random splitting of training and test sets.
We use CVXPY and the MOSEK solver to solve the underlying optimization problem \cref{eq:SCP2}.
We perform 20 independent trials for each $N$ and average the results.

For both regularization settings, adding more sampled hidden layer weights makes the SCP approximation more refined and therefore decreases the training loss. When the regularization strength $\beta$ is $10^{-4}$, the test accuracy increases, peaks, and then decreases as $N$ increases. The accuracy drops when $N$ is large, possibly because of the overfitting caused by a lack of sparsity. As a comparison, training ANNs using \cref{alg:train} with $P_s$ set to 120 achieves an average accuracy of $79.80\%$ and an average training loss of $0.2428$ on the same dataset. Directly optimizing the non-convex cost function \cref{eq:nonconvex_general} using gradient descent back-propagation with the width $m$ set to $2 P_s = 240$ achieves an $81.14\%$ average test accuracy and a $0.3560$ average cost. Thus, with a proper choice of $N$, the prediction performance of the SCP convex training approach is on par with \cref{alg:train} and traditional back-propagation SGD. When the regularization strength $\beta$ is $10^{-2}$, the test accuracy of the ANNs trained with the SCP method generally increases with $N$.

To verify the performance of the proposed training approach on larger-scale data, we use the SCP method to train ANNs on the MNIST handwritten digits database \citep{lecun2010mnist} for binary classification between digits ``2'' and ``8''  ($d=784$ and $n=11809$) using the binary cross-entropy loss. The SCP training formulation \cref{eq:SCP2} is solved with the ISTA algorithm \citep{Beck2009AFI}. With the number of sampled weights $N$ set to 39365 (a much larger value than $P_s$ in the ADMM experiments, corresponding to an optimality level of $\xi\psi=0.3$), the SCP formulation \cref{eq:SCP2} achieves a test accuracy of $99.45\%$. Compared with the ADMM approach discussed in \Cref{sec:ADMM}, the SCP formulation is able to train much wider ANNs with a similar amount of computational power. In summary, this result demonstrates the performance and efficiency advantage of the SCP formulation \cref{eq:SCP2} for medium or large machine learning problems.

\subsection{Hinge Loss Convex Adversarial Training -- The Optimization Landscape} \label{sec:adv_opt_landscape}

This subsection shows that the convex loss landscape and the non-convex landscape overlap within an $\ell_\infty$-norm-bounded additive perturbation set around a training point $x_k$, and thereby verifies that the convex objective \cref{eq:rob_gen_cvx_1} provides an exact certification of the non-convex loss function at training data points.

\begin{figure}
    \centering
    \begin{subfigure}{.48\textwidth}
        \centering
        \adjustbox{trim={.04\width} {.04\height} {.03\width} {.02\height},clip} {\includegraphics[width=1.06\textwidth]{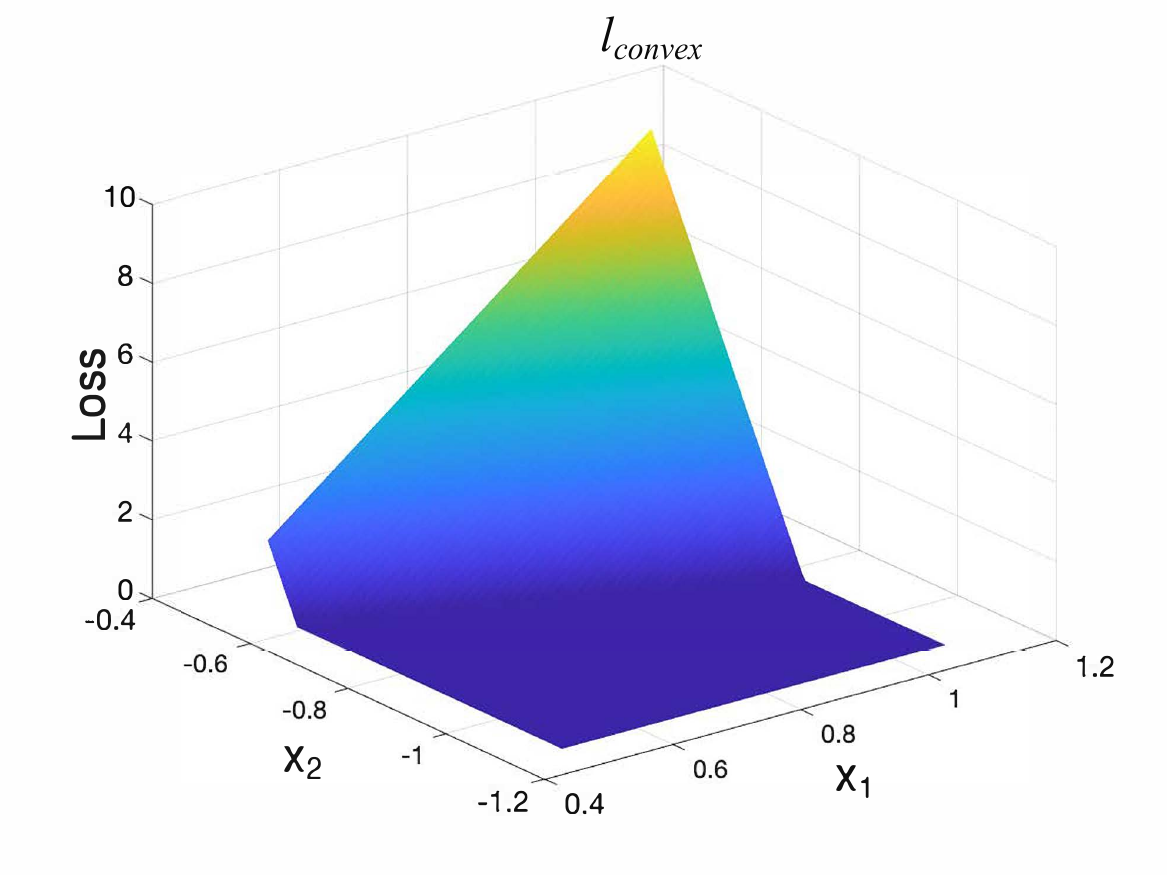}}
        \vspace{-2mm}
        \caption{The loss landscape of the convex objective $\ell_{\text{convex}}$ for $\norm{\delta}_\infty \leq 0.3$.}
        \label{fig:landscapes_a}
    \end{subfigure} \;
    \begin{subfigure}{.48\textwidth}
        \centering
        \adjustbox{trim={.04\width} {.04\height} {.03\width} {.02\height},clip} {\includegraphics[width=1.06\textwidth]{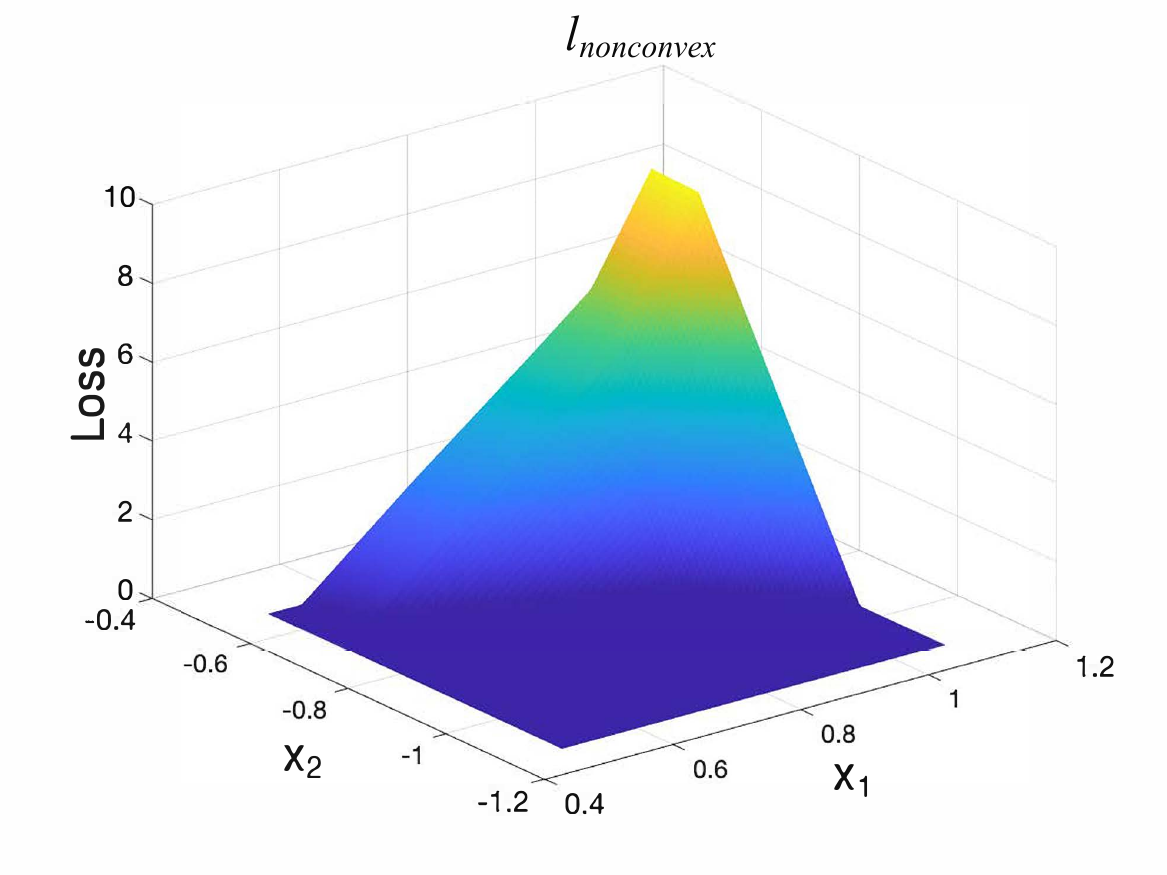}}
        \vspace{-2mm}
        \caption{The loss landscape of the non-convex objective $\ell_{\text{nonconvex}}$ for $\norm{\delta}_\infty \leq 0.3$.}
        \label{fig:landscapes_b}
    \end{subfigure}
    
    \begin{subfigure}{.48\textwidth}
        \centering
        \adjustbox{trim={.04\width} {.04\height} {.03\width} {.02\height},clip} {\includegraphics[width=1.06\textwidth]{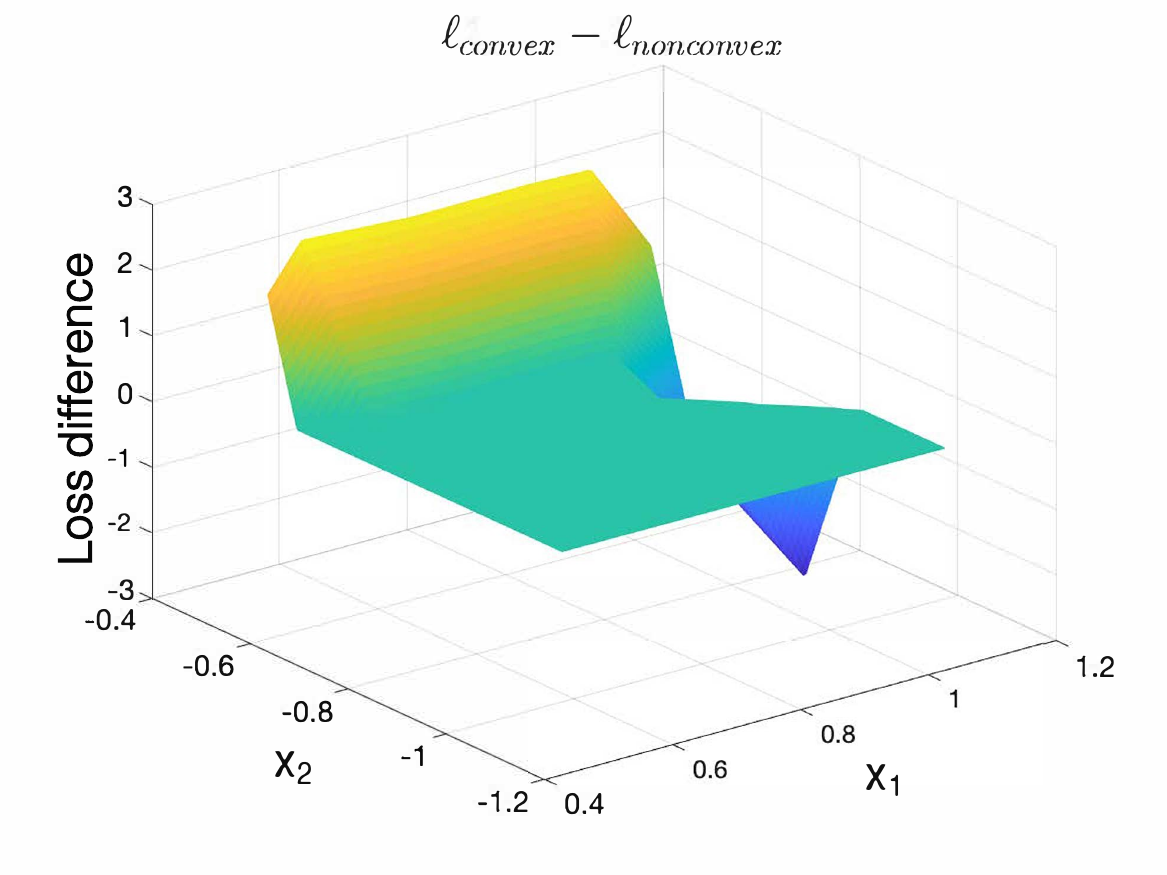}}
        \vspace{-2mm}
        \caption{$\ell_{\text{convex}} - \ell_{\text{nonconvex}}$ for $\norm{\delta}_\infty \leq 0.3$.}
        \label{fig:landscapes_c}
    \end{subfigure} \;
    \begin{subfigure}{.48\textwidth}
        \centering
        \adjustbox{trim={.04\width} {.04\height} {.03\width} {.02\height},clip} {\includegraphics[width=1.06\textwidth]{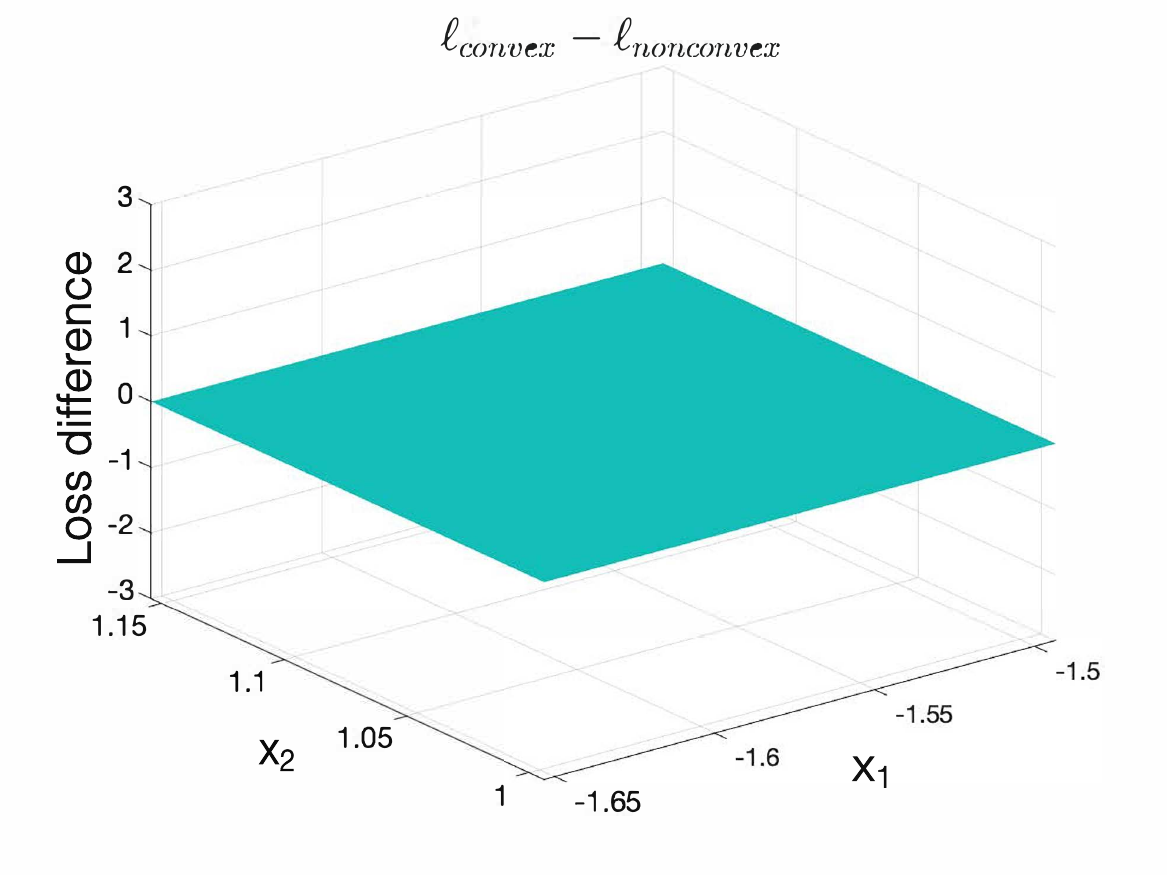}}
        \vspace{-2mm}
        \caption{$\ell_{\text{convex}} - \ell_{\text{nonconvex}}$ zoomed into $\norm{\delta}_\infty \leq 0.08$.}
        \label{fig:landscapes_d}
    \end{subfigure}
    \vspace{-2mm}
    \caption{Illustrations of the optimization landscapes of the convex and non-convex training formulations.}
    \label{fig:landscapes}
    \vspace{-3mm}
\end{figure}

The visualizations are based on the 2-dimensional experiment described in \Cref{sec:AT_exp_2D}. We use \cref{alg:adv_train} to train a robust ANN on the 2-dimensional dataset with $\epsilon = 0.08$, $P_s = 360$, and $\beta = 10^{-9}$. We then randomly select one of the training points $x_k$ and plot the loss around $x_k$ for the convex objective \cref{eq:rob_gen_cvx_1} and the non-convex objective \cref{eq:robust_general}. Specifically, for $\norm{\delta}_\infty \leq 0.3$, we plot
\begin{equation*}
    \ell_{\text{convex}} = \Big( 1 - y_k \cdot \sum\iP d_{ik} (x_k+\delta)^\top (v_i^\star - w_i^\star) \Big) 
    \text{ and } 
    \ell_{\text{nonconvex}} = \Big( 1 - y_k \cdot \sum\jm \big( (x_k+\delta)^\top u^\star_j \big)_+ \alpha^\star_j \Big),
\end{equation*}
where $d_{ik}$ is the $k\th$ entry of $D_i$, $y_k$ is the training label corresponding to $x_k$. Moreover, $v_i^\star$, $w_i^\star$ are the optimizers returned by \cref{alg:adv_train} and $u_j^\star$ and $\alpha_j^\star$ are the ANN weights recovered from $v_i^\star$ and $w_i^\star$ with \cref{eq:recover_weights}. The plots are shown in \cref{fig:landscapes_a}, \ref{fig:landscapes_b}.

For a clearer visualization, we also plot $\ell_{\text{convex}} - \ell_{\text{nonconvex}}$ in \cref{fig:landscapes_c} and zoom in to the $\ell_\infty$ norm ball with radius $\epsilon = 0.08$ in \cref{fig:landscapes_d}. When $\ell_{\text{convex}} - \ell_{\text{nonconvex}}$ is zero, the convex objective provides an exact certificate for the non-convex loss function. \cref{fig:landscapes_d} shows that for $\norm{\delta}_\infty \leq 0.08$, the difference is zero, supporting the finding that for ANNs trained with \cref{alg:adv_train}, the convex objective offers an exact certificate around the training points.

\subsection{Hinge Loss Convex Adversarial Training with Different Regularizations} \label{sec:more_2d_adv_exp}

\begin{figure}
    \centering
    \includegraphics[trim={7mm 6mm 1cm 0}, clip, width=.244\textwidth]
    {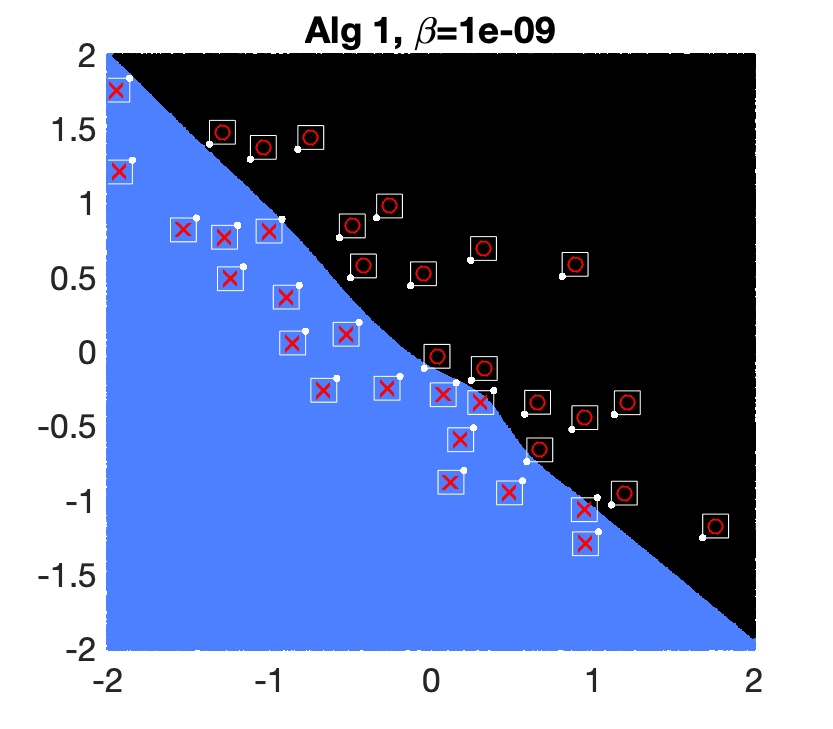}
    \includegraphics[trim={7mm 6mm 1cm 0}, clip, width=.244\textwidth]
    {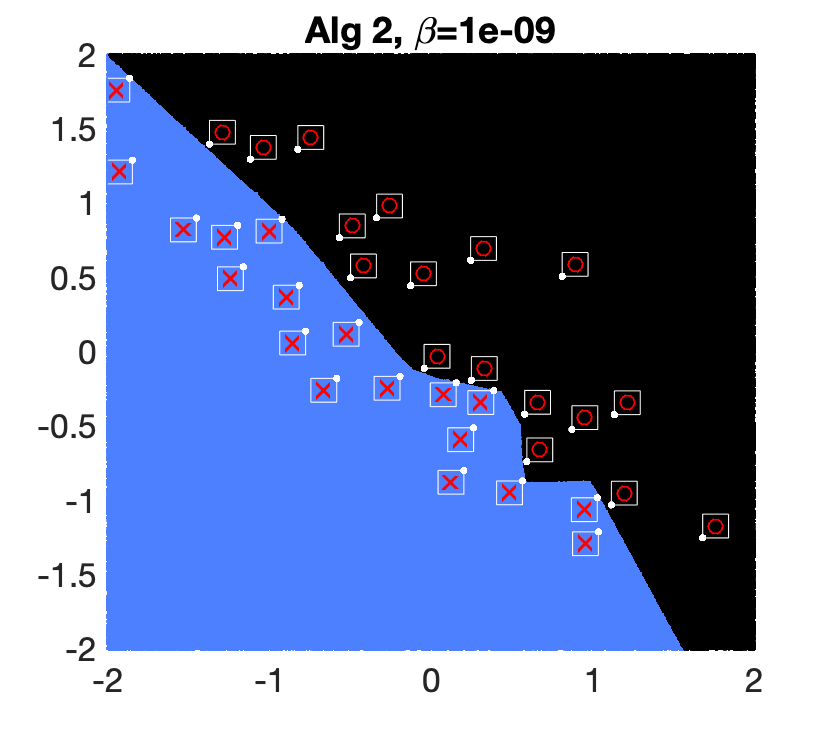}
    \includegraphics[trim={7mm 6mm 1cm 0}, clip, width=.244\textwidth]
    {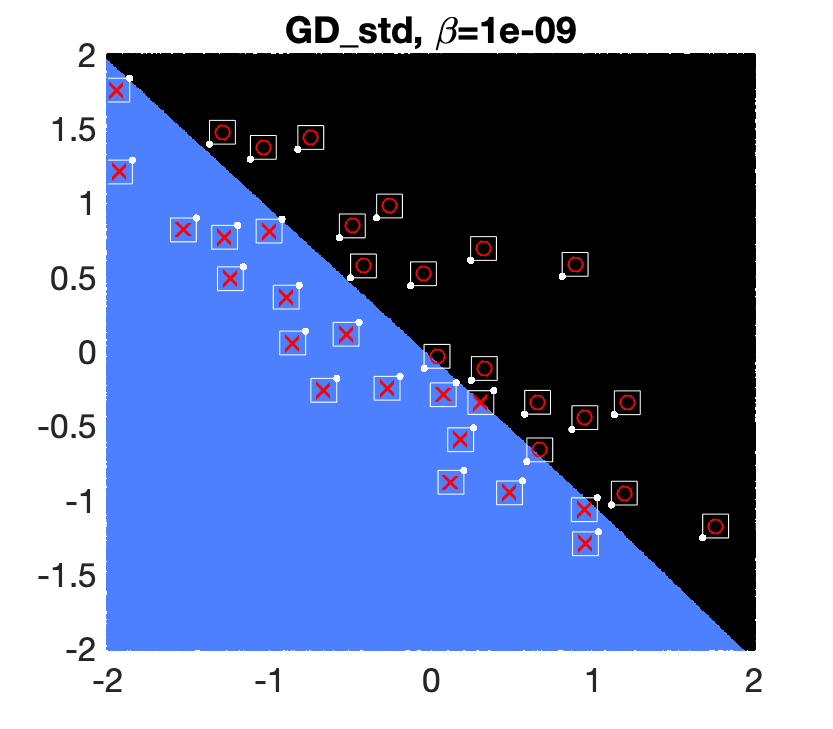}
    \includegraphics[trim={7mm 6mm 1cm 0}, clip, width=.244\textwidth]
    {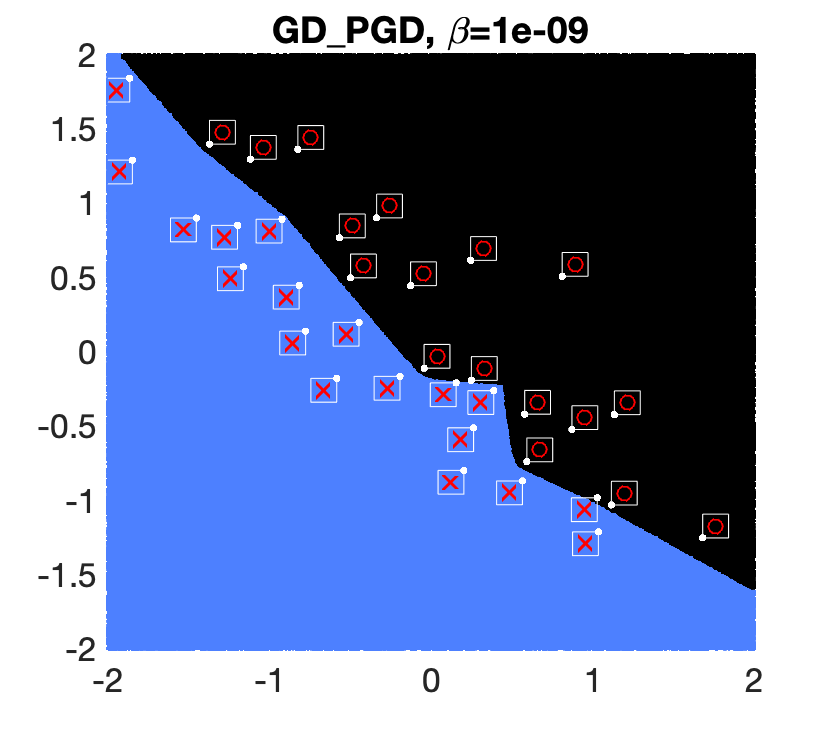}
    \vspace{1mm}
    \includegraphics[trim={7mm 6mm 1cm 0}, clip, width=.244\textwidth]
    {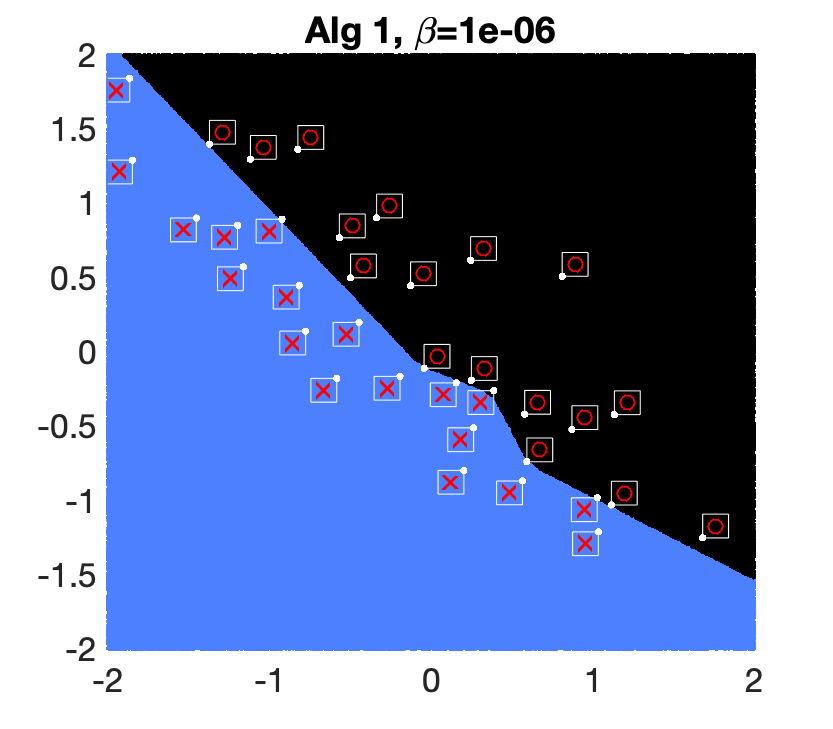}
    \includegraphics[trim={7mm 6mm 1cm 0}, clip, width=.244\textwidth]
    {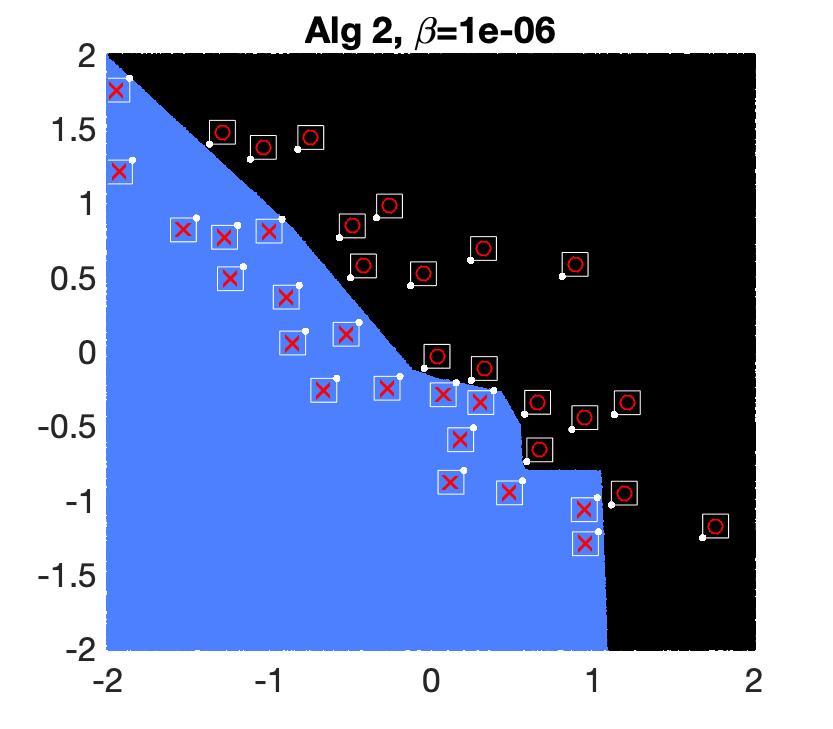}
    \includegraphics[trim={7mm 6mm 1cm 0}, clip, width=.244\textwidth]
    {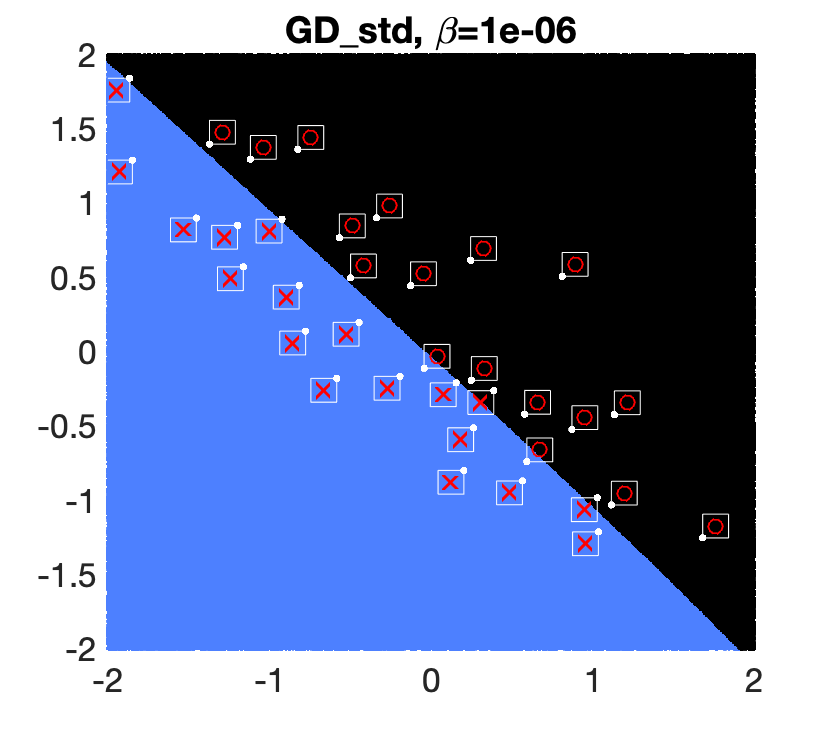}
    \includegraphics[trim={7mm 6mm 1cm 0}, clip, width=.244\textwidth]
    {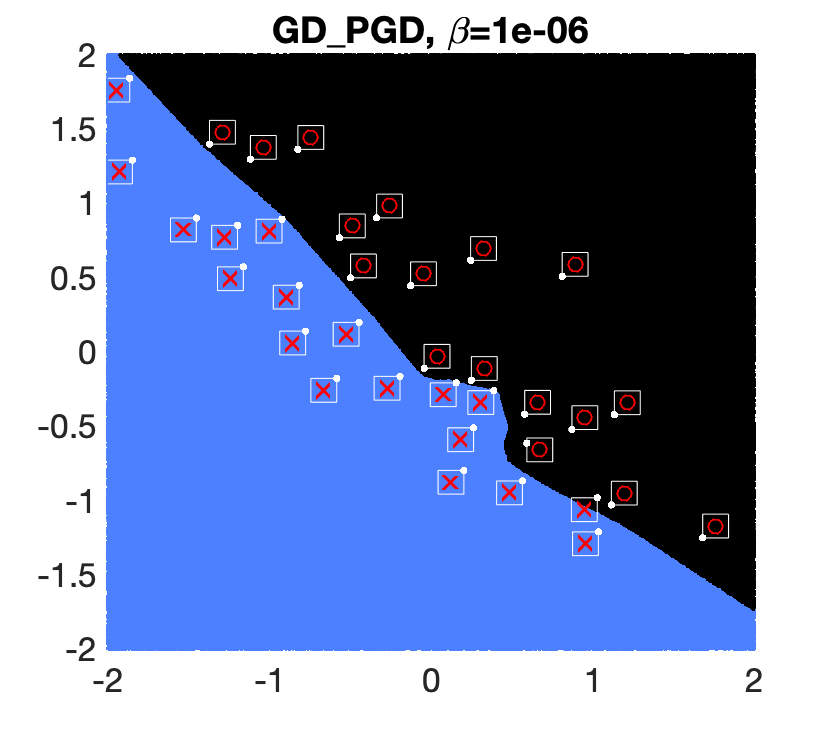}
    \vspace{1mm}
    \includegraphics[trim={7mm 6mm 1cm 0}, clip, width=.244\textwidth]
    {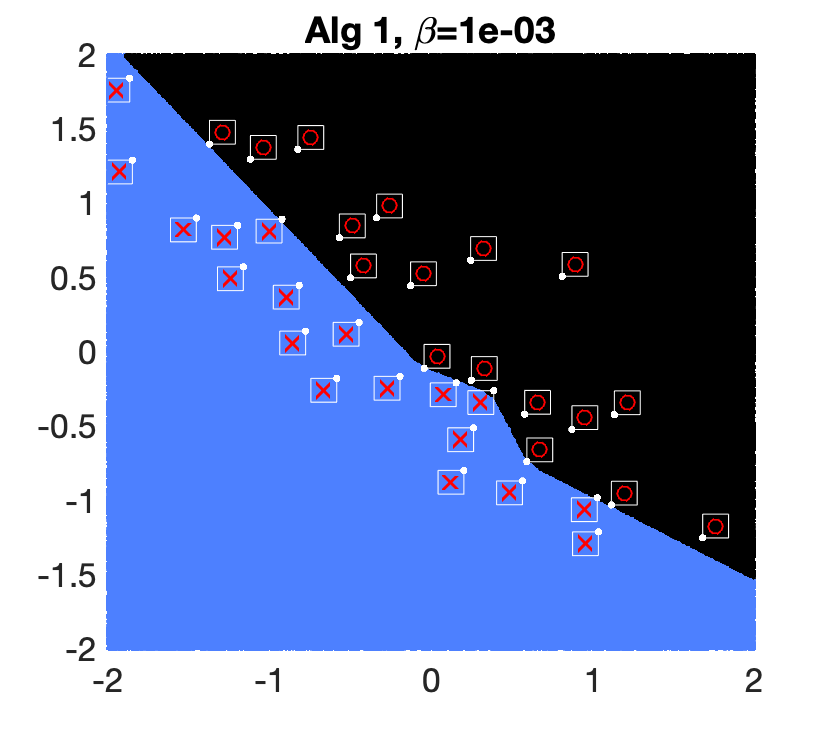}
    \includegraphics[trim={7mm 6mm 1cm 0}, clip, width=.244\textwidth]
    {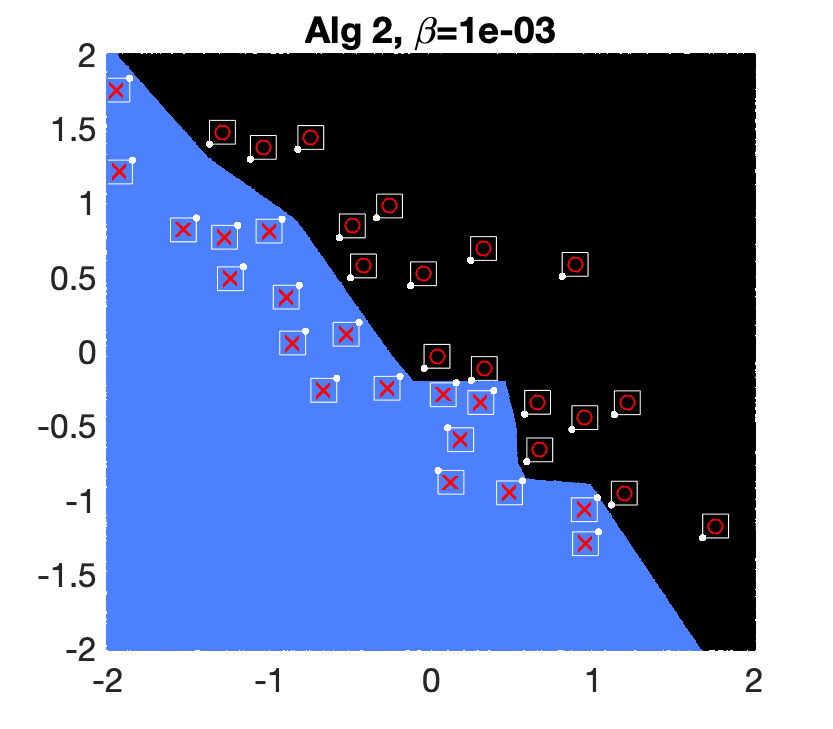}
    \includegraphics[trim={7mm 6mm 1cm 0}, clip, width=.244\textwidth]
    {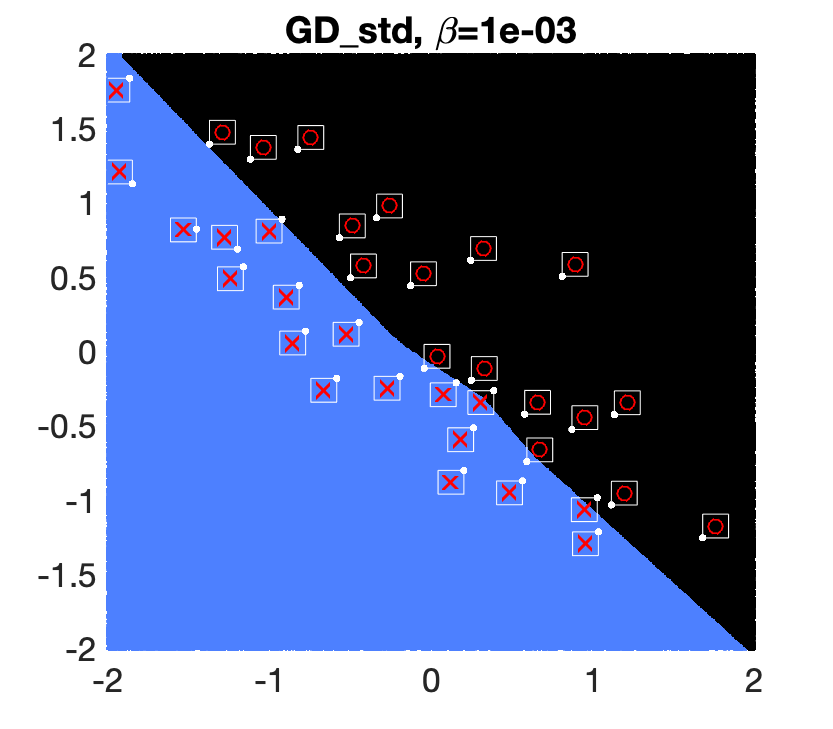}
    \includegraphics[trim={7mm 6mm 1cm 0}, clip, width=.244\textwidth]
    {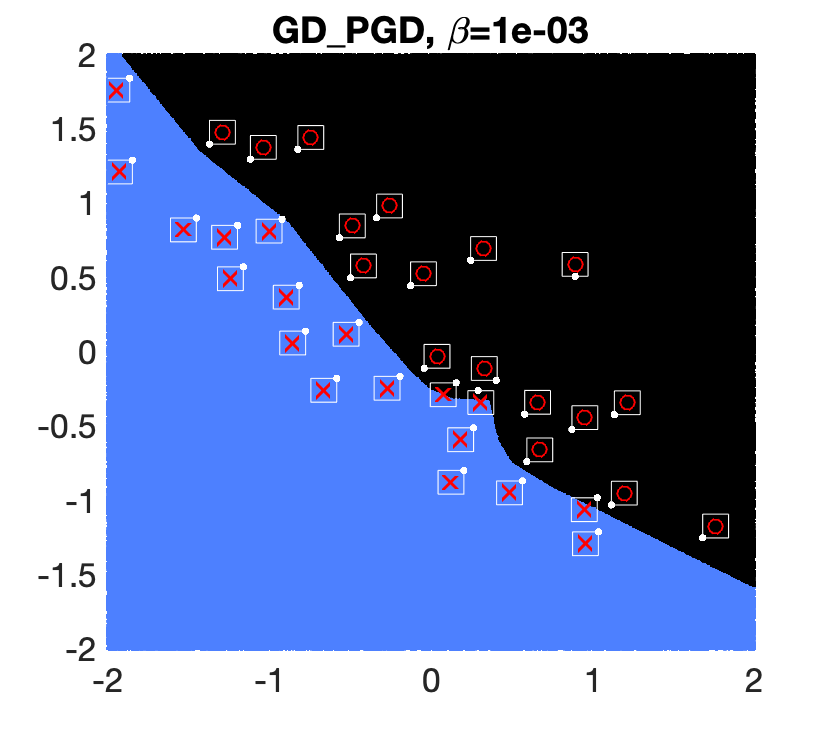}
    \vspace{1mm}
    \includegraphics[trim={7mm 6mm 1cm 0}, clip, width=.244\textwidth]
    {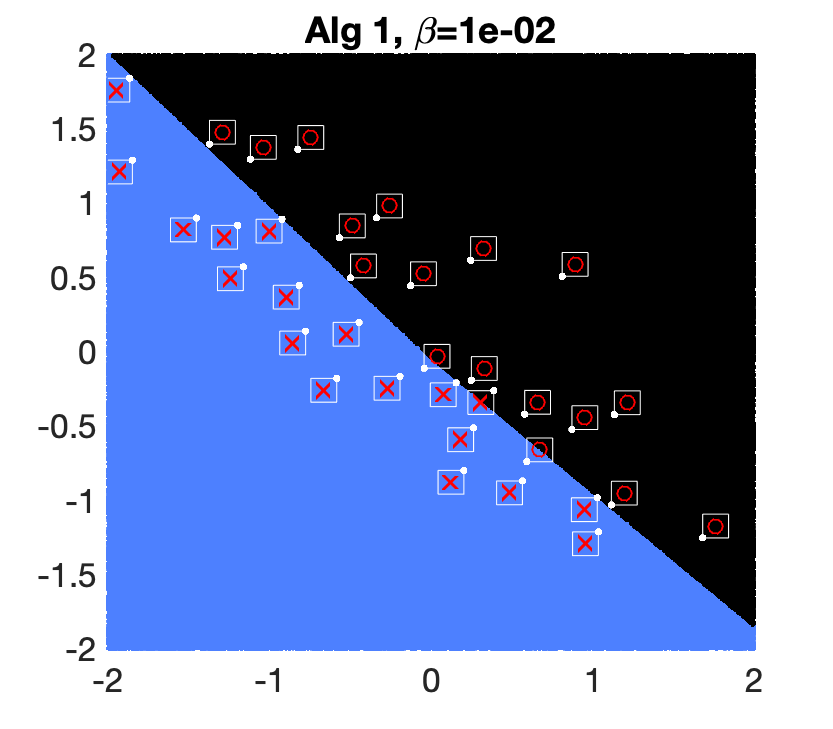}
    \includegraphics[trim={7mm 6mm 1cm 0}, clip, width=.244\textwidth]
    {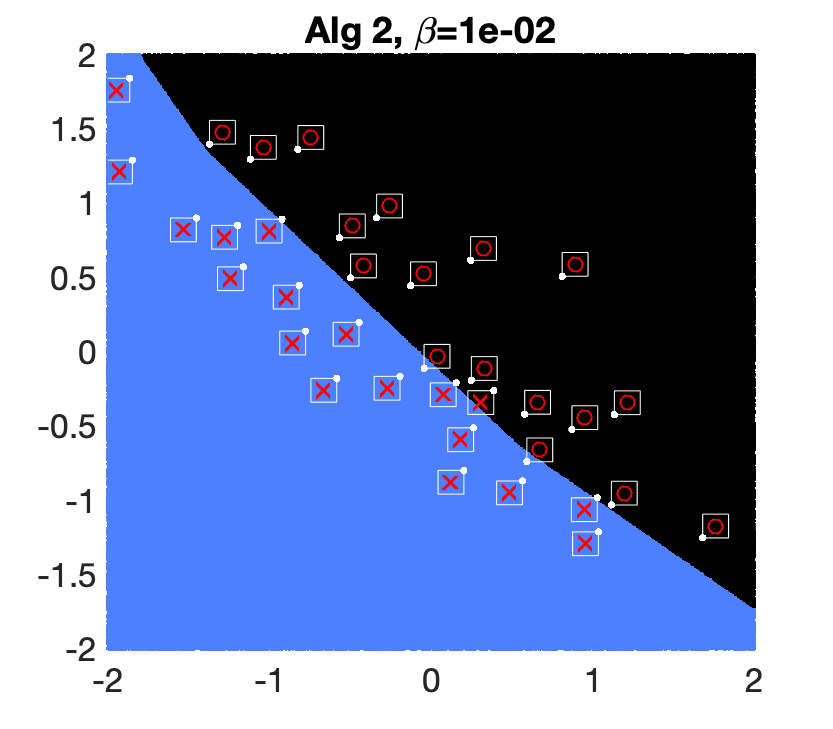}
    \includegraphics[trim={7mm 6mm 1cm 0}, clip, width=.244\textwidth]
    {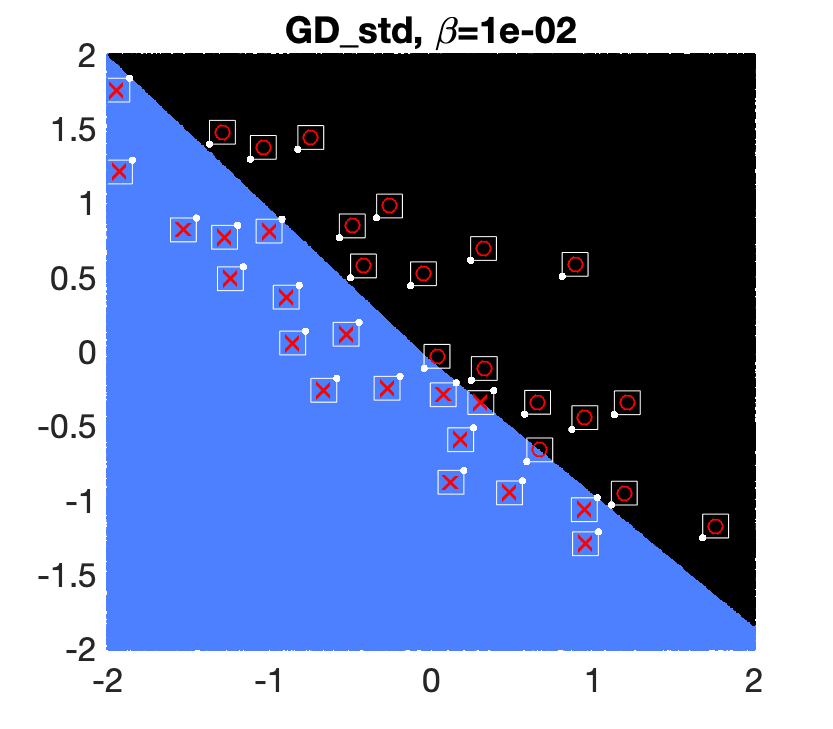}
    \includegraphics[trim={7mm 6mm 1cm 0}, clip, width=.244\textwidth]
    {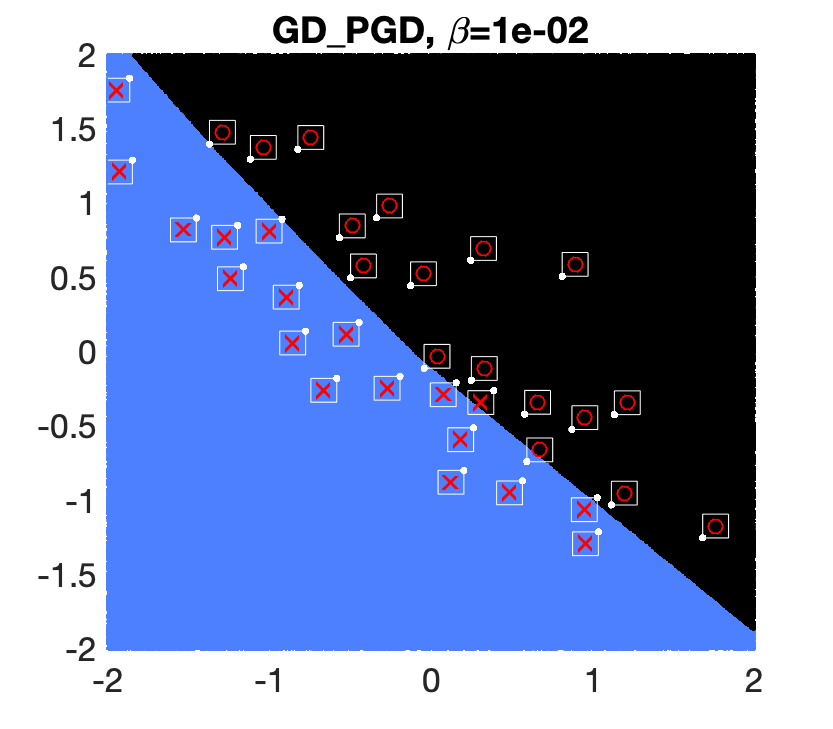}
    \caption{Decision boundaries obtained from various methods with $\beta$ set to $10^{-9}$, $10^{-6}$, $10^{-3}$, and $10^{-2}$.}
    \label{fig:7}
    \vspace{-3mm}
\end{figure}

We now compare the decision boundaries obtained from the convex training algorithms and back-propagation algorithms. As shown in \cref{fig:7}, the two standard training methods (\cref{alg:train} and GD-std) learned decision boundaries that separated the training points but failed to separate the perturbation boxes. Note that \cref{alg:train} learned slightly more sophisticated boundaries while GD-std learned near-linear boundaries that were very close to one of the positive training points \textcolor{red}{$\times$}.

The convex adversarial training method given by \cref{alg:adv_train} learnes boundaries that separate all perturbation boxes when $\beta$ was $10^{-3}$, $10^{-6}$, or $10^{-9}$. This behavior matches the theoretical illustration of adversarial training \citep[Figure 3]{madry2018towards}, and verifies that \cref{alg:adv_train} works as intended. When the regularization is too strong ($\beta = 10^{-2}$), the robust boundary becomes smoothed out and very similar to the standard training boundaries.
The traditional adversarial training method GD-PGD learns boundaries that separate most perturbation boxes. However, the boundaries cut through the box at around $(1, -1)$ when $\beta$ is $10^{-3}$, $10^{-6}$, or $10^{-9}$. This behavior is likely caused by GD-PGD's worse convergence due to the non-convexity. When $\beta$ is too large, the GD-PGD boundary also becomes smoothed out.

\subsection{Squared Loss Convex Adversarial Training} \label{sec:sl_simulation}
\begin{figure}[t] %
    \centering
    \begin{minipage}{.47\textwidth}
      \centering
      \adjustbox{trim={.08\width} {.01\height} {.08\width} {.01\height},clip} {\includegraphics[width=1.14\textwidth]{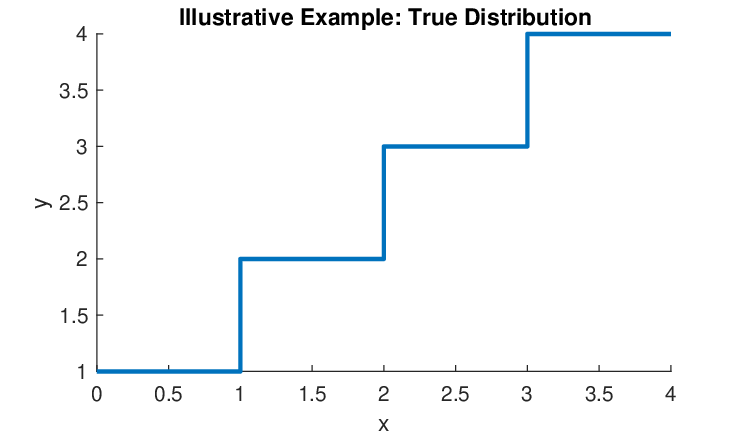}}
      \vspace{-1mm}
      \captionof{figure}{The true relationship between the data $x$ and the targets $y$ used in the illustrative example in \Cref{sec:sl_simulation}. The training ($n=8$ points) and test ($n=100$ points) sets are uniformly sampled from the distribution.}
      \label{fig:te}
    \end{minipage} \;
    \begin{minipage}{.505\textwidth}
      \centering
      \adjustbox{trim={.08\width} {.01\height} {.08\width} {.01\height},clip} {\includegraphics[width=1.17\textwidth]{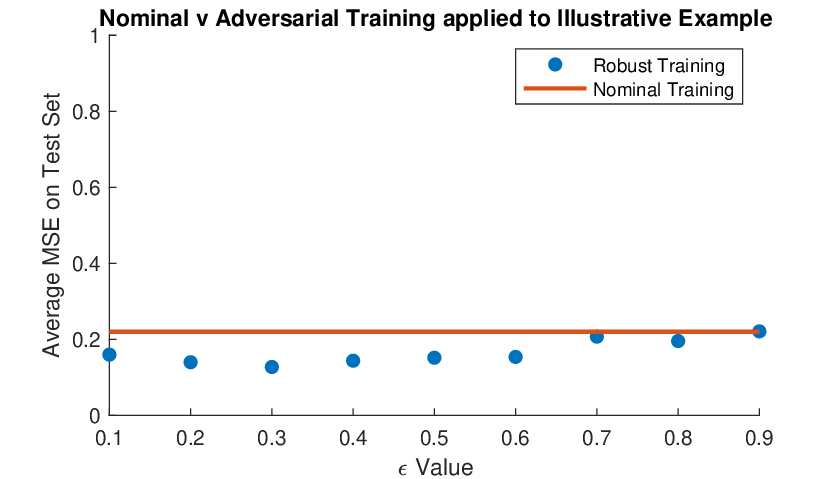}}
      \captionof{figure}{The robust training approach \cref{eq:SOCP3} outperforms the standard approach for different $\eps\in\{0.1, ..., 0.9\}$ on the dataset studied in \Cref{sec:sl_simulation}.}
      \label{fig:exp_sl}
    \end{minipage}
    \vspace{-2mm}
\end{figure}

The performance of the proposed robust optimization problem \cref{eq:SOCP3} is compared with the standard training problem \cref{eq:convex_general} on an illustrative 1-dimensional dataset.
\cref{fig:te} shows the true relationship between the data vector $X$ and the target output $y$. Training data are constructed by uniformly sampling eight points from this distribution, and test data are constructed by uniformly sampling 100 points. A bias term is included by concatenating a column of ones to $X$.

The training and test procedure are repeated for 100 trials with convex standard training (\cref{alg:train}). For convex adversarial training (\cref{alg:adv_train}), we varied the perturbation radius $\epsilon = 0.1, \dots, 0.9$. The training and test procedure was carried out for ten trials for each $\epsilon$. \cref{fig:exp_sl} reports the average test mean square error (MSE) for each setup.

The adversarial training procedure outperforms standard training for all $\epsilon$ choices. We further observe that the average MSE is the lowest at $\epsilon \approx 0.3$. This behavior arises as the robust problem attempts to account for all points within the uncertainty interval around the sampled training points. When $\epsilon$ is too small, the robust problem approaches the standard training problem. Larger values of $\epsilon$ cause the uncertainty interval to overestimate the constant regions of the true distribution, increasing the MSE.

\section{Experiment Setting Details}

\subsection{ADMM Hyperparameters} \label{sec:ADMMparam}

\begin{table}[!tb]
    \centering
    \caption{Hyperparameter settings used for the ADMM experiments.}
    \label{tbl:ADMMhyp}
    \begin{small}
    \begin{tabular}{l|c|c|c|c|c|c|c}
        \toprule
        & \cref{fig:ADMMConv1} & \cref{fig:ADMMConv2} & \cref{fig:ADMMAcc} & \cref{fig:ADMMCom} & \cref{tbl:ADMMMnist} & \cref{tbl:ADMMFMnist} & \cref{tbl:ADMMRBCD} \scriptsize{(ADMM-RBCD)} \\
        \midrule
        $\rho$     & 0.4    & 0.4    & 0.1    & 0.1    & 0.1   & 0.4   & 0.01  \\
        $\gamma_a$ & 0.01   & 0.4    & 0.1    & 0.1    & 0.1   & 0.16  & 0.01  \\
        $\beta$    & 0.0005 & 0.0005 & 0.0005 & 0.0001 & 0.001 & 0.001 & 0.001 \\
        \bottomrule
    \end{tabular}
    \end{small}
\end{table}

The proposed ADMM algorithm has two hyperparameters: a penalty hyperparameter $\rho$ and a step size $\gamma_a$.
The hyperparameters used in the experiments in this paper are shown in \cref{tbl:ADMMhyp}.
In most experiments, we select $\gamma_a = \rho$, a common choice for the ADMM algorithm.
The penalty parameter $\rho$ controls the level of infeasibility of $v$ and $w$.
Note that while ADMM guarantees to converge to an optimal feasible solution, the optimization variables may be infeasible in intermediate steps.
The feasibility of $v_i$ and $w_i$ to \cref{eq:convex_general} is emphasized when $\rho$ is large, while a low objective value is emphasized when $\rho$ is small.
For the purpose of finding optimal $u_j$ and $\alpha_j$ that minimize \cref{eq:nonconvex_general}, a balance between feasibility and low objective is required.
In practice, if there exists a significant gap between the objective of \cref{eq:convex_general} and the training loss \cref{eq:nonconvex_general}, then $\rho$ should be increased.
If the objective of \cref{eq:convex_general} struggles to reduce, then $\rho$ should be decreased.

\subsection{FGSM and PGD Details} \label{sec:PGDdetails}

The hinge loss has a flat part with zero gradient. To generate adversarial examples even in this part, we treat it as the ``leaky hinge loss'' via the model $\max \{ \zeta (1 - \hat{y} \cdot y), 1 - \hat{y} \cdot y \}$, where $\zeta \to 0^+$. Hence, the PGD update \cref{eq:PGD} amounts to
\begin{equation*} \label{eq:PGD_hinge}
    \widetilde{x}^{t+1} = \Pi_{\gX} \textstyle \Big( \tilde{x}^t - \gamma_p \cdot \sgn \big( y \cdot \sum_{j:\ x^\top u_j \geq 0} (u_j \alpha_j) \big) \Big), \quad \widetilde{x}^0 = x.
\end{equation*}
where the projection step can be performed by clipping the coordinates that deviate more than $\epsilon$ from $x$. In the following experiments, we use $\gamma_p = \epsilon/30$ and run PGD for $T = 40$ steps. On the other hand, the FGSM calculation can again be regarded as a special case of PGD where $T = 1$.

\newpage
\section{Convex Adversarial Training Extensions}

\subsection{Convex Squared Loss Adversarial Training} \label{sec:sqr_loss}

The squared loss $\ell(\widehat{y}, y) = \frac{1}{2} \enorm{\widehat{y}-y}^2$ is another commonly used loss function in machine learning. Consider the non-convex training problem of a one-hidden-layer ReLU ANN trained with the $\ell_2$-regularized squared loss:
\begin{align}\label{nonconvex_sl}
    \min_{(u_j, \alpha_j)\jm} \frac{1}{2} \bigg|\bigg| \sum\jm (X u_j)_+ \alpha_j - y & \bigg|\bigg|_2^2 + \frac{\beta}{2}\sum\jm \big( \norm{u_j}_2^2+\alpha_j^2 \big).
\end{align}

Coupling this nominal problem with the perturbation set $\gX$ gives us the robust counterpart as
\begin{align} \label{eq:robustnonconvex_sl}
    & \min_{(u_j, \alpha_j)\jm} 
    \begin{pmatrix}
        \displaystyle \max\allDelta \frac{1}{2} \bigg|\bigg| \sum\jm \big( (X+\Delta)u_j \big)_+ \alpha_j - y \bigg|\bigg|_2^2 + \frac{\beta}{2} \sum\jm \big( \norm{u_j}_2^2+\alpha_j^2 \big)
    \end{pmatrix}.
\end{align}

Applying \cref{THM:CVX_MINIMAX} and \cref{CORO:ROB_CONSTRAINT} leads to the following formulation as an upper bound on \cref{eq:robustnonconvex_sl}:
\begin{align} \label{eq:sl_minimax_convex}
    \min_{(v_i, w_i)\iPhat} & 
    \begin{pmatrix}
        \displaystyle \max\allDelta \frac{1}{2} \Bigg|\Bigg|\sum_{i=1}^{\widehat{P}} D_i (X+\Delta)(v_i-w_i) - y \Bigg|\Bigg|_2^2 + \beta \sum\iPhat \big( \norm{v_i}_2 + \norm{w_i}_2 \big) \hfill
    \end{pmatrix} \\[-.5mm]
    \ST \quad & (2 D_i - I_n) X v_i \geq \epsilon \norm{v_i}_1, \quad (2 D_i - I_n) X w_i \geq \epsilon \norm{w_i}_1, \quad \alliinPhat. \nonumber
\end{align}

Solving the maximization over $\Delta$ in closed form leads to the next result, with the proof provided in \Cref{sec:ROBUST_SOCP}.

\begin{theorem} \label{thm:robust_SOCP}
    The optimization problem \cref{eq:sl_minimax_convex} is equivalent to the convex program:
    \begin{align}\label{eq:SOCP3}
        & \min_{(v_i, w_i)\iPhat, a, z} a + \beta\sum\iPhat (\enorm{v_i}+\enorm{w_i}) \\
        & \quad\;\; \ST \quad (2 D_i - I_n) X v_i \geq \epsilon \norm{v_i}_1, \quad
        (2 D_i - I_n) X w_i \geq \epsilon \norm{w_i}_1, \quad \alliinPhat \nonumber \\[-1mm]
        & \qquad \qquad\ z_k\geq \bigg| \sum\iPhat D_{ik} x_k^\top (v_i-w_i) - y_k \bigg| \nonumber + \epsilon \biggnorm{ \sum\iPhat D_{ik}(v_i-w_i) }_1, \quad \forall k\in[n] \nonumber \\
        & \qquad \qquad\ z_{n+1}\geq \big| 2a-\tfrac{1}{4} \big|, \;\; \enorm{z}\leq 2a + \tfrac{1}{4} \nonumber.
   \end{align}
\end{theorem}

Problem \cref{eq:SOCP3} is a convex optimization that can train robust ANNs. However, directly using \cref{eq:SOCP3} for adversarial training can be intractable due to the large number of constraints that arise when we include all $D_i$ matrices associated with all $\Delta$ such that $X+\Delta \in \gX$. To this end, one can use the approximation in \cref{alg:adv_train} and sample a subset of the diagonal matrices $D_1, \dots, D_{P_s}$. As before, the optimality gap can be characterized with \cref{thm:prac}.

\subsection{Convex Adversarial Training for ConvNets} \label{sec:otherANN}

While our discussions explicitly focus on one-hidden-layer scalar-output ReLU networks, the derived training methods can be used for more sophisticated ANN architectures. As discussed above, greedily training one-hidden-layer ANNs leads to a well-performing deep network \citep{belilovsky19a}. Leveraging recent works that reform the training of more complex ANNs into convex programs \citep{Ergen21b, Ergen21a, Sahiner21}, our analysis can also extend to those ANNs because most convex training formulations share similar structures. Specifically, these convex training formulations rely on binary matrices to represent ReLU activation patterns and rely on convex (and often linear) constraints to enforce the patterns, with different regularizations revealing the sparse properties of different architectures. Coupling layer-wise training \citep{belilovsky19a} and SCP convex training recovers multi-layer ELMs.

As an example, we now extend our convex adversarial training analysis to various CNN formulations used in \citep{Ergen21b}. 

The paper \citep{Ergen21b} shows that the convex ANN training approach extends to various CNN architectures. Taking advantage of this result, the convex adversarial training formulations similarly generalize. In this part of the appendix, we change our notations to align with \citep{Ergen21b}. For example, the robust counterpart of the average pooling two-layer CNN convex training formulation (cf. Equations (4) and (26) in \citep{Ergen21b}) is:
\begin{align*}
    \min_{\{v_i,w_i\}\iPconv} & \Bigg( \max_{X_k \in \gX_k} \ell \bigg( \sum\iPconv \sum_{k=1}^K \Dbar_i^k X_k (w_i - v_i), \mathbf{y} \bigg) + \beta \sum\iPconv \big( \enorm{v_i} + \enorm{w_i} \big) \Bigg) \\
    \text{s.t.} \;\; & \min_{X_k \in \gX_k} \big( 2 \Dbar_i^k - I_n \big) X_k w_i \geq 0, \;\; \min_{X_k \in \gX_k} \big( 2 \Dbar_i^k - I_n \big) X_k v_i \geq 0, \;\; \forall i, k,
\end{align*}
where $v_i, w_i \in \sR^{\dbar}$ for all $i\in[\Pconv]$ and $\dbar$ is the convolutional filter size. Moreover, $X_k$ is the $k\th$ patch of the data matrix $X$ and $\gX_k$ is the corresponding perturbation set of the patch $X_k$. Furthermore, $\{\Dbar_1, \cdots, \Dbar_{\Pconv}\}$ is the set formed by all diagonal binary matrices that represent possible ReLU activation patterns associated with $\mathbf{M} \coloneqq \begin{bmatrix} X_1^\top & \cdots & X_{\Pconv}^\top \end{bmatrix}^\top$ and $\Dbar_i^k$ denotes the $k\th$ $\dbar$-by-$\dbar$ diagonal block of $\Dbar_i$.

The next step would be to show that the above formulation is equivalent to a classic convex optimization. Note that each robust constraint is an LP subproblem that can be solved in closed form, which means that the robust constraints can be cast as equivalent classic constraints. When $\ell(\cdot)$ is the squared loss, the above equation becomes a robust second-order cone program (SOCP), which is known to be a convex optimization (similar to \cref{eq:sl_minimax_convex}). Otherwise, if $\ell(\cdot)$ is monotonously increasing or decreasing in the CNN output $\widehat{\mathbf{y}}$ (examples include the hinge loss and the binary cross-entropy loss), the inner maximization problem
\begin{align*}
    \argmax_{X_k \in \gX_k} \ell \bigg( \sum\iPconv \sum_{k=1}^K \Dbar_i^k X_k (w_i - v_i), \mathbf{y} \bigg)
\end{align*}
reduces to
\begin{align*}
    \argmax_{X_k \in \gX_k} \sum\iPconv \sum_{k=1}^K \Dbar_i^k X_k (w_i - v_i) \;\; \text{ or } \;\; \argmin_{X_k \in \gX_k} \sum\iPconv \sum_{k=1}^K \Dbar_i^k X_k (w_i - v_i), 
\end{align*}
which are LPs that can be solved in closed form. Substituting the closed-form solution yields the desired convex adversarial training formulations.

Similarly, for max pooling two-layer CNNs, the robust counterpart becomes (cf. Equation (7) of \citep{Ergen21b}):
\begin{align*}
    \min_{\{v_i,w_i\}\iPconv} & \Bigg( \max_{X_k \in \gX_k} \ell \bigg( \sum\iPconv \sum_{k=1}^K \Dbar_i^k X_k (w_i - v_i), \mathbf{y} \bigg) + \beta \sum\iPconv \big( \enorm{v_i} + \enorm{w_i} \big) \Bigg) \\
    \text{s.t.} \quad & \min_{X_k \in \gX_k} \big( 2 \Dbar_i^k - I_n \big) X_k w_i \geq 0, \;\; \min_{X_k \in \gX_k} \big( 2 \Dbar_i^k - I_n \big) X_k v_i \geq 0, \;\; \forall i, k. \\
    & \min_{X_k \in \gX_k} \Dbar_i^k X_k v_i \geq \max_{X_j \in \gX_j} \Dbar_i^k X_j v_i, \; \forall i, j, k, \\[-.5mm]
    & \min_{X_k \in \gX_k} \Dbar_i^k X_k w_i \geq \max_{X_j \in \gX_j} \Dbar_i^k X_j w_i, \; \forall i, j, k.
\end{align*}
where each additional robust constraint is an LP subproblem solvable in closed form.

The same robust optimization techniques can be applied to three-layer CNNs (see Equation (11) in \citep{Ergen21b}) and derive corresponding convex adversarial training formulations. In general, the convex standard training formulations for different NNs / CNNs share very similar structures. Therefore, many convex standard training formulations can be ``robustified'' by recasting as mini-max formulations. Whether these mini-max formulations can be reformed into classic convex optimizations depends on the specific structures of the problems. For CNNs with two or three layers considered in \citep{Ergen21b}, such classic convex formulations can be derived.

Similarly, the ADMM splitting scheme, discussed in \Cref{sec:ADMM}, also applies to the above CNN formulations. The CNN training formulations also belong to the family of convex training formulations outlined in \cref{eq:ct_general}, and can be similarly split into loss function terms, regularization terms, and linear inequality constraints.

\subsection{\texorpdfstring{$\ell_p$}{lp} Norm-Bounded Perturbation Set for Hinge Loss} \label{sec:lpnorm}
\cref{thm:inner_max} can be extended to the following $\ell_p$ norm-bounded perturbation set: 
\begin{equation*}
    \widetilde{\gX} = \big\{ X + \Delta \in \sR^{n \times d} \ \big| \ 
    \Delta = [\delta_1 \; \cdots \; \delta_n]^\top, \ 
    \norm{\delta_k}_p \leq \epsilon, \ \forall k \in [n] \big\}.
\end{equation*}

In the case of performing binary classification with a hinge-lossed ANN, the convex adversarial training problem then becomes:
\begin{align} \label{eq:hinge_adv_D_lp}
    \min_{(v_i, w_i)\iPhat} & 
    \begin{pmatrix}
        \displaystyle \frac{1}{n} \: \sum_{k=1}^n \bigg( 1 - y_k \sum\iPhat d_{ik} x_k^\top (v_i - w_i) + \epsilon \cdot \biggnorm{ \sum\iPhat d_{ik} (v_i - w_i) }_{p*} \bigg)_+ \\[2mm]
        \hfill + \beta \sum\iPhat \Big( \norm{v_i}_2 + \norm{w_i}_2 \Big)
    \end{pmatrix} \\
    \ST & \quad (2 D_i - I_n) X v_i \geq \epsilon \norm{v_i}_{p*}, \quad (2 D_i - I_n) X w_i \geq \epsilon \norm{w_i}_{p*}, \quad \alliinPhat, \nonumber
\end{align}
where $D_1, \dots, D_{\widehat{P}}$ are all distinct diagonal matrices associated with $\diag([X u \geq 0])$ for all possible $u \in \mathbb{R}^d$ and all $X+\Delta$ at the \textit{boundary} of $\widetilde{\gX}$. Note that $\norm{\cdot}_{p*}$ is the dual norm of $\norm{\cdot}_p$.

\newpage
\section{Proofs}

\subsection{Proof of \texorpdfstring{\cref{thm:prac}}{Theorem 2.2}} \label{sec:pracproof}

We start by recasting the semi-infinite constraint of the dual formulation \cref{eq:dual} as $\max_{\norm{u}_2 \leq 1} |v^\top (Xu)_+| \leq \beta$ and obtain
\begin{align*}
    \max_{\norm{u}_2\leq1} \big| v^\top (X u)_+ \big| = & \max_{\norm{u}_2\leq1} \big| v^\top \diag([Xu\geq0]) X u \big| = \max_{i\in[P]} \bigg( \begin{matrix} \displaystyle \max_{\substack{\norm{u}_2 \leq 1 \\ (2D_i - I_n) Xu \geq 0}} \end{matrix} \big| v^\top D_i X u \big| \bigg),
\end{align*}
where the last equality holds by the definition of the $D_i$ matrices: $D_1 \dots, D_P$ are all distinct matrices that can be formed by $\diag([Xu\geq0])$ for some $u\in\R^d$. The constraint $(2D_i - I_n) X u \geq 0$ is equivalent to $D_i X u \geq 0$ and $(I_n - D_i) X u \leq 0$, which forces $D_i = \diag([Xu\geq0])$ to hold.

Therefore, the dual formulation \cref{eq:dual} can be recast as
\begin{align} \label{eq:dual2}
    \max_v -\ell^*(v) \qquad \ST \; \begin{matrix} \displaystyle \max_{\substack{\norm{u}_2 \leq 1 \\ (2D_i - I_n) Xu \geq 0}} \end{matrix} \big| v^\top D_i X u \big| \leq \beta,\ \forall i \in [P].
\end{align}

To form a tractable convex program that provides an approximation to \cref{eq:dual2}, one can independently sample a subset of the diagonal matrices. One possible sampling procedure is presented in \cref{alg:train}. The sampled matrices, denoted as $D_1, \dots, D_{P_s}$, can be used to construct the relaxed problem:
\begin{align} \label{eq:drelax}
    d_{s1}^\star = \max_v -\ell^*(v) \qquad \ST \; \begin{matrix} \displaystyle \max_{\substack{\norm{u}_2 \leq 1 \\ (2D_h - I_n) X u \geq 0}} \end{matrix} \big| v^\top D_h X u \big| \leq \beta,\ \alliinPs.
\end{align}

The optimization problem \cref{eq:drelax} is convex with respect to $v$. \citep{Pilanci20a} has shown that \cref{eq:dual2} has the same optimal objective as its dual problem \cref{eq:convex_general}. By following precisely the same derivation, it can be shown that \cref{eq:drelax} has the same optimal objective as \cref{eq:prac_clean} and $p_{s1}^\star = d_{s1}^\star$. Moreover, if an additional diagonal matrix $D_{P_s+1}$ is independently randomly sampled to form \cref{eq:prac_clean2}, then we also have $p_{s2}^\star = d_{s2}^\star$, where
\begin{align*} \label{eq:drelax2}
    d_{s2}^\star = \max_v -\ell^*(v) \qquad \ST \; \begin{matrix} \displaystyle \max_{\substack{\norm{u}_2 \leq 1 \\ (2D_h - I_n) X u \geq 0}} \end{matrix} \big| v^\top D_h X u \big| \leq \beta,\ \alliinPss.
\end{align*}

Thus, the level of suboptimality of \cref{eq:drelax} compared with \cref{eq:dual2} is the level of suboptimality of \cref{eq:prac_clean} compared with \cref{eq:convex_general}. Notice that by introducing a slack variable $w\in\sR$, \cref{eq:dual2} can be represented as an instance of the UCP with $n+1$ optimization variables, defined in \citep{Calafiore2005}:
\begin{equation*}
    \max_{v,w:\ w \leq -\ell^*(v)} w \qquad \ST \; \begin{matrix} \displaystyle \max_{\substack{\norm{u}_2 \leq 1 \\ (2D_i - I_n) X u \geq 0}} \end{matrix} \big| v^\top D_i X u \big| \leq \beta,\ \forall i \in [P].
\end{equation*}

The relaxed problem \cref{eq:drelax} can be regarded as a corresponding SCP. Suppose that $w^\star, v^\star$ is a solution to the sampled convex problem \cref{eq:drelax}. It can be concluded from \citep[Theorem 1]{Calafiore2005} and \citep[Theorem 1]{Campi09} that if $P_s \geq \min \big\{ \frac{n+1}{\psi \xi} - 1, \frac{2}{\xi} ( n+1-\log\psi ) \big\}$, then $v^\star$ satisfies the original constraints of the UCP \cref{eq:dual2} with high probability. Specifically, with probability no smaller than $1-\xi$, we have
\begin{equation*}
    \sP \Big\{ D\in\gD \ : \begin{matrix} \displaystyle \max_{\substack{\norm{u}_2 \leq 1 \\ (2D - I_n) X u \geq 0}} \end{matrix} \big| v^{\star\top} D X u \big| > \beta \Big\} \leq \psi.
\end{equation*}
where $\gD$ denotes the set of all diagonal matrices that can be formed by $\diag([X u \geq 0])$ for some $u\in\R^d$, which is the set formed by $D_1, \dots, D_P$. 

Since $D_{P_s+1}$ is randomly sampled from $\gD$, we have
\begin{equation*}
    \sP \Big\{ D\in\gD \ : \begin{matrix} \displaystyle \max_{\substack{\norm{u}_2 \leq 1 \\ (2D - I_n) X u \geq 0}} \end{matrix} \big| v^{\star\top} D X u \big| > \beta \Big\} =
    \sP \Big\{ \begin{matrix} \displaystyle \max_{\substack{\norm{u}_2 \leq 1 \\ (2D_{P_s+1} - I_n) X u \geq 0}} \end{matrix} \big| v^{\star\top} D_{P_s+1} X u \big| > \beta \Big\}
\end{equation*}

Thus, with probability no smaller than $1-\xi$, it holds that
\begin{equation*}
    \sP \Big\{ \begin{matrix} \displaystyle \max_{\substack{\norm{u}_2 \leq 1 \\ (2D_{P_s+1} - I_n) X u \geq 0}} \end{matrix} \big| v^{\star\top} D_{P_s+1} X u \big| > \beta \Big\} \leq \psi.
\end{equation*}

Moreover, $d_{s2}^\star < d_{s1}^\star$ if and only if $\big| v^{\star\top} D_{P_s+1} X u \big| > \beta$ with $d_{s2}^\star = d_{s1}^\star$ otherwise. The proof is completed by noting that $p_{s1}^\star = d_{s1}^\star$ and $p_{s2}^\star = d_{s2}^\star$. $\hfill \square$

\subsection{Proof of \texorpdfstring{\cref{thm:ADMM}}{Theorem 3.1}} \label{sec:ADMMproof}

We start by rewriting \cref{eq:4} as
\begin{align} \label{eq:5}
    \min_{v,s,u:\ s \geq 0} f_1(u) + f_2(v,s) \quad \ST \quad E_1 u - E_2 \begin{bmatrix} v \\ s \end{bmatrix} = 0,
\end{align}
where $f_1(u) = \ell(F u, y)$, $f_2(v,s) = \beta \norm{v}_{2,1}$, $E_1 = \begin{bmatrix} I \\ G \end{bmatrix}$, and $E_2 = I$.

Furthermore, let $L(u,v,s,\nu,\lambda)$ denote the augmented Lagrangian:
\begin{align*}
    & L(u,v,s,\nu,\lambda) \coloneqq \\[-1mm]
    & \hspace{15mm} f_1 (u) + \beta \norm{v}_{2,1} + \sI_{\geq0}(s) + \frac{\rho}{2} \Big( \norm{u - v + \lambda}_2^2 - \norm{\lambda}_2^2 \Big) + \frac{\rho}{2} \Big( \norm{G u - s + \nu}_2^2 - \norm{\nu}_2^2 \Big)
\end{align*}

Theorem 3.1 in \citep{Hong17} shows that the ADMM algorithm converges linearly when the objective satisfies seven conditions. We show that these conditions are all satisfied for \cref{eq:5} given the assumptions of \cref{thm:ADMM} in this paper:

\begin{enumerate}[(a)]
    \setlength\itemsep{.2em}
    \item It can be easily shown that \cref{eq:5} attains a global solution because the feasible set of the equivalent problem \cref{eq:convex_general} is non-empty.
    \item We can then decompose $f_1(u)$ into $g_1(F u) \coloneqq \ell(F u, y)$ and $h_1(u) \coloneqq 0$ and define $h_2(\cdot) \coloneqq f_2(\cdot)$. When the loss $\ell(\yhat,y)$ is convex with respect to $\yhat$, the functions $g_1(\cdot), h_1(\cdot), h_2(\cdot)$ are all convex and continuous.
    \item When $\ell(\yhat,y)$ is strictly convex and continuously differentiable with a uniform Lipschitz continuous gradient with respect to $\yhat$, the function $g_1(\cdot)$ is strictly convex and continuously differentiable with a uniform Lipschitz continuous gradient.
    \item The epigraph of $h_1(\cdot) = 0$ is a polyhedral set. Moreover, $h_2(v,s) = \norm{v}_{2,1} = \linebreak \sum\iP (\enorm{v_i} + \enorm{w_i})$ by definition.
    \item The constant function $h_1(\cdot)$ is trivially finite. Furthermore, for all $u, v, s$ that make $L(u,v,s,\nu,\lambda)$ finite, it must hold that $f_1 (u) < +\infty$, $v < +\infty$, and $s \geq 0$. Therefore, $h_2(\cdot)$ must be finite.
    \item $E_1$ and $E_2$ both have full column rank since the identity matrix has full column rank.
    \item When $u\to\infty$, we have $L(u,v,s,\nu,\lambda)\to\infty$. Hence, the solution to \cref{eq:ADMM1} must be finite as long as the initial points $u^0, v^0, s^0, \lambda^0, \nu^0$ are finite. The solutions to \cref{eq:ADMM2} and \cref{eq:ADMM3} are also finite, since the closed-form solutions are derived in \Cref{sec:svupdates}. Therefore, the sequence $\{(u^k, v^k, s^k, \lambda^k, \nu^k)\}$ is finite. Thus, there exist finite $u_{\max}, v_{\max}, s_{\max}$ such that \cref{eq:5} is equivalent to the formulation below:
    \begin{align} \label{eq:6}
        \min_{v,s,u:\ s\geq0} \quad & f_1(u) + f_2(v,s) \\[-1mm]
        \ST \quad & E_1 u - E_2 \begin{bmatrix} v \\ s \end{bmatrix} = 0, \;\; \norm{u}_\infty \leq u_{\max}, \;\; \norm{v}_\infty \leq v_{\max}, \;\; \norm{s}_\infty \leq s_{\max}. \nonumber
    \end{align}
    Furthermore, the ADMM algorithm that solves \cref{eq:6} is equivalent to \cref{alg:ADMM}. The feasible set of \cref{eq:6} is a compact polyhedral set formed by the $\ell_\infty$ norm constraints, the non-negativity constraints, and the linear equality constraints.
\end{enumerate}

Thus, by the application of \citep[Theorem 3.1]{Hong17}, the desired result holds true when the step size $\gamma_a$ is sufficiently small. $\hfill \square$

\subsection{Proof of \texorpdfstring{\cref{thm:SCP}}{Theorem B.1}} \label{sec:SCPproof}

As discussed in \Cref{sec:itersampling}, strong duality holds between \cref{eq:noncvx2} and \cref{eq:UCP}, as well as between \cref{eq:SCP2} and \cref{eq:SCP}. Here, we introduce a slack variable $w$ and cast \cref{eq:UCP} as a canonical uncertain convex program with $n+1$ optimization variables and a linear objective, where $n$ is the number of training data:
\begin{align*}
    & \min_{(v,w)\in\gF} w \\
    \ST \quad & f(v,w,u) \coloneqq |v^\top (X u)_+| - \beta \leq 0, \; \forall u\in\gG \\
    & \gF = \big\{ v\in\R^n, w\in\R \,\big|\, \norm{y-v}_2^2 - 2 w \leq 0 \big\} \\
    & \hspace{.5mm} \gG = \big\{ u \,\big|\, \norm{u}_2 = 1 \big\} .
\end{align*}

By leveraging \citep[Theorem 1]{Calafiore2005} and \citep[Theorem 1]{Campi09}, we can conclude that if $N \geq \min \big\{ \frac{n+1}{\psi \gamma} - 1, \frac{2}{\gamma} ( n+1-\log\psi ) \big\}$, then with probability no smaller than $1-\gamma$, the solution $v^\star$ to the randomized problem \cref{eq:SCP} satisfies $\sP\{ u: \norm{u}_2=1, |v^{\star\top} (X u)_+| > \beta \} \leq \psi$. Since $u_{N+1}$ is randomly generated on the Euclidean norm sphere via a uniform distribution, it holds that $\sP\{ |v^{\star\top} (X u_{N+1})_+| > \beta \} \leq \psi$.

Consider the following dual formulation with the newly sampled hidden neuron $u_{N+1}$ included:
\begin{align} \label{eq:SCPaugd}
    d_{s4}^\star = \max_{v\in\R^n} -\ell^*(v) \quad \ST \quad & |v^\top (X u_i)_+| \leq \beta, \;\; \forall i\in[N+1].
\end{align}

Since \cref{eq:SCPaugd} and \cref{eq:SCP} share the same objective, it holds that $d_{s4}^\star < d_{s3}^\star$ if and only if $|v^{\star\top} (X u_{N+1})_+| > \beta$ with $d_{s4}^\star = d_{s3}^\star$ otherwise. The proof is completed by recalling that $p_{s3}^\star = d_{s3}^\star$ and $p_{s4}^\star = d_{s4}^\star$ due to strong duality. $\hfill \square$

\subsection{Details About the Strong Duality Between \texorpdfstring{\cref{eq:SCP}}{(B.5)} and \texorpdfstring{\cref{eq:SCP2}}{(B.2)}} \label{sec:SCPdual}

\subsubsection{General Loss Functions}

In this part of the appendix, we explicitly derive the relationship between the optimal solutions $(\alpha_i^\star)\iN$ and $v^\star$ for the purpose of recovering the dual optimizers from the primal optimizers. 

The SCP training formulation \cref{eq:SCP2} is equivalent to the following constrained optimization:
\begin{align} \label{eq:A1}
    \min_{r, (\alpha_i)\iN} \ell (r, y) + \beta \sum\iN |\alpha_i| \qquad \ST \quad r = \sum\iN \big( X u_i \big)_+ \alpha_i,
\end{align}
and a solution to \cref{eq:SCP2} is also optimal for \cref{eq:A1}. The optimization problem \cref{eq:A1} is then equivalent to the minimax problem
\begin{align} \label{eq:A2}
    \min_{r, (\alpha_i)\iN} \bigg( \max_v \ell(r, y) + \beta \sum\iN |\alpha_i| + v^\top \Big( \sum\iN (X u_i)_+ \alpha_i - r \Big) \bigg).
\end{align}

The outer minimization is convex over $r$ and $(\alpha_i)\iN$, while the inner maximization is concave over $v$. Thus, by the Sion's minimax theorem \citep{Sion1958}, the optimization problem \cref{eq:A2} is equivalent to:
\begin{align*}
    & \max_v \bigg( \min_r \Big( \ell(r,y) - v^\top r \Big) + \min_{(\alpha_i)\iN} \Big( \beta \sum\iN |\alpha_j| + v^\top \sum\iN (X u_j)_+ \alpha_j \Big) \bigg) \\
    = & \max_v \bigg( -\max_r \Big( v^\top r - \ell(r,y) \Big) \quad \ST \quad \big| v^\top (X u_i)_+ \big| \leq \beta,\ \forall i\in[N] \bigg) \\
    = & \max_v -\ell^*(v) \quad \ST \quad \big| v^\top (X u_i)_+ \big| \leq \beta,\ \forall i\in[N],
\end{align*}
which is \cref{eq:SCP}. The first equality holds because
\begin{align*}
    \min_{(\alpha_i)\iN} \Big( \beta \sum\iN |\alpha_j| + v^\top \sum\iN (X u_j)_+ \alpha_j \Big) = \begin{cases} 0, & \big| v^\top (X u_i)_+ \big| \leq \beta,\ \forall i\in[N], \\ \infty, & \text{otherwise.} \end{cases}
\end{align*}

Therefore, with the optimal $(\alpha_i^\star)\iN$, one can calculate $r^\star$ via $r^\star = \sum\iN \big( X u_i \big)_+ \alpha_i^\star$, and recover $v^\star$ by solving the following LP:
\begin{align*}
    v^\star = \argmax_v -v^\top r^\star \quad \ST \quad \big| v^\top (X u_i)_+ \big| \leq \beta,\ \forall i\in[N].
\end{align*}

\subsubsection{Squared Loss}

In this part, we prove the relationship between $(\alpha_i^\star)\iN$ and $v^\star$ by deriving the Karush--Kuhn--Tucker (KKT) conditions for the special case when the squared loss is considered.
In this case, the SCP training formulation \cref{eq:SCP2} reduces to
\begin{align*}
    \min_{(\alpha_i)\iN} \frac{1}{2} \Bignorm{\sum\iN \big( X u_i \big)_+ \alpha_i - y}_2^2 + \beta \sum\iN |\alpha_i|,
\end{align*}
which is equivalent to
\begin{align} \label{eq:A3}
    \min_{r, (\alpha_i)\iN} \frac{1}{2} \norm{r}_2^2 + \beta \sum\iN |\alpha_i| \qquad \ST \quad r = \sum\iN \big( X u_i \big)_+ \alpha_i - y.
\end{align}

By introducing a dual vector variable $v \in \R^n$, we can write the Lagrangian of \cref{eq:A3} as:
\begin{align*}
    L_{\text{SCP}} \Big( v, r, (\alpha_i)\iN \Big) = & \frac{1}{2} \norm{r}_2^2 + \beta \sum\iN |\alpha_i| + v^\top \Big( \sum\iN \big( X u_i \big)_+ \alpha_i - y - r \Big) \\
    = & \Big( \frac{1}{2} r^\top + v^\top \Big) r + \Big( \beta \sum\iN |\alpha_i| + v^\top \sum\iN \big( X u_i \big)_+ \alpha_i \Big) + v^\top y
\end{align*}

$L_{\text{SCP}} \big( v, r, (\alpha_i)\iN \big)$ is smooth with respect to $r$. Thus, by the Lagrangian stationarity condition, at optimum, we must have $\nabla_r L \big( v^\star, r^\star, (\alpha_i^\star)\iN \big) = r^\star + v^\star = 0$. By the primal feasibility condition, we must have $r^\star = \sum\iN \big( X u_i \big)_+ \alpha_i^\star - y$. Thus, at the optimum, $v^\star = y - \sum\iN \big( X u_i \big)_+ \alpha_i^\star$.

\subsection{Proof of \texorpdfstring{\cref{THM:CVX_MINIMAX}}{Theorem 4.1}} \label{sec:CVX_MINIMAX}

Before proceeding with the proof, we first present the following result borrowed from \citep{Pilanci20a}.

\begin{lemma} \label{lemma:RECOVER}
    For a given data matrix $X$ and $(v_i, w_i)\iP$, if $(2 D_i - I_n) X v_i \geq 0$ and $(2 D_i - I_n) X w_i \geq 0$ for all $i \in [P]$, then we can recover the corresponding ANN weights \linebreak $(u_{v,w_j}, \alpha_{v,w_j})\jms$ using the formulas in \cref{eq:recover_weights}, and it holds that
    \vspace{-1mm}
    \begin{align} \label{MertEquality}
    \ell \bigg( \sum\iP D_i X (v_i - w_i), y \bigg) \; + & \; \beta \sum\iP \big( \norm{v_i}_2 + \norm{w_i}_2 \big) \nonumber \\
    = & \; \ell \bigg( \sum\jms (X u_{v,w_j})_+ \alpha_{v,w_j}, y \bigg) + \frac{\beta}{2} \sum\jms \big( \norm{u_{v,w_j}}_2^2 + \alpha_{v,w_j}^2 \big).
    \end{align}
\end{lemma}

\cref{THM:PILANCI} implies that the non-convex cost function \cref{eq:nonconvex_general} has the same objective value as the following finite-dimensional convex optimization problem:
\begin{equation} \label{eq:theorem_5}
\begin{aligned}
    q^\star = \min_{(v_i, w_i)\iP} & \ell \bigg( \sum\iP D_i X (v_i - w_i), y \bigg) + \beta \sum\iP \big(\norm{v_i}_2 + \norm{w_i}_2\big) \\
    \ST \quad & (2 D_i - I_n) X v_i \geq 0, \; (2 D_i - I_n) X w_i \geq 0, \;\; \alliinP
\end{aligned}
\end{equation}
where $D_1, \dots, D_P$ are all of the matrices in the set of matrices $\gD$, which is defined as the set of all distinct diagonal matrices $\diag([X u \geq 0])$ that can be obtained for all possible $u \in \mathbb{R}^d$. We recall that the optimal neural network weights can be recovered using \cref{eq:recover_weights}.

Consider the following optimization problem:
\begin{equation}\label{eq:redundant_D}
\begin{aligned}
    \widetilde{q}^\star = \min_{(v_i, w_i)\itildeP} & \ell \bigg( \sum\itildeP D_i X (v_i - w_i), y \bigg) + \beta \sum\itildeP \big(\norm{v_i}_2 + \norm{w_i}_2\big) \\
    \ST \quad & (2 D_i - I_n) X v_i \geq 0, \; (2 D_i - I_n) X w_i \geq 0, \;\; \forall i \in [\widetilde{P}]
\end{aligned}
\end{equation}
where additional $D$ matrices, denoted as $D_{P+1}, \dots, D_{\widetilde{P}}$, are introduced. These additional matrices are still diagonal with each entry being either 0 or 1, while they do not belong to $\gD$. They represent ``infeasible hyperplanes'' that cannot be achieved by the sign pattern of $X u$ for any $u\in\R^d$. 

\begin{lemma} \label{lemma:RED_D}
    It holds that $\widetilde{q}^\star = q^\star$, meaning that the optimization problem \cref{eq:redundant_D} has the same optimal objective as \cref{eq:theorem_5}.
\end{lemma}
The proof of \cref{lemma:RED_D} is given in \Cref{sec:RED_D}.

The robust minimax training problem \cref{eq:robust_general} considers an uncertain data matrix $X+\Delta$. Different values of $X+\Delta$ within the perturbation set $\gU$ can result in different $D$ matrices. 
Now, we define $\widehat{\gD} = \bigcup_\Delta \gD_\Delta$, where $\gD_\Delta$ is the set of diagonal matrices for a particular $\Delta$ such that $X+\Delta \in \gU$.
By construction, we have $\gD_\Delta \subseteq \widehat{\gD}$ for every $\Delta$ such that $X+\Delta \in \gU$. Thus, if we define $D_1, \dots, D_{\widehat{P}}$ as all matrices in $\widehat{\gD}$, then for every $\Delta$ with the property $X+\Delta \in \gU$, the optimization problem
\begin{equation} \label{eq:cvx_allD}
\begin{aligned}
    \min_{(v_i, w_i)\iPhat} & \ell \bigg( \sum\iPhat D_i (X+\Delta) (v_i - w_i), y \bigg) + \beta \sum\iPhat (\norm{v_i}_2 + \norm{w_i}_2) \\
    \ST \quad & (2 D_i - I_n) (X+\Delta) v_i \geq 0, \; (2 D_i - I_n) (X+\Delta) w_i \geq 0, \;\: \alliinPhat
\end{aligned}
\end{equation}
is equivalent to
\begin{equation*}
        \min_{(u_j, \alpha_j)\jm} \ell \bigg( \sum\jm ((X+\Delta)u_j)_+ \alpha_j, y \bigg) + \frac{\beta}{2} \sum\jm \big( \norm{u_j}_2^2 + \alpha_j^2 \big)
\end{equation*}
as long as $m \geq \widehat{m}^\star$ with $\widehat{m}^\star = |\{i : v_i^\star(\Delta) \neq 0\}| + |\{i : w_i^\star(\Delta) \neq 0\}|$, where $(v_i^\star(\Delta), w_i^\star(\Delta))\iPhat$ denotes an optimal point to \cref{eq:cvx_allD}.

Now, we focus on the minimax training problem with a convex objective given by
\begin{equation} \label{eq:minimax1}
\begin{aligned}
    \min_{(v_i, w_i)\iPhat\in\gF} 
    \begin{pmatrix}
        \displaystyle \max\allDeltaU \ell \bigg( \sum\iPhat D_i (X+\Delta) (v_i - w_i), y \bigg) + \beta \sum\iPhat \big( \norm{v_i}_2 + \norm{w_i}_2 \big) \hfill \\[4mm]
        \quad\ \ST \;\; (2 D_i - I_n) (X+\Delta) v_i \geq 0, \; (2 D_i - I_n) (X+\Delta) w_i \geq 0, \; \alliinPhat
    \end{pmatrix},
\end{aligned}
\end{equation}
where $\gF$ is defined as:
\vspace{1mm}
\begin{equation*}
    \bigg\{ (v_i, w_i)\iPhat \; \bigg| \; \begin{matrix}
        \; \exists\Delta:X+\Delta\in\gU \hfill \\ \ST \; (2 D_i - I_n) (X+\Delta) v_i \geq 0, \; (2 D_i - I_n) (X+\Delta) w_i \geq 0, \; \alliinPhat
    \end{matrix} \bigg\}.
\end{equation*}

The introduction of the feasible set $\gF$ is to avoid the situation where the inner maximization over $\Delta$ is infeasible and the objective becomes $- \infty$, leaving the outer minimization problem unbounded.

Moreover, consider the following problem:
\begin{equation} \label{eq:minimax2}
\begin{aligned}
    \min_{(v_i, w_i)\iPhat} & 
    \begin{pmatrix}
        \displaystyle \ell \bigg( \sum\iPhat D_i (X+\DeltaStar) (v_i - w_i), y \bigg) + \beta \sum\iPhat \big( \norm{v_i}_2 + \norm{w_i}_2 \big)
    \end{pmatrix} \\
    \ST \quad & (2 D_i - I_n) (X+\DeltaStar) v_i \geq 0, \; (2 D_i - I_n) (X+\DeltaStar) w_i \geq 0, \;\; \alliinPhat,
\end{aligned}
\end{equation}
where $\DeltaStar$ is the optimal point for $\max\allDeltaU \ \ell \left( \sum\iPhat D_i (X+\Delta) (v_i - w_i), y \right)$. 
Note that the inequality constraints are dropped for the maximization here compared to \cref{eq:minimax1}.

The optimization problem \cref{eq:minimax1} gives a lower bound on \cref{eq:minimax2}. To prove this, we first rewrite \cref{eq:minimax2} as:
\begin{align*}
    \min_{(v_i, w_i)\iPhat} & f \big( (v_i, w_i)\iPhat \big) \text{, where } f \big( (v_i, w_i)\iPhat \big) = \\
    & \begin{cases}
        \ell \Big( \sum\iPhat D_i (X+\DeltaStar) (v_i - w_i), y \Big) & (2 D_i - I_n) (X+\DeltaStar) v_i \geq 0, \; \alliinPhat \\[-1mm]
        \hspace{2cm} + \beta \sum\iPhat \big( \norm{v_i}_2 + \norm{w_i}_2 \big), & (2 D_i - I_n) (X+\DeltaStar) w_i \geq 0, \; \alliinPhat \\[2.5mm]
        + \infty, & \text{otherwise}.
    \end{cases}
\end{align*}

Now, we analyze \cref{eq:minimax1} by considering the following three cases.

Case 1: For some $(v_i, w_i)\iPhat$, $\DeltaStar$ is optimal for the inner maximization of \cref{eq:minimax1} and the inequality constraints are inactive. This happens whenever $\DeltaStar$ is feasible for the particular choice of $(v_i, w_i)\iPhat$. In other words, $(2 D_i - I_n) (X+\DeltaStar) v_i \geq 0$ and $(2 D_i - I_n) (X+\DeltaStar) w_i \geq 0$ hold true for all $i\in[\widehat{P}]$. For these $(v_i, w_i)\iPhat$, we have:
\begin{align*}
    & \begin{pmatrix}
        \displaystyle \max\allDeltaU \ell \bigg( \sum\iPhat D_i (X+\Delta) (v_i - w_i), y \bigg) + \beta \sum\iPhat \big( \norm{v_i}_2 + \norm{w_i}_2 \big) \\[4mm]
        \ST \:\: (2 D_i - I_n) (X+\Delta) v_i \geq 0, \; (2 D_i - I_n) (X+\Delta) w_i \geq 0, \; \alliinPhat
    \end{pmatrix} \\[1mm]
    & \hspace{15mm} = \ell \bigg( \sum\iPhat D_i (X+\DeltaStar) (v_i - w_i), y \bigg)
        + \beta \sum\iPhat \big( \norm{v_i}_2 + \norm{w_i}_2 \big).
\end{align*}

Case 2: For some $(v_i, w_i)\iPhat$, $\DeltaStar$ is infeasible, while some $\Delta$ within the perturbation bound satisfies the inequality constraints. Suppose that among the feasible $\Delta$'s,
\begin{align*}
    \DeltaStarTil = & \argmax\allDeltaU \ell \bigg( \sum\iPhat D_i (X+\Delta) (v_i - w_i), y \bigg) + \beta \sum\iPhat \big( \norm{v_i}_2 + \norm{w_i}_2 \big) \\
    & \ST \:\: (2 D_i - I_n) (X+\Delta) v_i \geq 0, \; (2 D_i - I_n) (X+\Delta) w_i \geq 0, \; \alliinPhat.
\end{align*}
In this case,
\begin{align*}
    & \begin{pmatrix}
        \displaystyle \max\allDeltaU \ell \bigg( \sum\iPhat D_i (X+\Delta) (v_i - w_i), y \bigg) + \beta \sum\iPhat \big( \norm{v_i}_2 + \norm{w_i}_2 \big) \hfill \\[4mm]
        \ST \:\: (2 D_i - I_n) (X+\Delta) v_i \geq 0, \; (2 D_i - I_n) (X+\Delta) w_i \geq 0, \; \alliinPhat
    \end{pmatrix} \\[1mm]
    & \hspace{15mm} = \ell \bigg( \sum\iPhat D_i (X+\DeltaStarTil) (v_i - w_i), y \bigg)
        + \beta \sum\iPhat \big( \norm{v_i}_2 + \norm{w_i}_2 \big)
\end{align*}

Case 3: For all other $(v_i, w_i)\iPhat$, the objective value is $+ \infty$ since they do not belong to $\gF$.

Therefore, \cref{eq:minimax1} can be rewritten as
\begin{align*}
    & \min_{(v_i, w_i)\iPhat} g \big( (v_i, w_i)\iPhat \big), \ \text{where } g \big( (v_i, w_i)\iPhat \big) = \\
    & \; \begin{cases}
        \ell \Big( \sum\iPhat D_i (X+\DeltaStar) (v_i - w_i), y \Big) & (2 D_i - I_n) (X+\DeltaStar) v_i \geq 0, \; \alliinPhat \\[-1mm]
        \hspace{23mm} + \beta \sum\iPhat \big( \norm{v_i}_2 + \norm{w_i}_2 \big), & (2 D_i - I_n) (X+\DeltaStar) w_i \geq 0, \; \alliinPhat \\[2.5mm]
        & \exists j: (2 D_j - I_n) (X+\DeltaStar) v_j < 0 \\
        \ell \Big( \sum\iPhat D_i (X+\DeltaStarTil) (v_i - w_i), y \Big) & \hspace{2.3mm} \text{or} \ (2 D_j - I_n) (X+\DeltaStar) w_j < 0 \\[-1mm]
        \hspace{23mm} + \beta \sum\iPhat \big( \norm{v_i}_2 + \norm{w_i}_2 \big), & \exists \Delta: \ (2 D_i - I_n) (X+\Delta) v_i \geq 0, \; \alliinPhat \\
        & \hspace{8.7mm} (2 D_i - I_n) (X+\Delta) w_i \geq 0, \; \alliinPhat \\[2.5mm]
        + \infty, & \text{otherwise}
    \end{cases}
\end{align*}

Hence, $g((v_i, w_i)\iPhat) = f((v_i, w_i)\iPhat)$ for all $(v_i, w_i)\iPhat$ belonging to the first and the third cases. $g((v_i, w_i)\iPhat) < f((v_i, w_i)\iPhat)$ for all $(v_i, w_i)\iPhat$ belonging to the second case. Thus, $\min_{(v_i, w_i)\iPhat} g((v_i, w_i)\iPhat) \leq \min_{(v_i, w_i)\iPhat} f((v_i, w_i)\iPhat)$. This concludes that \cref{eq:minimax1} is a lower bound to \cref{eq:minimax2}. 

Let $(v_{\text{minimax}_i}^\star, w_{\text{minimax}_i}^\star)\iPhat$ denote an optimal point for \cref{eq:minimax2}. It is possible that for some $\Delta: X+\Delta \in \gU$, the constraints $(2 D_i - I_n) (X+\Delta) v_{\text{minimax}_i}^\star \geq 0$ and $(2 D_i - I_n) (X+\Delta) w_{\text{minimax}_i}^\star \geq 0$ are not satisfied for all $i\in[\widehat{P}]$.
In light of \cref{lemma:RECOVER}, at those $\Delta$ where such constraints are violated, the convex problem \cref{eq:minimax2} does not reflect the cost of the ANN. For these infeasible $\Delta$, the input-label pairs $(X+\Delta, y)$ can have a high cost in the ANN and potentially become the worst-case adversary. However, these $\Delta$ are ignored in \cref{eq:minimax2} due to the infeasibility.
Since adversarial training aims to minimize the cost over the worst-case adversaries generated upon the training data whereas \cref{eq:minimax2} may sometimes miss the worst-case adversaries, \cref{eq:minimax2} does not fully accomplish the task of adversarial training.
In fact, by applying \cref{THM:PILANCI} and \cref{lemma:RED_D}, it can be verified that \cref{eq:minimax1} and \cref{eq:minimax2} are lower bounds to \cref{eq:robust_general} as long as $m \geq \widehat{m}^\star$:
\begin{equation*}
    \begin{aligned}
        & \min_{(u_j, \alpha_j)\jm} 
        \begin{pmatrix}
            \displaystyle \max\allDeltaU \ell \bigg( \sum_{j=1}^m \big( (X+\Delta) u_j \big)_+ \alpha_j, y \bigg) + \frac\beta2 \sum_{j=1}^m \Big( \norm{u_j}_2^2 + \alpha_j^2 \Big)
        \end{pmatrix} \\
        & \hspace{15mm} \geq \min_{(u_j, \alpha_j)\jm} \ell \bigg( \sum_{j=1}^m \big( (X+\DeltaStar) u_j \big)_+ \alpha_j, y \bigg) + \frac\beta2 \sum_{j=1}^m \Big( \norm{u_j}_2^2 + \alpha_j^2 \Big) \\
        & \hspace{15mm} = \begin{pmatrix}
            \displaystyle \min_{(v_i, w_i)\iPhat} \ell \bigg( \sum\iPhat D_i (X+\DeltaStar) (v_i - w_i), y \bigg) + \beta \sum\iPhat \big( \norm{v_i}_2 + \norm{w_i}_2 \big) \\[4mm]
            \ST \; (2 D_i - I_n) (X+\DeltaStar) v_i \geq 0, \; (2 D_i - I_n) (X+\DeltaStar) w_i \geq 0, \; \alliinPhat
        \end{pmatrix}.
    \end{aligned}
\end{equation*}

To address the feasibility issue, we can apply robust optimization techniques (\citep{boyd2004convex} Section 4.4.2) and replace the constraints in \cref{eq:minimax2} with robust convex constraints, which will lead to \cref{eq:rob_gen_cvx}. 
Let $\big( (v_{\text{rob}_i}^\star, w_{\text{rob}_i}^\star)\iPhat, \Delta_{\text{rob}}^\star \big)$ denote an optimal point of \cref{eq:rob_gen_cvx} and let $(u_{\text{rob}_j}^\star, \alpha_{\text{rob}_j}^\star)\jmhs$ be the ANN weights recovered from $(v_{\text{rob}_i}^\star, w_{\text{rob}_i}^\star)\iPhat$ with \cref{eq:recover_weights}, where $\mhs$ is the number of nonzero weights.
In light of \cref{lemma:RECOVER}, since the constraints $(2 D_i - I_n) (X+\Delta) v_{\text{rob}_i}^\star \geq 0$ and $(2 D_i - I_n) (X+\Delta) w_{\text{rob}_i}^\star \geq 0$ for all $i\in[\widehat{P}]$ apply to all $X+\Delta\in\gU$, all $X+\Delta\in\gU$ satisfy the equality
\begin{align*}
    \displaystyle & \ell \bigg( \sum\iPhat D_i (X+\Delta) (v_{\text{rob}_i}^\star - w_{\text{rob}_i}^\star), y \bigg) + \beta \sum\iPhat \big( \norm{v_{\text{rob}_i}^\star}_2 + \norm{w_{\text{rob}_i}^\star}_2 \big) \\
    & \hspace{30mm} = \ell \bigg( \sum\jmhs \big( (X+\Delta) u_{\text{rob}_j}^\star \big)_+ \alpha_{\text{rob}_j}^\star, y \bigg) + \frac{\beta}{2} \sum\jmhs \big( \norm{u_{\text{rob}_j}^\star}_2^2 + \alpha_{\text{rob}_j}^{\star 2} \big).
\end{align*} 

Thus, since 
\begin{equation*}
    \displaystyle \Delta_{\text{rob}}^\star = \argmax\allDeltaU \ell \bigg( \sum\iPhat D_i (X+\Delta) (v_{\text{rob}_i}^\star - w_{\text{rob}_i}^\star), y \bigg) + \beta \sum\iPhat \big( \norm{v_{\text{rob}_i}^\star}_2 + \norm{w_{\text{rob}_i}^\star}_2 \big),
\end{equation*}
we have 
\begin{equation*}
    \displaystyle \Delta_{\text{rob}}^\star = \argmax\allDeltaU \ell \bigg( \sum\jmhs \big( (X+\Delta) u_{\text{rob}_j}^\star \big)_+ \alpha_{\text{rob}_j}^\star, y \bigg) + \frac{\beta}{2} \sum\jmhs \big( \norm{u_{\text{rob}_j}^\star}_2^2 + \alpha_{\text{rob}_j}^{\star 2} \big),
\end{equation*}
giving rise to:
\begin{equation*}
    \begin{aligned}
        & \ell \bigg( \sum\iPhat D_i (X+\Delta_{\text{rob}}^\star) (v_{\text{rob}_i}^\star - w_{\text{rob}_i}^\star), y \bigg) + \beta \sum\iPhat \big( \norm{v_{\text{rob}_i}^\star}_2 + \norm{w_{\text{rob}_i}^\star}_2 \big) \\
        & \hspace{25mm} = \ell \bigg( \sum\jmhs \big( (X+\Delta_{\text{rob}}^\star) u_{\text{rob}_j}^\star \big)_+ \alpha_{\text{rob}_j}^\star, y \bigg) + \frac{\beta}{2} \sum\jmhs \big( \norm{u_{\text{rob}_j}^\star}_2^2 + \alpha_{\text{rob}_j}^{\star 2} \big) \\
        & \hspace{25mm} = \max\allDeltaU \ell \bigg( \sum\jmhs \big( (X+\Delta) u_{\text{rob}_j}^\star \big)_+ \alpha_{\text{rob}_j}^\star, y \bigg) + \frac\beta2 \sum\jmhs \big( \norm{u_{\text{rob}_j}^\star}_2^2 + \alpha_{\text{rob}_j}^{\star 2} \big) \\
        & \hspace{25mm} \geq \min_{(u_j, \alpha_j)\jmhs} 
        \begin{pmatrix}
            \displaystyle \max\allDeltaU \ell \bigg( \sum\jmhs \big( (X+\Delta) u_j \big)_+ \alpha_j, y \bigg) + \frac\beta2 \sum\jmhs \big( \norm{u_j}_2^2 + \alpha_j^2 \big)
        \end{pmatrix}
    \end{aligned}
\end{equation*}

Therefore, \cref{eq:rob_gen_cvx} is an upper bound to \cref{eq:robust_general}. $\hfill \square$

\subsection{Proof of \texorpdfstring{\cref{CORO:ROB_CONSTRAINT}}{Corollary 4.2}} \label{sec:ROB_CONSTRAINT}

Define $E_i = 2 D_i - I_n$ for all $i\in[\widehat{P}]$.
Note that each $E_i$ is a diagonal matrix, and its diagonal elements are either -1 or 1. Therefore, for each $i\in[\widehat{P}]$, we can analyze the robust constraint $\min\allDeltaU E_i (X+\Delta) v_i \geq 0$ element-wise (for each data point).
Let $e_{ik}$ denote the $k\th$ diagonal element of $E_i$ and $\delta_{ik}^\top$ denote the $k\th$ element of $\Delta$ that appears in the $i^{\th}$ constraint. We then have:
\begin{equation} \label{eq:hinge_constraint_min}
    \begin{pmatrix}
        \displaystyle \min_{\norm{\delta_{ik}}_\infty \leq \epsilon} e_{ik} (x_k^\top + \delta_{ik}^\top) v_i
    \end{pmatrix} = \begin{pmatrix}
        \displaystyle e_{ik} x_k^\top v_i + \min_{\norm{\delta_{ik}}_\infty \leq \epsilon} e_{ik} \delta_{ik}^\top v_i
    \end{pmatrix} \geq 0
\end{equation}

The minima of the above optimization problems are achieved at $\delta_{ik}^{\star\star} = \epsilon \cdot \sgn(e_{ik} v_i) = \epsilon \cdot e_{ik} \cdot \sgn(v_i)$.

Note that as $\epsilon$ approaches 0, $\delta_{ik}^{\star\star}$ and $\Delta_{\text{rob}}^\star$ in \cref{THM:CVX_MINIMAX} both approach 0, which means that the gap between the convex robust problem \cref{eq:hinge_adv_D} and the non-convex adversarial training problem \cref{eq:hinge_adv} diminishes.
Substituting $\delta_k^{\star\star}$ into \cref{eq:hinge_constraint_min} yields that
\begin{equation*}
\begin{aligned}
    \Big( e_{ik} x_k^\top v_i - \epsilon \norm{e_{ik} v_i}_1 \Big)
    = \Big( e_{ik} x_k^\top v_i - \epsilon \norm{v_i}_1 \Big) \geq 0.
\end{aligned}
\end{equation*}

Vertically concatenating $e_{ik} x_k^\top v_i - \epsilon \norm{v_i}_1 \geq 0$ for all $i\in[\widehat{P}]$ gives the vectorized representation $E_i X v_i - \epsilon \norm{v_i}_1 \geq 0$, which leads to \cref{eq:robust_constraint}. Since the constraints on $w$ are exactly the same, we also have that $\min\allDeltaU E_i (X+\Delta) w_i \geq 0$ is equivalent to $E_i X w_i - \epsilon \norm{w_i}_1 \geq 0$ for all $i\in[\widehat{P}]$.

\subsection{Proof of \texorpdfstring{\cref{thm:inner_max}}{Theorem 4.3}} \label{sec:INNER_MAX}

The regularization term is independent of $\Delta$. Thus, it can be ignored for the purpose of analyzing the inner maximization. Note that each $D_i$ is diagonal, and its diagonal elements are either 0 or 1. Therefore, the inner maximization of \cref{eq:hinge_ADV_D_minmax} can be analyzed element-wise (i.e. independently maximize the cost at each data point).

The maximization problem of the loss at each data point is:
\begin{equation} \label{eq:hinge_inner_max}
    \max_{\norm{\delta_k}_\infty \leq \epsilon} \bigg( 1 - y_k \sum\iP d_{ik} (x_k^\top+\delta_k^\top) (v_i - w_i) \bigg)_+, \\
\end{equation}
where $d_{ik}$ is the $k\th$ diagonal element of $D_i$ and $\delta_k^\top$ is the $k\th$ row of $\Delta$. One can write:
\begin{equation*}
\begin{aligned}
    & \max_{\norm{\delta_k}_\infty \leq \epsilon} \bigg( 1 - y_k \sum\iP d_{ik} (x_k^\top+\delta_k^\top) (v_i - w_i) \bigg)_+ \\
    & \hspace{35mm} = \bigg( \max_{\norm{\delta_k}_\infty \leq \epsilon} 1 - y_k \sum\iP d_{ik} (x_k^\top+\delta_k^\top) (v_i - w_i) \bigg)_+ \\
    & \hspace{35mm} = \bigg( 1 - y_k \sum\iP d_{ik} x_k^\top (v_i - w_i)
    - \min_{\norm{\delta_k}_\infty \leq \epsilon} \delta_k^\top y_k \sum\iP d_{ik} (v_i - w_i) \bigg)_+ .\\
\end{aligned}
\end{equation*}

The optimal solution to $\displaystyle \min_{\norm{\delta_k}_\infty \leq \epsilon} \delta_k^\top y_k \sum\iP d_{ik} (v_i - w_i)$ is $\displaystyle \delta_{\text{hinge}_k}^\star = - \epsilon \cdot \sgn \Big( y_k \sum\iP d_{ik} (v_i - w_i)^\top \Big)$, or equivalently: 
\begin{equation*} 
    \Delta_\text{hinge}^\star = - \epsilon \cdot \sgn \Big( \sum\iP D_i y (v_i - w_i)^\top \Big).
\end{equation*}

Substituting $\delta_{\text{hinge}_k}^\star$ into \cref{eq:hinge_inner_max}, we find the optimal objective of the optimization problem \cref{eq:hinge_inner_max} to be
\begin{equation*}
\begin{aligned}
    & \bigg ( 1 - y_k \sum\iP d_{ik} x_k^\top (v_i - w_i)
    + \epsilon \bigg|\bigg| y_k \sum\iP d_{ik} (v_i - w_i)\bigg|\bigg| _1 \bigg)_+ \\
    & \hspace{35mm} = \bigg( 1 - y_k \sum\iP d_{ik} x_k^\top (v_i - w_i)
    + \epsilon |y_k| \bigg|\bigg| \sum\iP d_{ik} (v_i - w_i) \bigg|\bigg|_1 \bigg)_+ .
\end{aligned}
\end{equation*}

Therefore, the overall loss function is:
\begin{equation*}
    \frac{1}{n} \: \sum_{k=1}^n \bigg( 1 - y_k \sum\iP d_{ik} x_k^\top (v_i - w_i) + \epsilon |y_k| \bigg|\bigg| \sum\iP d_{ik} (v_i - w_i) \bigg|\bigg|_1 \bigg)_+ .
\end{equation*}

In the case of binary classification, $y = \{-1, 1\}^n$, and thus $|y_k| = 1$ for all $k\in[n]$. Therefore, the above is equivalent to
\begin{equation} \label{hinge_obj}
    \frac{1}{n} \: \sum_{k=1}^n \bigg( 1 - y_k \sum\iP d_{ik} x_k^\top (v_i - w_i) + \epsilon \bigg|\bigg| \sum\iP d_{ik} (v_i - w_i) \bigg|\bigg|_1 \bigg)_+
\end{equation}
which is the objective of \cref{eq:hinge_adv_D}. This completes the proof. $\hfill \square$

\subsection{Proof of \texorpdfstring{\cref{thm:robust_SOCP}}{Theorem E.1}} \label{sec:ROBUST_SOCP}

We first exploit the structure of \cref{eq:sl_minimax_convex} and reformulate it as the following robust second-order cone program (SOCP) by introducing a slack variable $a\in\sR$:
\begin{align} \label{eq:SOCP2}
        & \min_{(v_i, w_i)\iPhat, a} a + \beta\sum\iPhat (\enorm{v_i}+\enorm{w_i}) \\
        & \quad\ \ST \quad  (2 D_i - I_n) X v_i \geq \epsilon \norm{v_i}_1, \;\;
        (2 D_i - I_n) X w_i \geq \epsilon \norm{w_i}_1, \quad \alliinPhat \nonumber \\
        & \qquad \qquad \max\allDelta \begin{Vmatrix} \begin{bmatrix} \sum\iPhat 
        D_i(X+\Delta)(v_i-w_i) - y \\ 2a-\frac{1}{4} \end{bmatrix} \end{Vmatrix}_2 \leq 2a+\tfrac{1}{4}, \quad \alliinPhat. \nonumber
\end{align}

Then, we need to establish the equivalence between \cref{eq:SOCP2} and \cref{eq:SOCP3}. To this end, we consider the constraints of \cref{eq:SOCP2} and argue that these can be recast as the constraints given in \cref{eq:SOCP3}.
One can write:
\begin{align*}
    & \max\allDelta \Bigg|\Bigg| \begin{bmatrix} \sum\iPhat D_i(X+\Delta)(v_i-w_i) - y \\ 2a-\frac{1}{4} \end{bmatrix} \Bigg|\Bigg|_2 \leq 2a+\frac{1}{4} \\
    \Longleftrightarrow &
    \max_{\norm{\delta_k}_\infty\leq \epsilon,\ \forall k\in[n]}  \begin{Vmatrix} \begin{bmatrix} 
        \sum\iPhat d_{i1}(x_1^\top-\delta_1^\top)(v_i-w_i) - y_1 \\ \sum\iPhat d_{i2}(x_2^\top-\delta_2^\top)(v_i-w_i) - y_2\\
        \vdots \\
        \sum\iPhat d_{in}(x_n^\top-\delta_n^\top)(v_i-w_i) - y_n\\
        2a-\frac{1}{4} 
    \end{bmatrix} \end{Vmatrix} _2 \leq 2a+\frac{1}{4} \\
    \Longleftrightarrow & 
    \max_{\norm{\delta_k}_\infty \leq \epsilon,\ \forall k\in[n]} 
    \Bigg(\sum\kn \Big(\sum\iPhat d_{ik}(x_k^\top-\delta_k^\top)(v_i-w_i) - y_k\Big)^2 + \Big(2a-\frac{1}{4}\Big)^2 \Bigg)^\frac{1}{2} \leq 2a+\frac{1}{4},
\end{align*}
where $d_{ik}$ is the $k\th$ diagonal element of $D_i$ and $\delta_k^\top$ is the $k\th$ row of $\Delta$. The above constraints can be rewritten by introducing slack variables $z\in\mathbb{R}^{n+1}$ as
\begin{equation*}
\begin{aligned}
  & \textstyle z_k \geq \Big| \sum\iPhat d_{ik} x_k^\top (v_i-w_i) - y_k \Big| + \epsilon \Bignorm{\sum\iPhat d_{ik} (v_i-w_i)}_1, \ \forall k \in [n] \\
  & z_{n+1} \geq \big| 2a-\tfrac{1}{4} \big|, \quad \norm{z}_2 \leq 2a+\tfrac{1}{4}.
\end{aligned}
\end{equation*}
\hfill $\square$

\subsection{Proof of \texorpdfstring{\cref{thm:CEadv}}{Theorem 4.4}} \label{sec:CEadvthm}

The inner maximization of \cref{eq:ce_adver1} can be analyzed separately for each $y_k$. For every index $k$ such that $y_k=0$, it holds that $\sum\kn \big( -2\yhat_k y_k + \log(e^{2\yhat_k}+1) \big)$ monotonously increases with respect to $\yhat_k$. Thus, we need to find $\delta_k$ that maximizes $\yhat_k$ in order to maximize the objective. Therefore, the worst-case adversary $\delta_k^\star$ is 
\begin{align} \label{eq:deltastar_kzero}
    \delta_{\kzero}^\star = \argmax_{\norm{\delta_k}_\infty \leq \epsilon} \bigg( \sum\iPhat d_{ik} \delta_k^\top (v_i - w_i) \bigg) = \epsilon \cdot \sgn \big( \sum\iPhat d_{ik} (v_i-w_i)^\top \big).
\end{align}

For each index $k$ such that $y_k=1$, it holds that $\sum\kn \big( -2\yhat_k \cdot y_k + \log(e^{2\yhat_k}+1) \big)$ \linebreak monotonously decreases with respect to $\yhat_k$. Thus, we need to minimize $\yhat_k$. Therefore, 
\begin{align} \label{eq:deltastar_kone}
    \delta_{\kone}^\star = \argmin_{\norm{\delta_k}_\infty \leq \epsilon} \bigg( \sum\iPhat d_{ik} \delta_k^\top (v_i - w_i) \bigg) = -\epsilon \cdot \sgn \big( \sum\iPhat d_{ik} (v_i-w_i)^\top \big).
\end{align}

The two cases can be combined as $\delta_k^\star = -\epsilon \cdot \sgn \Big( (2 y_k - 1) \sum\iPhat d_{ik} (v_i-w_i)^\top \Big)$. Concatenating $\delta_1^\star, \dots, \delta_n^\star$ back into the matrix form yields the worst-case perturbation matrix $\Delta_\text{BCE}^\star = -\epsilon \cdot \sgn \Big( (2 y - 1) \sum\iPhat D_i (v_i-w_i)^\top \Big)$.

Moreover, notice that the objective is separable based on those $k$ such that $y_k=0$ and those $k$ such that $y_k=1$:
\begin{align}
    \sum\kn \Big( -2 \yhat_k & y_k + \log (e^{2 \yhat_k} + 1) \Big) \nonumber \\
    = & \sum\kone \Big( -2 \yhat_k + \log( e^{2 \yhat_k} + 1 ) \Big) + \sum\kzero \log \big( e^{2 \yhat_k} + 1 \big) \nonumber \\
    = & \sum\kone \log \Big( \frac{e^{2 \yhat_k} + 1}{e^{2 \yhat_k}} \Big) + \sum\kzero \log \big( e^{2 \yhat_k} + 1 \big) \nonumber \\
    = & \sum\kone \log \big( e^{-2 \yhat_k} + 1 \big) + \sum\kzero \log \big( e^{2 \yhat_k} + 1 \big) \nonumber \\
    = & \sum\kone \log \bigg( \exp \Big( -2 \sum\iPhat d_{ik} x_k^\top (v_i-w_i) + 2 \epsilon \cdot \Bignorm{\sum\iPhat d_{ik} (v_i-w_i)}_1 \Big) + 1 \bigg)  \label{eq:ce_adv_prf1} \\
    & + \sum\kzero \log \bigg( \exp \Big( 2 \sum\iPhat d_{ik} x_k^\top (v_i-w_i) + 2 \epsilon \cdot \Bignorm{\sum\iPhat d_{ik} (v_i-w_i)}_1 \Big) + 1 \bigg)  \label{eq:ce_adv_prf2} \\
    = & \sum\kn \log \Bigg( \exp \bigg( 2 \Big( (2y_k-1) \sum\iPhat d_{ik} x_k^\top (v_i-w_i) + \epsilon \cdot \Bignorm{\sum\iPhat d_{ik} (v_i-w_i)}_1 \Big) \bigg) + 1 \Bigg) \nonumber \\
    = & \sum\kn f \circ g_k \big( \{v_i,w_i\}\iPhat \big), \nonumber
\end{align}
where \cref{eq:ce_adv_prf1} and \cref{eq:ce_adv_prf2} are obtained by substituting in \cref{eq:deltastar_kzero} and \cref{eq:deltastar_kone}, and $f(\cdot)$, $g(\cdot)$ are defined in \cref{eq:convex_ce_adver}. Substituting the term $\sum\kn \big( -2 \yhat_k y_k + \log (e^{2 \yhat_k} + 1) \big)$ in \cref{eq:ce_adver1} with the term $\sum\kn f \circ g_k \big( \{v_i,w_i\}\iPhat \big)$ yields the formulation \cref{eq:convex_ce_adver}. Since the function $f(\cdot)$ is convex non-decreasing and $g(\cdot)$ is convex, the optimization problem \cref{eq:convex_ce_adver} is convex. $\hfill \square$

\subsection{Proof of \texorpdfstring{\cref{lemma:RED_D}}{Lemma F.2}} \label{sec:RED_D}

According to \citep{Pilanci20a}, recovering the ANN weights by substituting \cref{eq:recover_weights} into \cref{eq:theorem_5} leads to
\begin{equation*}
\begin{aligned}
    q^\star = & \min_{(v_i, w_i)\iP} \ell \Bigg( \sum\iP D_i X (v_i - w_i), y \Bigg) + \beta \sum\iP \Big( \norm{v_i}_2 + \norm{w_i}_2 \Big) \\
    = & \min_{(u_j, \alpha_j)\jms} \ell \Bigg( \sum\jms (X u_j)_+ \alpha_j, y \Bigg) + \frac{\beta}{2} \sum\jms \Big( \norm{u_j}_2^2 + \alpha_j^2 \Big)
\end{aligned}
\end{equation*}

Similarly, we can recover the network weights from the solution $(\widetilde{v}_i^\star, \widetilde{w}_i^\star)\itildeP$ of \cref{eq:redundant_D} using
\begin{equation} \label{eq:recover_weights_1} 
    \begin{aligned}
        (\util_{j_{1 i}}, \altil_{j_{1 i}}) = \rbr*{\dfrac{\vits}{\sqrt{\norm{\vits}_2}}, \sqrt{\norm{\vits}_2}}, \;\;
        (\util_{j_{2 i}}, \altil_{j_{2 i}}) = \rbr*{\dfrac{\wits}{\sqrt{\norm{\wits}_2}}, -\sqrt{\norm{\wits}_2}}, \;\; \forall i \in [\widetilde{P}].
    \end{aligned}
\end{equation}

Unlike in \cref{eq:recover_weights}, zero weights are not discarded in \cref{eq:recover_weights_1}. For simplicity, we use $\util_1, \dots, \util_{\mts}$ to refer to the hidden layer weights and use $\altil_1, \dots, \altil_{\mts}$ to refer to the output layer weights recovered using \cref{eq:recover_weights_1}. Since $(\vits, \wits)\itildeP$ is a solution to \cref{eq:redundant_D}, it satisfies $(2 D_i - I_n) X \vits \geq 0$ and $(2 D_i - I_n) X \wits \geq 0$ for all $i \in [\widetilde{P}]$. Thus, we can apply \cref{lemma:RECOVER} to obtain:
\vspace{-1mm}
\begin{align*}
    \widetilde{q}^\star = & \ell \bigg(\sum\itildeP D_i X (\vits - \wits), y \bigg) + \beta \sum\itildeP \Big(\norm{\vits}_2 + \norm{\wits}_2 \Big) \\
    = & \ell \bigg(\sum_{j=1}^{\mts} (X \util_j^\star)_+ \alpha_j, y \bigg) + \frac{\beta}{2} \sum_{j=1}^{\mts} \Big( \norm{\util_j^\star}_2^2 + \altil_j^{\star 2} \Big) \\
    \geq & \min_{(u_j, \alpha_j)_{j=1}^{\mts}} \ell \bigg(\sum_{j=1}^{\mts} (X u_j)_+ \alpha_j, y \bigg) + \frac{\beta}{2} \sum_{j=1}^{\mts} \Big( \norm{u_j}_2^2 + \alpha_j^2 \Big)
\end{align*}

Since $\widetilde{P} \geq P$, $m^\star \leq 2 P$ and $\mts = 2 \widetilde{P}$, we have $\widetilde{m}^\star \geq m^\star$. Therefore, according to Section 2 and Theorem 6 of \citep{Pilanci20a}, we have:
\begin{align*}
    q^\star = & \min_{(u_j, \alpha_j)\jms} \ell \bigg( \sum_{j=1}^{m^\star} (X u_j)_+ \alpha_j, y \bigg) + \frac{\beta}{2} \sum\jms \Big( \norm{u_j}_2^2 + \alpha_j^2 \Big) \\
    = & \min_{(u_j, \alpha_j)_{j=1}^{\widetilde{m}^\star}} \ell \bigg( \sum_{j=1}^{\widetilde{m}^\star} (X u_j)_+ \alpha_j, y \bigg) + \frac{\beta}{2} \sum_{j=1}^{\widetilde{m}^\star} \Big( \norm{u_j}_2^2 + \alpha_j^2 \Big) \leq \widetilde{q}^\star.
\end{align*}

The above inequality $q^\star \leq \widetilde{q}^\star$ shows that an ANN with more than $m$ neurons in the hidden layer will yield the same loss as the ANN with $m$ neurons when optimized.

Note that \cref{eq:redundant_D} can always attain $q^\star$ by simply substituting in the optimal solution of \cref{eq:theorem_5} and assigning zeros to all other additional $v_i$ and $w_i$, implying that $q^\star \geq \widetilde{q}^\star$. Since $q^\star$ is both an upper bound and a lower bound on $\widetilde{q}^\star$, we have $\widetilde{q}^\star = q^\star$. Therefore, as long as all matrices in $\mathcal{D}$ are included, the existence of redundant matrices does not change the optimal objective value. $\hfill \square$

\end{document}